\documentclass{article}

\PassOptionsToPackage{numbers, compress, sort}{natbib}

\usepackage[final]{neurips_2023}

\usepackage[utf8]{inputenc} 
\usepackage[T1]{fontenc}    
\usepackage{hyperref}       
\usepackage{url}            
\usepackage{booktabs}       
\usepackage{amsfonts}       
\usepackage{nicefrac}       
\usepackage{microtype}      
\usepackage{xcolor}         
\hypersetup{
    colorlinks = true,
    linkbordercolor = {white},
    citecolor={olive}
}
\usepackage{multirow}
\usepackage{float}
\usepackage{amssymb}
\usepackage{array}
\usepackage{subcaption}
\usepackage{mathtools} 
\usepackage{nicematrix}
\usepackage{stmaryrd}
\usepackage{relsize}
\usepackage{colortbl}
\usepackage{nccmath}
\usepackage{bbm}

\title{Scalable Transformer for PDE Surrogate Modeling}

\author{%
  Zijie Li \\
  Department of Mechanical Engineering\\
  Carnegie Mellon University\\
  Pittsburgh, PA 15213 \\
  \texttt{zijieli@andrew.cmu.edu} \\
  \And
  Dule Shu \\
  Department of Mechanical Engineering\\
  Carnegie Mellon University\\
  Pittsburgh, PA 15213 \\
  \texttt{dules@andrew.cmu.edu} \\
  \And
  Amir Barati Farimani \\
  Department of Mechanical Engineering\\
  Carnegie Mellon University \\
  Pittsburgh PA, USA \\
  \texttt{barati@cmu.edu}
}
\author{Zijie Li, Dule Shu, Amir Barati Farimani\\
Carnegie Mellon University \\
Mechanical Engineering Department \\
\texttt{\{zijieli, dules\}@andrew.cmu.edu \& barati@cmu.edu} \\
}

\begin{document}

\maketitle
\vspace{-1mm}
\begin{abstract}
  Transformer has shown state-of-the-art performance on various applications and has recently emerged as a promising tool for surrogate modeling of partial differential equations (PDEs). Despite the introduction of linear-complexity attention, applying Transformer to problems with a large number of grid points can be numerically unstable and computationally expensive. In this work, we propose Factorized Transformer (FactFormer), which is based on an axial factorized kernel integral. Concretely, we introduce a learnable projection operator that decomposes the input function into multiple sub-functions with one-dimensional domain. These sub-functions are then evaluated and used to compute the instance-based kernel with an axial factorized scheme. We showcase that the proposed model is able to simulate 2D Kolmogorov flow on a $256\times 256$ grid and 3D smoke buoyancy on a $64\times64\times64$ grid with good accuracy and efficiency. The proposed factorized scheme can serve as a computationally efficient low-rank surrogate for the full attention scheme when dealing with multi-dimensional problems. 
\end{abstract}

\vspace{-2mm}
\section{Introduction}

Various physics processes are modeled by partial differential equations (PDEs), from the interaction between atoms in molecular systems to large-scale cosmological phenomena. Solving PDEs advances the understanding of complex physical phenomena, enabling people to make accurate predictions, and make informed decisions across a wide range of scientific and engineering disciplines. Numerical solvers provide a practical way to simulate and predict PDEs since many PDEs are often difficult to solve analytically. Most numerical solvers divide the continuous domain into a discretized grid and reduce the continuous differential equations to algebraic equations via methods like  finite difference/element/volume methods or spectral method. Despite the theoretical guarantees behind them, their practical realization of specific problems can pose challenges that require careful expertise to overcome, such as a sufficient understanding of the underlying physics, or a fine-tailored mesh that resolves the necessary spatio-temporal scales. The interest in developing user-friendly and efficient PDE solvers, along with the success of deep learning models in many other areas \citep{jumper2021highly, silver2016mastering, He2016Residual, brown2020language}, has facilitated the emergence of neural-network-based PDE solvers, where the neural network can be used to parameterize the solution function of the target equation \citep{raissi2019physics}, or to approximate the solution operator\citep{lu2019deeponet, li2020gno}. Compared to many numerical solvers, neural PDE solvers appear to be more tolerant with coarse discretization \citep{stachenfeld2022learned}, and can be applied without explicit meshing \citep{raissi2019physics}. In addition, knowing the underlying equations are not strictly necessary for neural PDE solvers, which gives them the potential to simplify and accelerate the process of physics simulation based on PDEs.

Among various neural network designs, attention-based models (Transformer) \citep{Attention-NIPS-2017} have become state-of-the-art for a wide array of applications \citep{brown2020language, dosovitskiy2020image, jumper2021highly, carion2020detr}, which gives rise to a recent surge of interest in applying the Transformer to PDE modeling \citep{cao2021galerkin, guo2022transformer, li2023transformer, hao2023gnot, kissas2022learning, liu2022ht, ovadia2023vito, nguyen2022fourierformer, han2022predicting, fonseca2023continuous}. 
By viewing the input sequence as a function sampled on a discretization grid, attention can be interpreted as a learnable kernel integral \citep{kovachki2021neural, kissas2022learning, cao2022understand, guo2022transformer} or a learnable Galerkin projection \citep{cao2021galerkin}, and the sequence-to-sequence Transformer \citep{Attention-NIPS-2017} can be modified correspondingly to be better suited for PDE modeling \citep{kissas2022learning, cao2021galerkin, guo2022transformer, li2023transformer, hao2023gnot, ovadia2023vito, liu2023mitigating}. In these works, attention is typically applied to every grid point in the domain to exploit both the local and non-local structure of the system, and therefore a linear-complexity variant of attention is usually necessary. As the number of grid points grows exponentially with respect to the number of dimensions, this results in a very large attention matrix that computes the interaction between every pair of the grid points (despite this attention matrix is not evaluated explicitly in linear attention). Consequently, cascading a deep stack of attention layers introduces instability and relatively high computational cost on high-resolution grid. To alleviate these issues and improve the scalability of Transformer in PDE modeling, we propose a modified attention mechanism. Our model is inspired by the kernel integral viewpoint of softmax-free attention, with a factorized integration scheme motivated by the inherent low-rank structure of dot-product kernel matrix. More specifically, we propose a multi-dimensional factorized kernel integral with each kernel function in the integral having only single-dimensional domains. To calculate these axial kernels, we propose a learnable integral operator that is able to project the input function with high-dimensional domain into a set of sub-functions with single-dimensional domain. The computation of each axial kernel is quadratic with respect to the number of grid points along that the corresponding axis but does not grow with the number of dimensions, which alleviates the curse of dimensionality in standard attention.
With the modified attention mechanism, our proposed model can scale up to multi-dimensional problems with a large number of grid points and achieve competitive performance compared to state-of-the-art models. Moreover, we show that our factorized attention mechanism can reduce the computational cost and improve stability compared to softmax-free linear attention.\footnote{Code for this project is available at: \url{https://github.com/BaratiLab/FactFormer}.}

\begin{figure}[t]
    \centering
    \hspace{-2mm}
    \includegraphics[width=1.01\textwidth]{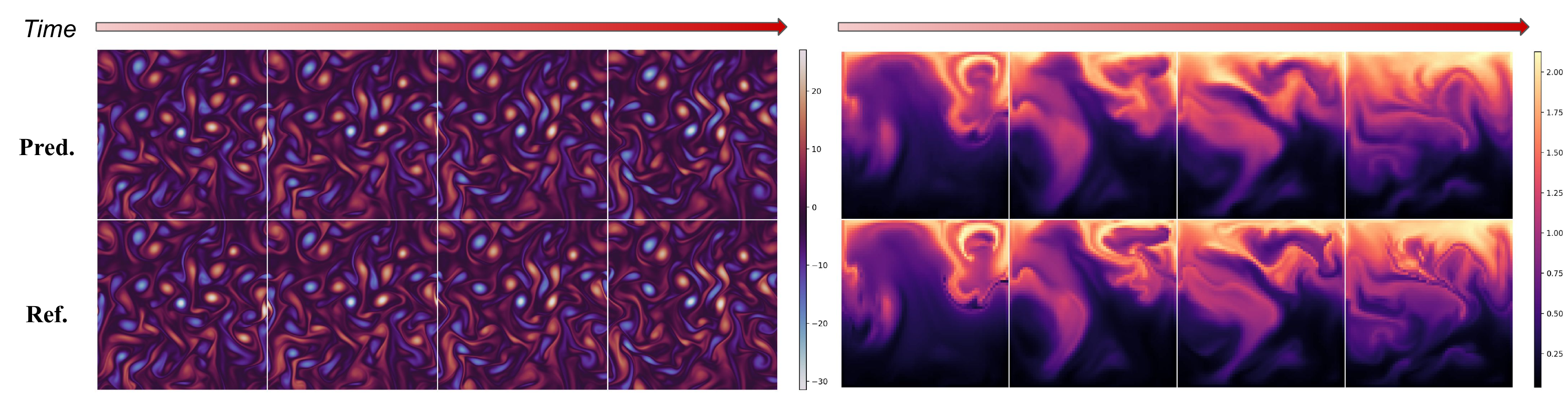}
    \vspace{-3mm}
    \caption{Model's prediction (pred.) and reference ground truth (ref.). \textbf{Left}: 2D Kolmogorov flow on $256\times 256$ grid; \textbf{Right}: 3D smoke buoyancy on $64\times 64\times 64$ grid ($zOy$ cross-section is shown).}
    \vspace{-5mm}
\end{figure}

\section{Related works}
\vspace{-2mm}

\paragraph{Neural PDE solver} 

Based on the emphases of model design, neural PDE solvers can generally be divided into the following groups. The first group of work focuses on using neural networks with mesh-specific architecture design (such as convolutional layers for uniform mesh, or a graph layer for irregular mesh) to learn the spatial and/or temporal correlation of the PDE data \citep{gupta2022towards, stachenfeld2022learned, brandstetter2022message, sanchezgonzalez2020learning, pfaff2021learning, Ummenhofer2020Lagrangian, prantl2022guaranteed, li2022fgn, lotzsch2022learning, Pant2021dlrom, fluidunet-kdd2020, li2019dpi, thuerey2020deep, janny2023eagle}. With input-target data collected, the training process can be conducted without the knowledge of underlying PDEs. This can be appealing when the physics of the system is unknown or partially known, such as large-scale climate modeling \citep{Rasp2020weatherbench, nguyen2023climax, lam2022graphcast, pathak2022fourcastnet}. The second group of work, namely the Physics-Informed Neural Networks (PINNs)\citep{raissi2019physics, zhu2023reliable, pang2019fpinns, lu2021deepxde, cai2021physics, karniadakis2021physics, sun2020surrogate, han2018solving, hao2023physicsinformed}, treat neural networks as a parametrization of the underlying solution function. PINNs incorporate the knowledge of the governing equations into the construction of loss function, which includes the residual of the PDE, the consistency with given boundary condition and initial condition. Unlike the previous group of works, PINNs do not necessarily need input-target data and can be trained solely based on equation loss. The third group of works, often referred to as the neural operator, focuses on learning a mapping between the function spaces\citep{kissas2022learning, kovachki2021neural, brandstetter2022clifford, li2020neural, brandstetter2022lipoint, li2020fourier, li2020multipole, lu2019deeponet, jin2022mionet, gupta2021multiwaveletbased, cao2021galerkin, hao2023gnot, ovadia2023vito, Lu2022fair, bhattacharya2021pcanet}. Neural operator has the generalization capability within a family of PDE and can potentially be adapted to different discretization without retraining. DeepONet \citep{lu2019deeponet} proposes a practical realization of the universal operator approximation theorem \citep{Universal-apprx-operator-IEEE-1995}.  Concurrent work graph neural operator \citep{li2020gno} proposes a learnable kernel integral to approximate the solution operator of parametric PDEs and the follow-up work Fourier Neural Operator (FNO) \citep{li2020fourier} achieves excelling accuracy and efficiency on certain types of problems. Broadly speaking, the operator learning can be conducted upon different types of function bases, such as the Fourier bases \citep{li2020fourier, wen2022ufno, tran2023factfno, guibas2021adaptive, rahman2023uno, kossaifi2023multigrid, kovachki2021universal}, wavelet bases \citep{gupta2021multiwaveletbased}, learned bases in an attention layer \citep{cao2021galerkin, li2023transformer}, or based on approximation of the Green’s function \citep{boulle2022learning, tang2022neural}. The training of neural operators can also be combined with the principle of PINNs to yield a more physically consistent prediction \citep{li2023physicsinformed, pi-deeponet}. Our model is closely related to the neural operator, as the major building blocks in our proposed model are a learnable projection operator and a learnable kernel integral operator.

In addition to direct surrogate modeling, neural networks can also be combined with numerical solvers to improve their accuracy and efficiency. For example, using a trained neural network to correct the error of the solver on the fly \citep{um2020solver, bar2019learning, pathak2020using, Kochkov2021mlcfd, dresdner2022correctspectral}, or doing offline high-fidelity reconstruction \citep{shu2023physics, fukami2019superres, li2022tpugan, Chu2017tempoGAN, meshfreeflownet}. 

\vspace{-2mm}
\paragraph{Transformer for Physics Simulation}
The Transformer model \citep{Attention-NIPS-2017} 
have gained outstanding popularity in natural language modeling \citep{devlin2018bert, brown2020language}, imagery data processing \citep{dosovitskiy2020image} and beyond \citep{jumper2021highly}. In the field of physics simulation, Transformer has drawn increasing research interest as a surrogate model for simulation, with its modeling capability demonstrated both as a neural PDE solver \citep{cao2021galerkin, li2023transformer, nguyen2022fourierformer, guo2022transformer, kissas2022learning, hao2023gnot, ovadia2023vito, han2022predicting, geneva2022transformers, janny2023eagle} and as a pure data-driven model in the absence of a known governing PDE \citep{cachayclimformer, gao2022earthformer, chattopadhyay2020deep, nguyen2023climax}. 
The dot-product attention can be considered as an approximation of an integral transform with a non-symmetric learnable kernel function \citep{cao2021galerkin, wright2021transformers,kissas2022learning, kovachki2021neural, cao2022understand, guo2022transformer}, which relates Transformer to other popular operator learning models such the FNO \citep{li2020fourier}. We will expand the discussion of Transformer under the kernel viewpoint in Section \ref{section:method}.

\vspace{-2mm}
\paragraph{Efficient Transformer}
Following the introduction of Transformer \citep{Attention-NIPS-2017}, various works have investigated ways of reducing the computational cost of standard scaled-dot product attention. The first line of work seeks to remove the softmax and make use of matrix associativity to derive linear complexity attention \citep{transformer-rnn, choromanski2022rethinking, shen2020efficient}, which has also been explored for PDE modeling\citep{cao2021galerkin, hao2023gnot, li2023transformer}. The second line of work tries to approximate the dot product between query and key matrix by exploiting the low-rank structure of it \citep{wang2020linformer, Kitaev2020Reformer, xiong2021nystromformer, zaheer2021bigbird, child2019generating, beltagy2020longformer, ho2020axial}. Our work is related to the first group of works with a softmax-free design, but still calculates the dot product between query and key first. Among the second line of work, Axial Transformer is closely related to our work, as both works have explored conducting attention in an axial fashion. However, the  derivation of attention matrix is different in the two works (see Section \ref{sect:factorized kernel} for detailed comparison). More generally, the exploitation of the multi-dimensional tensor structure in our proposed model can be related to tensor factorization methods \citep{tensor-decomp-review, tensor-train-decomposition} and their applications in various deep learning models \citep{ma2019tensorized, novikov2015tensorizing, yang2017tensorrnn, lebedev2015speedingup, kossaifi2023multigrid}. 
\vspace{-3mm}
\section{Method}
\label{section:method}
\vspace{-1mm}

\subsection{Attention mechanism}
\vspace{-1mm}
\paragraph{Standard attention} Given three sets of vectors, namely the queries $\{\mathbf{q}_i\}_{i=1}^{N_q}$, keys $\{\mathbf{k}_i\}_{i=1}^{N_k}$, and values $\{\mathbf{v}_i\}_{i=1}^{N_v}$ (assuming $N_k=N_v$), attention mechanism \citep{ neural-turing-machine-2014, attention-based-nmt-2015, Neural-Machine-Translation-2014, Attention-NIPS-2017} dynamically computes a weighted average of the values:
    $\mathbf{z}_i
    = \sum_{j=1}^{N_v}h(\mathbf{q}_i, \mathbf{k}_j)\mathbf{v}_j,$ 
where $\mathbf{q}_i, \mathbf{k}_i, \mathbf{v}_i \in \mathbb{R}^{1\times d}$,  $h(\cdot)$ is the weight function that determines the contribution of a specific value to the final output. An example of $h(\cdot)$ is the scaled-dot product with softmax \citep{Attention-NIPS-2017}: $\smaller h(\mathbf{q}_i,  \mathbf{k}_j) = \exp{(\mathbf{q}_i \mathbf{k}_j^T/\tau)} / \sum_{s}\exp{(\mathbf{q}_i \mathbf{k}_s^T/\tau)}$, and $\tau$ is usually chosen as $\tau=\sqrt{d}$. The queries/keys/values are usually obtained from inputs via learnable projection. In self-attention, all of them are computed from the same source as follow: 
\begin{equation}
\small
\label{eq:attn projection}
    \mathbf{q}_i=\mathbf{u}_i W_q, \mathbf{k}_i=\mathbf{u}_i W_k, \mathbf{v}_i=\mathbf{u}_i W_v, 
\end{equation} where $\mathbf{u}_i \in \mathbb{R}^{1\times d_{\text{in}}}$ is the input vector and $\{W_q, W_k, W_v \}\in \mathbb{R}^{d_{\text{in}}\times d}$ are learnbale projection matrices. In cross-attention, queries are derived from one input while keys and values are derived from another.
\vspace{-4mm}

\paragraph{Attention as learnable integral}  Under the hood of PDE modeling, the input sequence to the attention layer can be viewed as the sampling of input function on the discretization grid \citep{kovachki2021neural, cao2021galerkin, kissas2022learning, li2023transformer}. \citet{kovachki2021neural} propose that the scaled-dot product attention \citep{Attention-NIPS-2017} can be viewed as a special case of a Neural Operator \citep{kovachki2021neural}, where the attention amounts to the Monte Carlo approximation of the learnable kernel integral. \citet{cao2021galerkin} further proposes two interpretations of softmax-free attention. The first is to view attention as the Fredholm integral equation of the second kind with a learnable asymmetric dot-product kernel, and the second is to view it as a Peterov-Galerkin projection with learnable basis function. The softmax-free attention proposed by \citet{cao2021galerkin} is later extended in OFormer \citep{li2023transformer}, where Rotary Positional Encoding (RoPE) \citep{su2022roformer} is introduced to modulate the dot product and can be viewed as another special case of the kernel integral in Neural Operator style.\vspace{-1mm}

In this work, we continue on adopting the learnable kernel integral viewpoint of attention and view each channel of the hidden feature map as the sampling of a specific function on the discretization grid. Given query/key/value matrix $\{Q, K, V\} \in \mathbb{R}^{N\times d}$, their row vectors: $\mathbf{q}_i/\mathbf{k}_i/\mathbf{v}_i$, correspond to the sampling of a set of functions $\{q_l(\cdot), k_l(\cdot), v_l(\cdot)\}_{l=1}^d$ on grid point $x_i$, where $\{x_i\}_{i=1}^N$ discretizes the underlying domain. As a more concrete example, the $l$-th column (channel) of $\mathbf{q}_i$, represents the sampling of function $q_l(\cdot)$ on a grid point, i.e. $(\mathbf{q}_i)^l=q_l(x_i)$. Furthermore, softmax-free attention is equivalent to the numerical quadrature of a kernel integral:
\begin{equation}
\small
    (\mathbf{z}_i)^l=\sum_{s=1}^{N}w_s(\mathbf{q}_i \cdot \mathbf{k}_s)(\mathbf{v}_s)^l \approx \int_{\Omega} \kappa\left(x_i, \xi\right)v_l(\xi) d\xi,
    \label{eq:kernel integral}
\end{equation}
where $\mathbf{z}_i$ is the output vector, $\kappa\left(x, \xi\right)=\sum_{l=1}^dq_l(x)k_l(\xi)$ is an instance-based kernel and $w_s$ is the quadrature weight. Understanding attention from the perspective of the kernel has been an active topic of research \citep{wright2021transformers, choromanski2022rethinking, tsai2019transformer, cao2022understand}. The theoretical approximation power of different kernel integrals has also been analyzed under the context of PDE learning \citep{kissas2022learning, guo2022transformer, kovachki2021neural}.
\vspace{-1mm}

Note that the above kernel does not explicitly depend on the spatial coordinates $(x_i, \xi)$. For this work, we opt for a modified kernel formulation proposed in OFormer \citep{li2023transformer}, which modulates the dot product kernel with relative position. Assuming the underlying spatial domain is 1-D (which is sufficient for our proposed model, see next subsection), given query and key vectors $\mathbf{q}_i,\mathbf{k}_j$ and their corresponding spatial coordinates $x_i, x_j$, RoPE \citep{su2022roformer} ($g(\cdot, \cdot):\mathbb{R}^{1\times d}\times\mathbb{R}\mapsto\mathbb{R}^{1\times d}$) is defined as:\vspace{-1mm}
\begin{align}
    \small
    \label{eq:1d-rope}
    g \left(\mathbf{q}_i , x_i \right) = \mathbf{q}_i\mathbf{\Theta}\left(x_i\right), \quad &g \left(\mathbf{k}_j , x_j \right) = \mathbf{k}_j \mathbf{\Theta}\left(x_j\right)\\
    \text{where: }
    \mathbf{\Theta}\left(x_i\right)= \text{Diag}\left( R_1(x_i), \hdots, R_{d/2}(x_i) \right)&, {\small \quad}
    R_l = 
    \begin{bmatrix}
    \cos{(\lambda x_i \theta_l)} & -\sin{(\lambda x_i \theta_l)} \\
    \sin{(\lambda x_i \theta_l)} & \cos{(\lambda x_i \theta_l)}
    \end{bmatrix}, \notag
\end{align}
and $\lambda, \theta_l$ are hyperparameters. $\theta_l$ is usually chosen as $10000^{-2(l-1)/d}, l \in \{1, 2, \hdots, d/2\}$ following \citet{Attention-NIPS-2017} and \citet{su2022roformer}. $\lambda$ is a mesh-based weight that we heuristically set to $64$ throughout most problems. The projection function $\Theta(\cdot): \mathbb{R} \mapsto \mathbb{R}^d\times \mathbb{R}^d$ can explicitly modulate the dot product with relative position: ${\small
    g \left(\mathbf{q}_i , x_i \right) g\left(\mathbf{k}_j , x_j \right)^T = \mathbf{q}_i \Theta(x_i-x_j)\mathbf{k}_j^T}
$, thanks to the following property of rotation matrix: ${\small R_l(x_i)R_l(x_j)^T=R_l(x_i-x_j)}$. 
\vspace{-1mm}

To summarize, we will adopt attention mechanism in the following form for the proposed model (with modification discussed in the next subsection):
\vspace{-1mm}
\begin{equation}
    Z = w \tilde{Q}\tilde{K}^T V,
    \label{eq:matrix attn}
\end{equation}
where $\tilde{\oblong}$ denotes a matrix whose row vectors are RoPE encoded as in \eqref{eq:1d-rope}, e.g., $\tilde{Q}_i=g(\mathbf{q}_i, x_i)$, $w$ is the quadrature weight with a typical choice of $1/N$ for uniform quadrature rule, $Z$ is the output matrix. The query/key/value matrix $Q/K/V$ is derived from the input via learnable projections defined in \eqref{eq:attn projection}. The matrix product $\tilde{Q}\tilde{K}^T$ evaluates the kernel function $\kappa(\cdot,\cdot)$ on the discretization grid $\{x_i\}_{i=1}^N$.
\vspace{-2.5mm}

\subsection{Multidimensional factorized attention}

Compared to the standard scaled dot product attention that has quadratic complexity with respect to the length of the input sequence, the attention in \eqref{eq:matrix attn} can enjoy a linear complexity by making use of the associativity of matrix multiplication (calculate $\small \tilde{K}^TV$ first). In PDE modeling, the length of the input sequence is equal to the number of points on the underlying discretization grid. Assuming the $n$-dimensional domain is discretized by $S_1 \times S_2 \times \hdots \times S_n=N$ points, the softmax-free attention in \eqref{eq:matrix attn} will compute the kernel in \eqref{eq:kernel integral} with the dot product of $\small Q$ and $K$, which are $N$ by $d$ matrices with $N$ usually much larger than $d$. The kernel matrix computed is by design low-rank as it is the product of two tall and thin matrices. Meanwhile, attending a large number of grid points to each other can be unstable and the linear attention has a complexity that is quadratic to the channel dimension $d$, which can limit the scalability of the model in terms of its width. To improve the numerical stability and reduce the computational cost of the aforementioned attention mechanism, we propose a simple yet efficient way to modify the kernel integral discussed in the previous section which is motivated by the low-rank structure of attention. Essentially, our model computes the kernel integral in an axial factorized manner instead of convolving over all the grid points in the domain.
\vspace{-1mm}

For the following discussion, we will use tensor notation \citep{tensor-decomp-review} to describe the operation. We assume the data is represented on a uniform Eulerian grid and can be treated as \textit{$n$-way} tensor $U\in\mathbb{R}^{S_1 \times S_2 \times \hdots \times S_n}$\footnote{In practice we often have an additional mode for the channel, resulting in a \textit{($n+1$)-way} tensor.}. The product of it with a matrix ${ W\in \mathbb{R}^{\mathsmaller{ J\times S_m}}}$ across the $m$-th mode will result in a tensor of shape ${\small S_1  \times \hdots \times S_{m-1} \times J \times S_{m+1} \times \hdots \times S_n}$, whose elements are defined as:
\vspace{-2mm}
\begin{equation}
\smaller
    (U \times _m W)_{i_1 i_2 \hdots i_{m-1} j i_{m+1} \hdots i_n} = \sum_{i_m=1}^{S_m}U_{i_1 i_2\hdots i_m \hdots i_n} W_{ji_m}.
\end{equation}
\vspace{-4mm}

\paragraph{Learnable projection} The first major component of the proposed framework is a set of learnable integral operators $\small \{\mathcal{G}^{(1)}, \mathcal{G}^{(2)},\hdots, \mathcal{G}^{(n)}\}$ that projects the input function $u: \mathbb{R}^{n}\mapsto\mathbb{R}^d$ into a set of functions with one-dimensional domain $\{\phi^{(1)}, \phi^{(2)},\hdots, \phi^{(n)}\}\in\mathbb{R}\mapsto \mathbb{R}^d$, which is defined as:\vspace{-1mm}
\begin{align}
\smaller
    \label{eq:learnable proj}
   &\phi^{(m)}(x^{(m)}_i)=\mathcal{G}^{(m)}(u)(x^{(m)}_i) \\
    &=
    \mathsmaller{h^{(m)}}\left(
     w\medmath{\int}_{\medmath \Omega_1} 
     \medmath{\hdots} \medmath{\int}_{\medmath \Omega_n}
     \mathsmaller{ \gamma^{(m)}}
    \left(u
    \left(\xi_1, \medmath{\hdots}, \xi_{m-1}, \medmath{x^{(m)}_i}, \xi_{m+1} , \medmath{\hdots}, \xi_n\right)
    \right)
    d\xi_1 \medmath{\hdots} d\xi_{m-1} d\xi_{m+1}\medmath{\hdots} d\xi_n
    \right),\notag
\end{align}
where $\smaller h^{(m)}(\cdot) : \mathbb{R}^d \mapsto \mathbb{R}^d$ and $\small \gamma^{(m)}(\cdot): \mathbb{R}^d \mapsto \mathbb{R}^d$ are pointwise learnable functions and $ w=1/\medmath{(L_1L_2\cdots L_{m-1}L_{m+1}\cdots L_n)}$, with $L_m$ being the size of domain $\small \Omega_m$ discretized by $\{x^{(m)}_i\}_{i=1}^{S_m}$. In practice, we implement $h^{(m)}$ as a three-layer multi-layer perception (MLP) similar to the feedforward network in Transformer \citep{Attention-NIPS-2017} and $\gamma^{(m)}$ as a simple linear transformation. When the underlying grid is uniform, \eqref{eq:learnable proj} simply amounts to first transforming the input with pointwise learnable functions $\gamma^{(m)}(\cdot)$, applying mean pooling over all but the $m$-th spatial dimension, and then applying another pointwise learnable function $h^{(m)}$.

\paragraph{Factorized kernel integral}
\label{sect:factorized kernel}
\begin{figure}[t]
    \centering
    \includegraphics[width=0.95\linewidth]{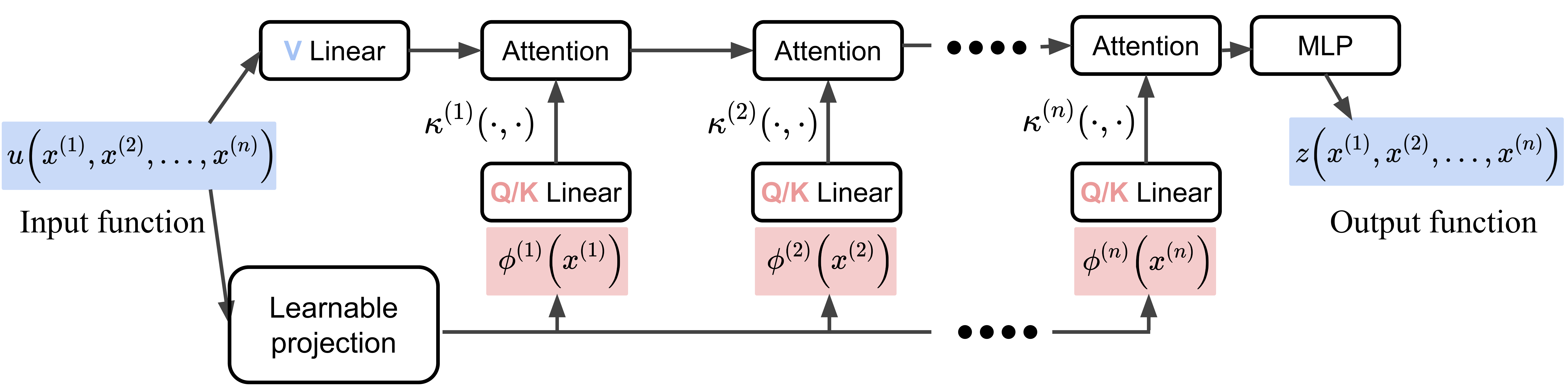}
    \vspace{-1mm}
    \caption{Schematic of the factorized kernel attention. \textbf{Upper path}: the input is transformed into the \textit{Value} via a linear transformation. \textbf{Lower path}: the input is first projected into multiple sub-functions with a one-dimensional domain. These sub-functions are then used to derive the \textit{Query} and \textit{Key} on each axis, and their dot products form the kernel function of the corresponding axis. The \textit{Value} is iteratively updated by the kernel integral transform along each axis and finally sent to an MLP.
    \label{fig:factorized kernel}}
    \vspace{-5mm}
\end{figure}
Equipped with the above projection module, we now introduce our factorized kernel integral scheme. More specifically, we propose to use the following integral to replace the kernel integral in \eqref{eq:kernel integral}:
\vspace{-2mm}
\begin{equation}
\begin{split}
\smaller
    \label{eq:factorized integral}
    &z\left(x^{(1)}_{i_1}, x^{(2)}_{i_2},\hdots, x^{(n)}_{i_n}\right)\\
&= \int_{\Omega_1}\kappa^{(1)}(x^{(1)}_{i_1}, \xi_1)\int_{\Omega_2}\kappa^{(2)}(x^{(2)}_{i_2}, \xi_2)\cdots\int_{\Omega_n}\kappa^{(n)}(x^{(n)}_{i_n}, \xi_n)v(\xi_1,\xi_2,\hdots, \xi_n) d\xi_1 d\xi_2\hdots d\xi_n,
\end{split}
\end{equation}
where kernels $\small \{\kappa^{(1)}, \hdots, \kappa^{(n)}\}: \mathbb{R}\times\mathbb{R}\mapsto\mathbb{R}$ are computed based on projected single-dimensional function along each axis, $v(\cdot): \mathbb{R}^n\mapsto\mathbb{R}^d$ is derived from the input function $u$ via linear transformation (just as the value in standard attention). Next, we will discuss how the above kernel integral is implemented in practice.
Using the learnable projection operator defined in \eqref{eq:learnable proj}, we can obtain $\smaller{ \{\hat{U}^{(1)},\hdots,\hat{U}^{(n)}\}}$ ($\hat{U}^{(m)}\in \mathbb{R}^{\mathsmaller{S_m \times d}}$) from the input $U\in\mathbb{R}^{\mathsmaller{S_1 \times \hdots \times S_n\times d}}$, where the $i_m$-th row of $\hat{U}$ is the evaluation of projected function $\phi^{(m)}(\cdot)$ at $x_{i_m}$: ${\smaller \hat{U}^{(m)}_{i_m}=\phi^{(m)}(x_{i_m})}$. Then we apply linear transformation on them to obtain the query/key matrix just as standard attention:
$
\smaller
    Q^{(m)} = \hat{U}^{(m)}W_q^{(m)}, K^{(m)} = \hat{U}^{(m)}W_k^{(m)},
$
where $\smaller \{W_q^{(m)}, W_k^{(m)}\} \in \mathbb{R}^{d\times d}$ are learnable matrices. The query and key are used to compute the kernel matrix $A^{(m)} \in \mathbb{R}^{S_m\times S_m}$:\vspace{-1mm}
\begin{equation}
\smaller
\label{eq:sub-kernel calculation}
    A^{(m)} =w_m\Tilde{Q}^{(m)} \left(\Tilde{K}^{(m)}\right)^{\mathsmaller T},
\end{equation}
where $w_m$ is the mesh weight, $\tilde{\oblong}$ denotes the RoPE encoded matrix as discussed in \eqref{eq:matrix attn}, the $i$-th row and $j$-th column of $A^{(m)}$ represents the kernel value $\kappa^{(m)}(\mathsmaller{x_{i}^{(m)}, x_{j}^{(m)}})$. Despite \eqref{eq:sub-kernel calculation} has a quadratic complexity with respect to the grid size $S_m$, this is an affordable cost for most of the problems where the axial grid size $S_m$ is mostly between $64$ to $512$. Meanwhile, the value $V\in \mathbb{R}^{\mathsmaller{S_1 \times \hdots \times S_n\times d}}$ is derived from the input via a linear transformation (i.e. $\small (m+1)$-th mode product): $V=U\times_{n+1}W_{v}$, where $ W_{v}\in{\mathbb{R}^{d\times d}}$ is again a learnable matrix. The overall factorized kernel integral is numerically approximated with the following tensor-matrix product (Figure \ref{fig:factorized kernel}):\vspace{-1mm}
\begin{equation}
     Z =\text{Att}(U)= V \times_1 A^{(1)} \times_2 A^{(2)} \times \hdots \times_n A^{(n)}.
     \label{eq:attention in product}
\end{equation}
In \eqref{eq:attention in product}, the computation of all kernel is of complexity $O(S_1^2d+S_2^2d+\hdots+S_n^2d)$, and the time complexity of a single tensor-matrix product $\smaller V\times_{m} A^{(m)}$ is $O(NS_md)$. After evaluating the tensor-matrix product,  the output tensor $Z$ will be sent to a pointwise feedforward network $f(\cdot):\mathbb{R}^d\mapsto\mathbb{R}^d$. To sum up, the update protocol of a single layer in our proposed factorized Transformer is defined as follows:\vspace{-1mm}
\begin{equation}
    U \leftarrow f\left(\text{IN}\left(\text{Att}\left(U\right)\right)\right) + U,
    \label{eq:overall update scheme}
\end{equation}
where $\text{Att}(\cdot)$ is the attention from \eqref{eq:attention in product}, $\text{IN}(\cdot)$ is instance normalization \citep{ulyanov2017instance} that normalizes each channel instance-wise.

\vspace{-1mm}
It is worth pointing out that the axial factorized kernel proposed here shares some similarities with the Axial Transformer proposed in \citet{ho2020axial}, but has two significant differences despite the connection. Firstly, Axial Transformer reduces the computational cost by constraining the context of attention along each axis (e.g. a pixel can only attend to other pixels on the same row), which amounts to moving all but one axis to the batch dimension. In this way computing the axial kernel matrix is of $O(N S_m d)$ complexity (recall $N=S_1\times \hdots \times S_n$) instead of  $O(S_m^2d)$ as in our model. And its overall computation of attention is relatively more expensive due to the presence of softmax. Secondly, the decomposition in Axial Transformer is not layer-wise. For example, in the first layer, the attention is conducted in a row-wise manner and then the second block will conduct attention in a column-wise manner, whereas our model decomposes attention along all axes into a tensor-matrix product within every layer. We provide an illustrative example in Figure \ref{fig:axial vs fact} of the Appendix.

\subsection{Training techniques}
\vspace{-1mm}

In this subsection, we will discuss several techniques used for training the model (including baselines) in our numerical experiments. In general, these techniques aim to alleviate the compounding error of autoregressive neural PDE solvers when applied to time-dependent PDEs. 
\vspace{-3mm}

\paragraph{Latent marching} It is proposed in the \citet{li2023transformer} that a simple pointwise learnable function $\varepsilon(\cdot, \cdot) \in \mathbb{R}^{d}\times \mathbb{R}_{>0}\mapsto \mathbb{R}^d$ can be used to propagate dynamics in the latent space with a fixed time interval $\Delta t$: $z\left(x,t+\Delta t\right)=z\left(x,t\right)+ \varepsilon\left(z\left(x,t\right), t\right)$, where $z$ is the output of the final attention layer. In practice, $\varepsilon$ is implemented as a pointwise MLP and is efficient to compute. 
Leveraging this technique, with one call to the neural solver, we can forward the state for multiple time steps (by marching in the latent space for $k$ steps), thus reducing the total number of calls by a ratio of $k$. This is in principle similar to the \textit{Temporal Bundling} technique proposed in Message-Passing Neural PDE solver (MP-PDE) \citep{brandstetter2023message}, yet different in practical realization. In MP-PDE, the multi-timestep prediction is implemented as first predicting the difference in time $\{d_1, d_2, \hdots, d_k\}$ and then adding them to the input $u_0$ by a forward Euler scheme in the physical space: $\hat{u}_k=u_0+d_k \Delta t$. In this work, we opt for the latent marching to predict multi-timesteps as the forward Euler scheme (in the physical space) is less stable for fluid problems with relatively large time step sizes.
\vspace{-2mm}

\paragraph{Pushforward} Neural PDE solvers are observed to be unstable for time-dependent problems. A small error or perturbation that occurs at the beginning of neural PDE solvers' prediction, can easily result in an unbounded rollout error. While there is hardly a universal method for guaranteeing their stability, a wide array of techniques have been proposed to improve the stability of neural PDE solvers, such as adding physics constraints \citep{wang2021learning, li2023physicsinformed, wang2022respecting}, rollout training \citep{li2020fourier} or adding random-walk noise \citep{sanchezgonzalez2020learning, stachenfeld2022learned, pfaff2021learning}. For this work, we adopt the \textit{pushforward} technique from MP-PDE, which amounts to rolling out the model for two steps during training and then letting the gradient only flows through the last step. This allows training the model on error-corrupted samples and promotes the stability of the model. From a practical perspective, this is straightforward to implement and also computationally much cheaper than standard rollout training.
\vspace{-2mm}

\section{Experiment}
\vspace{-3mm}
In this section, we will investigate our proposed model numerically on several challenging problems. Furthermore, we compare our model against softmax-free attention \citep{cao2021galerkin, li2023transformer}. The baseline models we compared against are Fourier Neural Operator (FNO) \citep{li2020fourier}, Factorized Fourier Neural Operator (F-FNO) \citep{tran2023factfno} and Dilated ResNet (Dil-ResNet) \citep{yu2016multiscale, He2016Residual}. FNO has been shown to have good accuracy on a wide range of PDE problems and is computationally very efficient owing to the Fast Fourier Transformation (FFT). F-FNO factorizes the spectral convolution in FNO into separate spectral convolution along different axes and adopt an improved residual connection formulation like Transformer \cite{Attention-NIPS-2017}. Dil-ResNet is recently introduced by \citet{stachenfeld2022learned} to learn the coarse-grained dynamics of turbulent flow and has demonstrated state-of-the-art performance across several problems. We adopt the implementation of Dil-ResNet with group normalization from PDEArena \citep{gupta2022multispatiotemporalscale}. On 2D steady-state problem where linear attention's computational cost is affordable, we also include the result from Galerkin Transformer \citep{cao2021galerkin}, which uses CNN to project the function onto a coarse grid and applies linear attention on the coarse grid. The implementation details of the proposed model and baselines are available in Section \ref{sec:implementation details}, \ref{sec:baseline details} of the Appendix.
\vspace{-2mm}
\subsection{Benchmark problems}
We first apply our model to three fluid-like systems, where the underlying physics patterns are sensitive to the spatiotemporal scale that discretization can resolve, and typically require fine discretization for classical numerical solvers. In these problems, the neural PDE solver is trained to predict the next frame (or multiple frames if using latent marching) given a context of previous frames. The number of context frames of Kolmogorov flow and isotropic turbulence is set to $10$ following \citep{li2023physicsinformed, li2020fourier}, and $4$ for smoke buoyancy similar to \citep{gupta2022multispatiotemporalscale}. We also consider a well-known steady-state problem-2D Darcy flow,  which has been studied in many of the previous works. Below we provide a brief description of each problem we studied. More details can be found in Section \ref{sec:dataset details} of the Appendix.

\vspace{-2mm}
\paragraph{2D Kolmogorov flow} The first example is 2D Kolmogorov flow governed by incompressible Navier-Stokes equation with a periodic boundary condition. The Reynolds number $\textit{Re}$ determines how turbulent the system will be. We adopt the setting of forced turbulence following \citet{Kochkov2021mlcfd} and generate the data by using the pseudo-spectral method to simulate fluid flow with Reynolds number $\textit{Re}=1000$. The objective is to predict the vorticity $\omega$ of the flow field within an interval $[t_0, t_0+T]$, where $T=1$s and $t_0$ is a random starting point in the sequence. We use a spatial grid of $256\times 256$ and temporal discretization of $\Delta t=0.0625$s (therefore $1$s corresponds to 16 frames) to train and evaluate the model. 
\vspace{-3mm}

\paragraph{3D isotropic turbulence} The second example is 3D isotropic turbulence governed by incompressible Navier-Stokes equation with a periodic boundary condition. The major difference from the first example is that the vortex stretching term is non-zero for three-dimensional flow. We use the 3D spectral simulator from \citet{Mortensen2016spectralDNS}, which simulates the forced turbulence described in \citet{3dturb-dns}. For generating the dataset, we simulate a system of Taylor Reynolds number $\textit{Re}_{\lambda}=84$ \citep{3dturb-dns}. The objective is to predict the pressure $p$ and velocity $\mathbf{u}$ from $t=0.5$ to $t=1$s (10 frames). The model is trained and evaluated on a $60\times60\times60$ spatial grid with $\Delta t=0.05$s. 

\vspace{-3mm}

\paragraph{3D smoke buoyancy} The third example is 3D buoyancy-driven flow, which depicts smoke volume rising in a closed domain. A similar system in 2D formulation has been studied in several previous works \citep{um2020solver, brandstetter2023clifford}. The underlying governing equation is the incompressible Navier-Stokes equation coupled with an advection equation. The boundary condition for the smoke field is Dirichlet while the boundary condition for the flow field is Neumann. The advection equation describes the motion of smoke, which is transported along the flow field. We modify the solver from \citep{gupta2022multispatiotemporalscale} that is implemented in \textit{phiflow}\citep{holl2020phiflow} to generate the data, with buoyancy factor set to $0.5$ and viscosity $\nu=0.003$. The objective is to predict the scalar density field of smoke $d$ and velocity of flow $\mathbf{u}$ from $t=3$ to $t=15$s (16 frames). The model is trained and evaluated on a $64\times64\times64$ spatial grid with $\Delta t=0.75$s. To account for non-periodic boundary conditions, we pad the domain for FNO variants and DilResNet following  the original works. For FactFormer, we append a simple CNN block after the model, which comprises $3$-by-$3$ convolutional layers with zero padding. 
\vspace{-3mm}

\paragraph{2D Darcy flow} In addition to the above time-depedendent systems, the fourth example is 2D steady-state problem from \citet{li2020fourier}. Given the diffusion coefficient, the model predicts the steady-state flow field. The boundary condition is also Dirichlet so we adopt settings for all models similar to the 3D smoke problem. 
\vspace{-2mm}

\subsection{Results and discussion}

For all the models, we study two protocols of training. The first is Latent Marching with Pushforward (denote as \textbf{LM}). The second is simply Autoregressive (denote as \textbf{AR}), where the model is rolled out for two steps during training. For LM models, each call to the model will output $k$ future steps. On 2D Kolmogorov flow/3D smoke buoyancy, $k$ is set to $4$, and $2$ for 3D isotropic turbulence. We interleave pushforward training with standard per-step training for LM models. The relative $L^2$ norm is used to train and measure the error of each model following \citet{li2020fourier}. The sequence-wise averaged error and the frame-wise error at the end frame are reported in Table \ref{tab:2d kmflow benchmark}, \ref{tab:3d iso benchmark}, \ref{tab:3d smoke benchmark}. We also report the time cost of simulating a sequence and the number of parameters for each model. The frame-wise error trends are shown in Figure \ref{fig:err 2d kmflow}, \ref{fig:prs err 3d iso}, \ref{fig:vel err 3d iso}, \ref{fig:dns err 3d smoke}, \ref{fig:vel err 3d smoke} in the Appendix. The visualization of predicted samples are provided in the Section \ref{sec:results vis} of Appendix.

\vspace{-1mm}
 We observe that Dil-ResNet has a slightly better per-frame fitting capability compared to the other models on 3D flow problems. As shown in the loss trend plots, it starts at a lower error compared to other models. This coincides with the observation in \citet{stachenfeld2022learned} where Dil-ResNet's performance is strong on 3D fluid problems. On 2D flow, F-FNO has the best accuracy compared to other models. Interestingly, FactFormer can catch up with Dil-ResNet on 2D Kolmogorov flow and 3D smoke buoyancy when the time duration becomes longer. Yet for shorter-term prediction - 3D isotropic turbulence, Dil-ResNet still has the best final accuracy. This suggests that the accuracy of long-term prediction can potentially benefit from exploiting the global structure that lies in the input. Nonetheless, compared to Dil-ResNet, FactFormer offers superior efficiency as indicated by the inference time (time cost of simulating a sequence). Since the training time is roughly proportional to the model forward time, on 3D problems Dil-ResNet generally takes 3-4 times longer to train. In terms of different training strategies, we find that AR models are less stable than multi-step training (LM) and computationally more expensive as it requires more calls to the neural solver. Despite the average error varies case by case, LM models' error generally accumulates slower on the problems we studied, whereas AR models quickly blow up in some cases. 

 \vspace{-1mm}
 Lastly, while Dil-ResNet has shown good accuracy for 3D flow problems, its performance is highly dependent on the training discretizations. As shown in the Figure \ref{fig:darcy loss}, without changing model architecture, its evaluation errors increases significantly when the resolution increases, while Transformer-based models and FNO models' performance are roughly invariant to the resolution. This highlights a major difference between CNN-based models and neural operators.

  \vspace{-2mm}
\begin{table}[H]
\centering
\scalebox{0.92}{
\begin{tabular}{ccccccccc} 
\toprule
\multirow{2}{*}{Model} & \multicolumn{2}{c}{FNO2D} & \multicolumn{2}{c}{F-FNO2D}  & \multicolumn{2}{c}{Dil-ResNet} & \multicolumn{2}{c}{FactFormer} \\
\cmidrule(lr){2-3}
\cmidrule(lr){4-5}
\cmidrule(lr){6-7}
\cmidrule(lr){8-9}
& AR${}^*$ & LM & AR${}^*$ & LM & AR & LM & AR & LM \\ 
\midrule
 $\omega$ avg. error & 0.3177 & \cellcolor{gray!15}0.2978 &\cellcolor{gray!15}\textbf{0.1486} & 0.2453 & 0.8156 & \cellcolor{gray!15}0.1655 & 0.8835 & \cellcolor{gray!15}0.1734 \\
 $\omega$ final error & 
 \cellcolor{gray!15}0.4423 & 0.4567& \cellcolor{gray!15}\textbf{0.2811}& 0.3861 & 
 1.1692 & \cellcolor{gray!15}0.3051 & 1.0963 & \cellcolor{gray!15}0.3017 \\
\midrule
{\small Inf. time (s)} & 0.73 & 0.81 & 0.86 & 1.01 & 4.69 & 1.78 & 3.14 & 1.38\\ 
{\small \# params (M)} & \multicolumn{2}{c}{85.1} & \multicolumn{2}{c}{3.7}  & \multicolumn{2}{c}{2.4} & \multicolumn{2}{c}{3.5} \\
\bottomrule
\end{tabular}}
\vspace{+1mm}
\captionsetup{width=0.97\linewidth}
\captionof{table}{ Evaluation results of 2D Kolmogorov flow. A batch size of 10 is used for inference. LM models predict 4 steps with each call to the model. Total prediction length is 16 steps. \textbf{AR${}^*$}: Since for 2D problem FNO variants can afford to rollout more steps during training, AR FNO rollout for 12 steps, AR F-FNO rollout for 6 steps, whereas other AR models rollout for 2 steps during training. For model that has complex parameters, each \texttt{cfloat} parameter count as two paramaters. \label{tab:2d kmflow benchmark}
}
\end{table}
%
\vspace{-12mm}
 \begin{table}[H]
 \centering
\scalebox{0.91}{
\begin{tabular}{ccccccccc} 
\toprule
\multirow{2}{*}{Model} & \multicolumn{2}{c}{FNO3D} & \multicolumn{2}{c}{F-FNO3D} & \multicolumn{2}{c}{Dil-ResNet} & \multicolumn{2}{c}{FactFormer} \\
\cmidrule(lr){2-3}
\cmidrule(lr){4-5}
\cmidrule(lr){6-7}
\cmidrule(lr){8-9}

& AR & LM & AR & LM  & AR & LM & AR & LM \\ 
\midrule
 $p$ avg. error 
 & 0.8080 &  \cellcolor{gray!15}0.4634 
 & \cellcolor{gray!15}0.3151 & 0.3264
 & \cellcolor{gray!15}\textbf{0.1725} & 0.1778 
 & 0.2989 &  \cellcolor{gray!15}0.2545 \\
  $p$ final error 
 & 1.1285 & \cellcolor{gray!15}0.6522 
 & 0.4250 & \cellcolor{gray!15}0.4159
 &  0.2573 &  \cellcolor{gray!15}\textbf{0.2448}
 & 0.4407 & \cellcolor{gray!15}0.3431  \\
 \cmidrule{1-1}
$\mathbf{u}$ avg. error  
& 0.3967 & \cellcolor{gray!15}0.3382
& \cellcolor{gray!15}0.2298 & 0.2303
& \cellcolor{gray!15}\textbf{0.1143} & 0.1250 
& 0.1775 & \cellcolor{gray!15}0.1670  \\
$\mathbf{u}$ final error
& 0.6561 & \cellcolor{gray!15}0.4735
& \cellcolor{gray!15}0.2799 & 0.2850
& 0.1675 &  \cellcolor{gray!15}\textbf{0.1671}
& 0.2594 &  \cellcolor{gray!15}0.2218  \\
\midrule
Inf. time (s) & 1.01 & 0.91 & 2.77 & 1.37 & 12.67 & 6.89 & 2.68& 1.31  \\ 
\# params (M) & \multicolumn{2}{c}{509.8} & \multicolumn{2}{c}{3.0} & \multicolumn{2}{c}{6.9} & \multicolumn{2}{c}{5.1} \\
\bottomrule
\end{tabular}}
\vspace{+1mm}
\captionsetup{width=0.97\linewidth}
\caption{ Evaluation results of 3D isotropic turbulence. A batch size of 4 is used for inference. LM models predict 2 steps with each call to the model. Total prediction length is 10 steps. \label{tab:3d iso benchmark}
}
\end{table}
 \begin{table}[H]
 \centering
\scalebox{0.91}{
\begin{tabular}{ccccccccc} 
\toprule
\multirow{2}{*}{Model} & \multicolumn{2}{c}{FNO3D} & \multicolumn{2}{c}{F-FNO3D} & \multicolumn{2}{c}{Dil-ResNet} & \multicolumn{2}{c}{FactFormer} \\
\cmidrule(lr){2-3}
\cmidrule(lr){4-5}
\cmidrule(lr){6-7}
\cmidrule(lr){8-9}

& AR & LM & AR & LM & AR & LM & AR & LM \\ 
\midrule
$d$ avg. error  
& 0.1607 & \cellcolor{gray!15}0.1344 
& \cellcolor{gray!15}0.1038 & 0.1236 &
\cellcolor{gray!15}\textbf{0.0843} & 0.0999 & 0.1017 & \cellcolor{gray!15}0.0942 \\
$d$ final error  
& 0.1775 & \cellcolor{gray!15}0.1287 
& 0.1415 & \cellcolor{gray!15}0.1219
& 0.1070 &\cellcolor{gray!15} 0.1062 &
0.1693 & \cellcolor{gray!15}\textbf{0.0941} \\
\cmidrule{1-1}
$\mathbf{u}$ avg. error  & 
0.5198 & \cellcolor{gray!15}0.4255 
& \cellcolor{gray!15}0.3419 & 0.3713 &
\cellcolor{gray!15}\textbf{0.2378} &  0.2747&
0.3537 & \cellcolor{gray!15}0.2592  \\
$\mathbf{u}$ final error  & 
1.0245 & \cellcolor{gray!15}0.6718 
& 0.8655 & \cellcolor{gray!15}0.6146
& 0.5372 & \cellcolor{gray!15}0.5023
& 0.7881 & \cellcolor{gray!15}\textbf{0.4482}  \\
\midrule
Inf. time (s) & 3.19 & 1.47 & 6.35 & 2.75 & 27.49 &  6.94& 5.61 & 2.62 \\ 
\# params (M) & \multicolumn{2}{c}{509.8} & \multicolumn{2}{c}{3.0} & \multicolumn{2}{c}{6.9} & \multicolumn{2}{c}{4.6} \\
\bottomrule
\end{tabular}}
\vspace{+1mm}
\captionsetup{width=0.97\linewidth}
\caption{ Evaluation results of 3D smoke buoyancy.  A batch size of 4 is used for inference. LM models predict 4 steps with each call to the model. Total prediction length is 16 steps.  \label{tab:3d smoke benchmark}}
\end{table}




\begin{minipage}{0.54\textwidth}
    \centering
    \includegraphics[width=\linewidth]{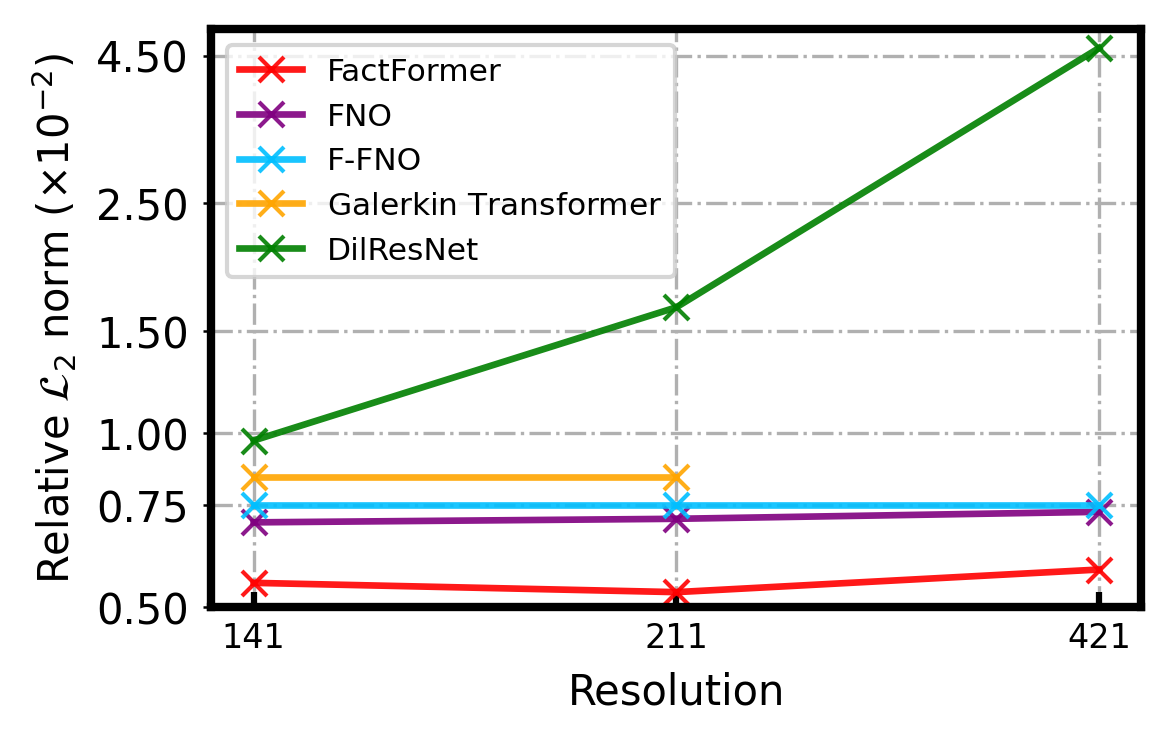}
    \captionsetup{width=0.95\linewidth}
    \vspace{-3mm}
    \captionof{figure}{Error on 2D Darcy flow with different training resolutions. Galerkin Transformer's result is taken from the original paper \citep{cao2021galerkin}.}
    \label{fig:darcy loss}
\end{minipage}
\begin{minipage}{0.45\textwidth}
\centering
    \includegraphics[width=\linewidth]{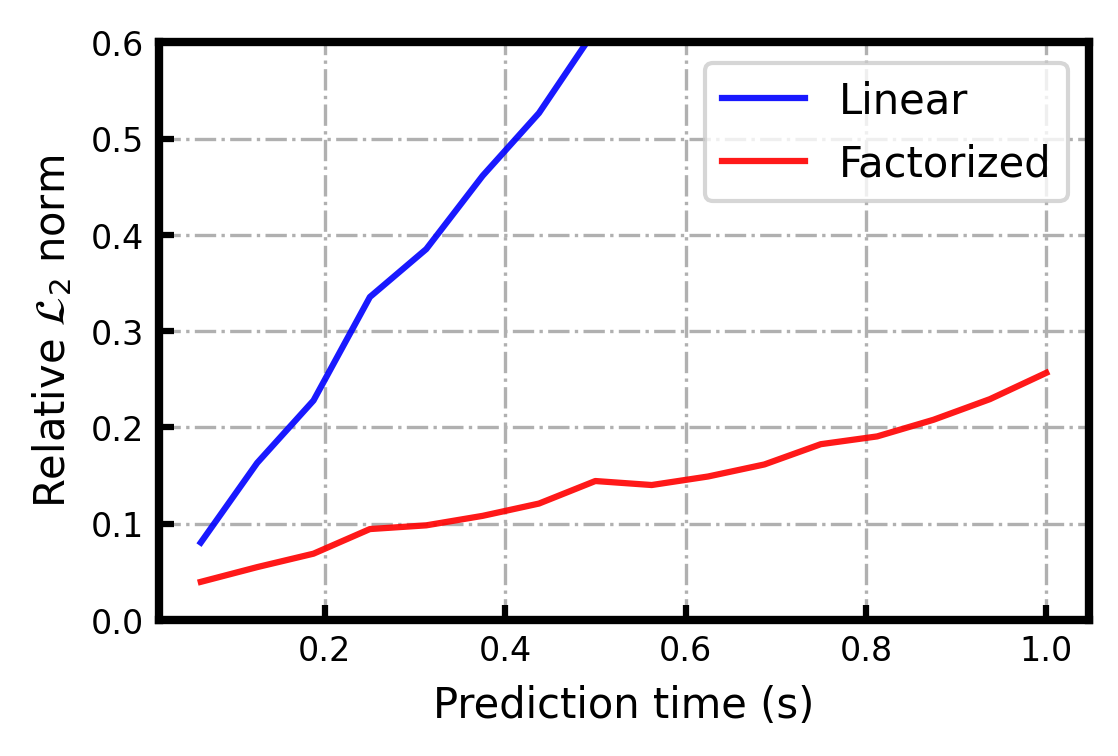}
    \vspace{+0.5mm}
    \captionsetup{width=0.95\linewidth}
    \captionof{figure}{Error trend of different Transformer models on 2D Kolmogorov flow with $128\times128$ grid.}
    \label{fig:err fact-vs-linear}
\end{minipage}

\begin{table}[H]
\centering
\scalebox{0.90}{
\begin{tabular}{cccccc} 
\toprule
\multirow{2}{*}{Model} & \multirow{2}{*}{Avg. rel. $L^2$ Norm} & \multicolumn{2}{c}{Fwd.} & \multicolumn{2}{c}{Fwd. + Bwd.} \\ 
\cmidrule(lr){3-4}
\cmidrule(lr){5-6}
 &  & Enc. time (s) & Prop. time (s) & Time (s) & Mem. (MB) \\ 
\midrule
Factorized attention & 0.1529  & 0.0202  &\multirow{3}{*}{0.0112}  & 0.0954 & 5217  \\
Linear attention (\texttt{matmul}) &\multirow{2}{*}{0.5853} & 0.1370 &  & 0.3013 & 12029 \\
Linear attention (\texttt{einsum}) & & 0.1333 & & 0.2938 & 12029\\
\bottomrule
\end{tabular}}
\vspace{+0.5mm}
\caption{\small Comparison between factorized and linear attention on their forward/backward computational cost, Mem. denotes the peak memory usage. The benchmark is carried out using PyTorch 1.8.2 on an RTX 3090, with a batch size of 4. \textbf{Enc. time}: the time spent on obtaining the latent encoding, primarily includes attention layers and feedforward layers after each attention layer; \textbf{Prop. time}: the time used to propagate dynamics in the latent space with a 3-layer MLP. \label{tab:benchmark attn}
}\vspace{-3mm}
\end{table}

\vspace{-2mm}
\subsection{Comparison against full attention}

In this subsection, we will present an ablation study of the proposed factorized attention mechanism with softmax-free attention (denoted as "linear attention") previously applied to PDE modeling\citep{cao2021galerkin, li2023transformer}. More specifically, we employ the attention from \citet{li2023transformer} (in the form of \eqref{eq:matrix attn}) to replace factorized attention in \eqref{eq:overall update scheme}, with $\small \tilde{K}, V$ normalized column-wise via instance normalization, 
e.g. $\smaller{\left||V_{\cdot,j }|\right|_2=1}$. To accommodate for the memory cost of linear attention, we further downsample the 2D Kolmogorov flow discussed in the last subsection to a $128\times128$ grid and train both linear and factorized attention models on it (with latent marching and pushforward trick).

\vspace{-2mm}
\paragraph{Comparison of performance}
We compare the accuracy and computational cost of the two attention mechanisms in Figure \ref{fig:err fact-vs-linear} and Table \ref{tab:benchmark attn}. While in principle full attention could have better approximation capacity than factorized attention, in practice we find that it performs worse than factorized attention on this problem we studied. Specifically, its rollout is less stable and results in a degraded accuracy. We hypothesize that this is due to the instability of iteratively calculating the attention matrix of a large size, as rolling out the prediction requires recursively calling the model multiple times. In addition to the accuracy improvement, the benchmark on computation empirically demonstrates the computational efficiency improvement of factorized attention over linear attention. We provide more detailed comparison between factorized attention and linear attention in the Section \ref{sec:further ablation study} of Appendix, where we observed consistent efficiency improvement with different grid sizes and model sizes.

\vspace{-3mm}

\begin{figure}[h]
\centering
    \vspace{-2mm}
    \begin{subfigure}{0.31\textwidth}
    \centering
        \includegraphics[width=\linewidth]{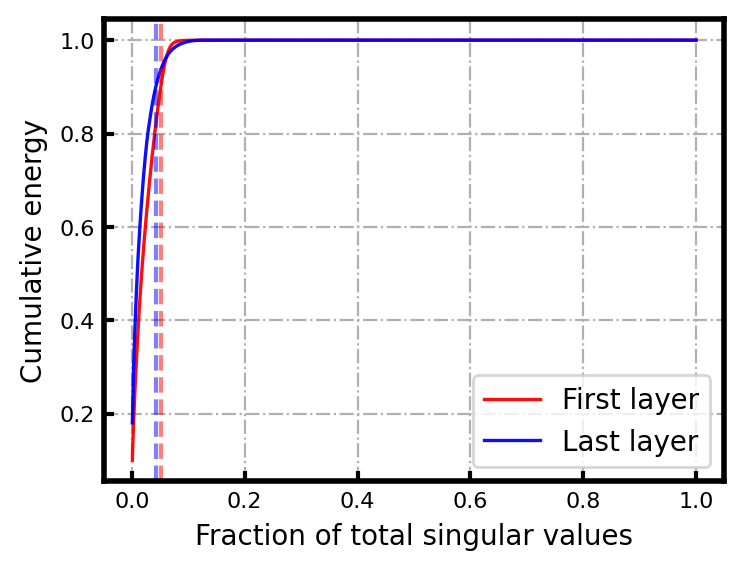}
        \caption{Full attention matrix $A$}
        \label{fig:full attn spectrum}
    \end{subfigure}
    \begin{subfigure}{0.31\textwidth}
    \centering

        \includegraphics[width=\linewidth]{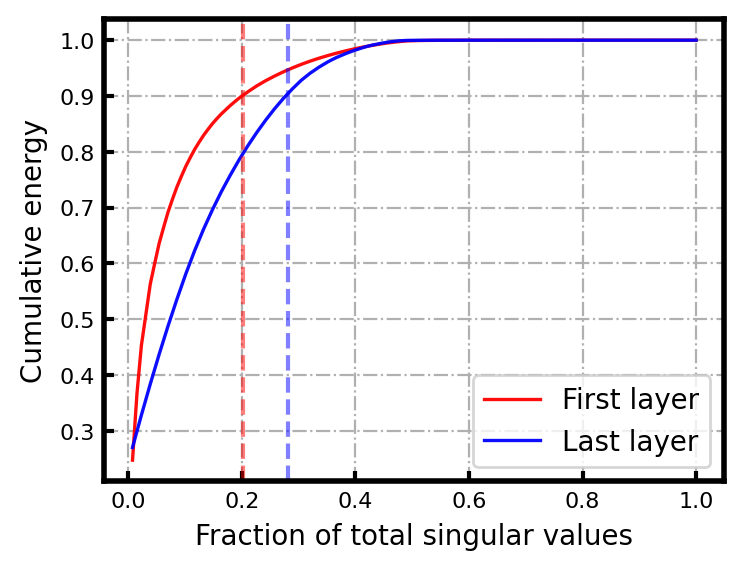}
        \caption{Attention matrix $A^{(1)}$}
        \label{fig:fact attn 1 spectrum}
    \end{subfigure}
    \begin{subfigure}{0.31\textwidth}
    \centering
        \includegraphics[width=\linewidth]{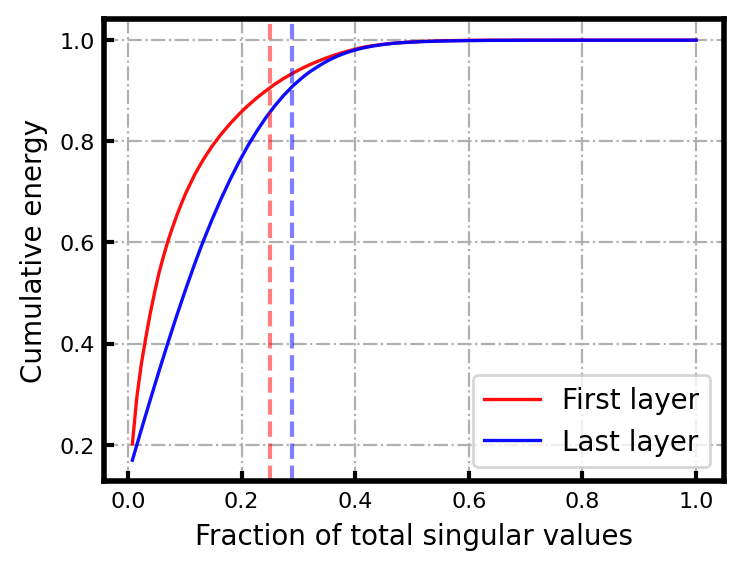}
        \caption{Attention matrix $A^{(2)}$}
        \label{fig:fact attn 2 spectrum}
    \end{subfigure}
    \vspace{-2mm}
    \caption{\small Spectrum of different attention matrices. \textbf{(a)}: Attention matrices from softmax-free attention. \textbf{(b), (c)}: Attention matrices for axis $x$ and $y$ from FactFormer. The vertical line indicates the fraction of singular values that capture $90\%$ of the total energy. Since the full attention matrix $A$ has a relatively large size ($16384 \times 16384$). Therefore its spectrum is computed via TruncatedSVD \citep{halko2010finding} with top $1024$ components truncated.}
    \vspace{-1mm}
\end{figure}

\paragraph{Pattern of attention matrices}
We also investigate the structure of different attention matrices. By construction, when using softmax-free attention to compute the kernel integral in \eqref{eq:kernel integral}, the kernel matrix $A=QK^T$ is going to have a low-rank structure since $\text{rank}(A)\leq \min(\text{rank}(Q), \text{rank}(K))$, $Q, K \in \mathbb{R}^{N\times d}$ and usually$N>>d$. After training, we compute the attention matrices based on $100$ samples and conduct singular value decomposition (SVD) on them. We define the total energy of the spectrum as the sum of singular values  $E=\sum_{i}\sigma_i$, where $\sigma_i$ is the $i$-th singular value and report the normalized cumulative energy histogram $\smaller b_k=\sum_{i=1}^k\sigma_i/\sum_{i}\sigma_i$ in Figure \ref{fig:full attn spectrum}, \ref{fig:fact attn 1 spectrum}, \ref{fig:fact attn 2 spectrum}. For each layer, the histogram is averaged across the attention matrices of all heads. We observe that for linear attention, its rank is relatively low as less than 5\% of the singular values capture over 90\% of the total energy, which is similar to the trend observed from previous works studying the rank of standard softmax-attention \citep{dong2021attention, bhojanapalli2021eigen, wang2020linformer}. Note that the spectrum of linear attention is based on a truncated SVD and therefore its rank will be even lower if a full SVD is performed. The highly low-rank structure of the full attention matrix hints the potential to approximate with or decomposed into smaller and more compact matrices, and our proposed factorized scheme is one example.

\vspace{-3mm}

\section{Conclusion}
\vspace{-2mm}

In this work, we propose an end-to-end Transformer for PDE modeling, which features a learnable projection operator and a factorized kernel integral. We demonstrate that the proposed model balances efficiency and accuracy well, making it a promising and scalable solution for PDE surrogate modeling. However, the proposed attention mechanism is still not free from the curse of dimensionality. The computation of the factorized kernel integral requires evaluating the function on all $S_1\times S_2\times\hdots\times S_m$ grid points. A future direction could be extending the factorization scheme to a more efficient tensor decomposition format like tensor-train. The proposed model currently exploits the uniform structure of the underlying grids and use mean pooling when doing projection, but non-uniform quadrature weight will be necessary when applying to non-uniform grids. It is also observed that the proposed model and other neural PDE solvers can be unstable due to the error accumulation when solving time-dependent systems.

\section*{Acknowledgement}
This work is supported by the National Science Foundation under Grant No. 1953222. The authors would like to thank the annonymous reviewers and area chair for their efforts and valuable feedback during the reviewing process. The authors would like to thank Dr. Shuhao Cao from University of Missouri - Kansas City for the comments regarding the backpropagation of feedforward layer and attention layer in Transformer. The authors would also like to thank Zhi Ye for the suggestions on benchmarking the computational cost of the model.
\newpage
\bibliographystyle{plainnat}
\bibliography{ref}

\begin{thebibliography}{127}
\providecommand{\natexlab}[1]{#1}
\providecommand{\url}[1]{\texttt{#1}}
\expandafter\ifx\csname urlstyle\endcsname\relax
  \providecommand{\doi}[1]{doi: #1}\else
  \providecommand{\doi}{doi: \begingroup \urlstyle{rm}\Url}\fi

\bibitem[lar()]{large-scale-kmflow}
\emph{Large-Scale Dynamics and Transition to Turbulence in the Two-Dimensional Kolmogorov Flow}, pages 374--396.
\newblock \doi{10.2514/5.9781600865831.0374.0396}.
\newblock URL \url{https://arc.aiaa.org/doi/abs/10.2514/5.9781600865831.0374.0396}.

\bibitem[Bahdanau et~al.(2014)Bahdanau, Cho, and Bengio]{Neural-Machine-Translation-2014}
Dzmitry Bahdanau, Kyunghyun Cho, and Yoshua Bengio.
\newblock Neural machine translation by jointly learning to align and translate, 2014.
\newblock URL \url{https://arxiv.org/abs/1409.0473}.

\bibitem[Bar-Sinai et~al.(2019)Bar-Sinai, Hoyer, Hickey, and Brenner]{bar2019learning}
Yohai Bar-Sinai, Stephan Hoyer, Jason Hickey, and Michael~P Brenner.
\newblock Learning data-driven discretizations for partial differential equations.
\newblock \emph{Proceedings of the National Academy of Sciences}, 116\penalty0 (31):\penalty0 15344--15349, 2019.

\bibitem[Beltagy et~al.(2020)Beltagy, Peters, and Cohan]{beltagy2020longformer}
Iz~Beltagy, Matthew~E. Peters, and Arman Cohan.
\newblock Longformer: The long-document transformer, 2020.

\bibitem[Bhattacharya et~al.(2021)Bhattacharya, Hosseini, Kovachki, and Stuart]{bhattacharya2021pcanet}
Kaushik Bhattacharya, Bamdad Hosseini, Nikola~B. Kovachki, and Andrew~M. Stuart.
\newblock Model reduction and neural networks for parametric pdes, 2021.

\bibitem[Bhojanapalli et~al.(2021)Bhojanapalli, Chakrabarti, Jain, Kumar, Lukasik, and Veit]{bhojanapalli2021eigen}
Srinadh Bhojanapalli, Ayan Chakrabarti, Himanshu Jain, Sanjiv Kumar, Michal Lukasik, and Andreas Veit.
\newblock Eigen analysis of self-attention and its reconstruction from partial computation, 2021.

\bibitem[Boull{\'e} et~al.(2022)Boull{\'e}, Kim, Shi, and Townsend]{boulle2022learning}
Nicolas Boull{\'e}, Seick Kim, Tianyi Shi, and Alex Townsend.
\newblock Learning green’s functions associated with time-dependent partial differential equations.
\newblock \emph{Journal of Machine Learning Research}, 23\penalty0 (218):\penalty0 1--34, 2022.

\bibitem[Brandstetter et~al.(2022{\natexlab{a}})Brandstetter, Berg, Welling, and Gupta]{brandstetter2022clifford}
Johannes Brandstetter, Rianne van~den Berg, Max Welling, and Jayesh~K Gupta.
\newblock Clifford neural layers for pde modeling.
\newblock \emph{arXiv preprint arXiv:2209.04934}, 2022{\natexlab{a}}.

\bibitem[Brandstetter et~al.(2022{\natexlab{b}})Brandstetter, Welling, and Worrall]{brandstetter2022lipoint}
Johannes Brandstetter, Max Welling, and Daniel~E Worrall.
\newblock Lie point symmetry data augmentation for neural {PDE} solvers.
\newblock In Kamalika Chaudhuri, Stefanie Jegelka, Le~Song, Csaba Szepesvari, Gang Niu, and Sivan Sabato, editors, \emph{Proceedings of the 39th International Conference on Machine Learning}, volume 162 of \emph{Proceedings of Machine Learning Research}, pages 2241--2256. PMLR, 17--23 Jul 2022{\natexlab{b}}.
\newblock URL \url{https://proceedings.mlr.press/v162/brandstetter22a.html}.

\bibitem[Brandstetter et~al.(2022{\natexlab{c}})Brandstetter, Worrall, and Welling]{brandstetter2022message}
Johannes Brandstetter, Daniel Worrall, and Max Welling.
\newblock Message passing neural pde solvers.
\newblock \emph{arXiv preprint arXiv:2202.03376}, 2022{\natexlab{c}}.

\bibitem[Brandstetter et~al.(2023{\natexlab{a}})Brandstetter, van~den Berg, Welling, and Gupta]{brandstetter2023clifford}
Johannes Brandstetter, Rianne van~den Berg, Max Welling, and Jayesh~K. Gupta.
\newblock Clifford neural layers for pde modeling, 2023{\natexlab{a}}.

\bibitem[Brandstetter et~al.(2023{\natexlab{b}})Brandstetter, Worrall, and Welling]{brandstetter2023message}
Johannes Brandstetter, Daniel Worrall, and Max Welling.
\newblock Message passing neural pde solvers, 2023{\natexlab{b}}.

\bibitem[Brown et~al.(2020)Brown, Mann, Ryder, Subbiah, Kaplan, Dhariwal, Neelakantan, Shyam, Sastry, Askell, et~al.]{brown2020language}
Tom Brown, Benjamin Mann, Nick Ryder, Melanie Subbiah, Jared~D Kaplan, Prafulla Dhariwal, Arvind Neelakantan, Pranav Shyam, Girish Sastry, Amanda Askell, et~al.
\newblock Language models are few-shot learners.
\newblock \emph{Advances in neural information processing systems}, 33:\penalty0 1877--1901, 2020.

\bibitem[Cachay et~al.()Cachay, Mitra, Hirasawa, Kim, Hazarika, Hingmire, Rasch, Singh, and Ramea]{cachayclimformer}
Salva~R{\"u}hling Cachay, Peetak Mitra, Haruki Hirasawa, Sookyung Kim, Subhashis Hazarika, Dipti Hingmire, Phil Rasch, Hansi Singh, and Kalai Ramea.
\newblock Climformer--a spherical transformer model for long-term climate projections.

\bibitem[Cai et~al.(2021)Cai, Mao, Wang, Yin, and Karniadakis]{cai2021physics}
Shengze Cai, Zhiping Mao, Zhicheng Wang, Minglang Yin, and George~Em Karniadakis.
\newblock Physics-informed neural networks (pinns) for fluid mechanics: A review.
\newblock \emph{Acta Mechanica Sinica}, 37\penalty0 (12):\penalty0 1727--1738, 2021.

\bibitem[Cao(2021)]{cao2021galerkin}
Shuhao Cao.
\newblock Choose a transformer: Fourier or galerkin.
\newblock In M.~Ranzato, A.~Beygelzimer, Y.~Dauphin, P.S. Liang, and J.~Wortman Vaughan, editors, \emph{Advances in Neural Information Processing Systems}, volume~34, pages 24924--24940. Curran Associates, Inc., 2021.
\newblock URL \url{https://proceedings.neurips.cc/paper_files/paper/2021/file/d0921d442ee91b896ad95059d13df618-Paper.pdf}.

\bibitem[Cao et~al.(2022)Cao, Xu, and Clifton]{cao2022understand}
Shuhao Cao, Peng Xu, and David~A. Clifton.
\newblock How to understand masked autoencoders, 2022.

\bibitem[Carion et~al.(2020)Carion, Massa, Synnaeve, Usunier, Kirillov, and Zagoruyko]{carion2020detr}
Nicolas Carion, Francisco Massa, Gabriel Synnaeve, Nicolas Usunier, Alexander Kirillov, and Sergey Zagoruyko.
\newblock End-to-end object detection with transformers, 2020.

\bibitem[Chandler and Kerswell(2013)]{chandler2013kmflow}
Gary~J. Chandler and Rich~R. Kerswell.
\newblock Invariant recurrent solutions embedded in a turbulent two-dimensional kolmogorov flow.
\newblock \emph{Journal of Fluid Mechanics}, 722:\penalty0 554–595, 2013.
\newblock \doi{10.1017/jfm.2013.122}.

\bibitem[Chattopadhyay et~al.(2020)Chattopadhyay, Mustafa, Hassanzadeh, and Kashinath]{chattopadhyay2020deep}
Ashesh Chattopadhyay, Mustafa Mustafa, Pedram Hassanzadeh, and Karthik Kashinath.
\newblock Deep spatial transformers for autoregressive data-driven forecasting of geophysical turbulence.
\newblock In \emph{Proceedings of the 10th international conference on climate informatics}, pages 106--112, 2020.

\bibitem[Chen and Chen(1995)]{Universal-apprx-operator-IEEE-1995}
Tianping Chen and Hong Chen.
\newblock Universal approximation to nonlinear operators by neural networks with arbitrary activation functions and its application to dynamical systems.
\newblock \emph{IEEE Transactions on Neural Networks}, 6\penalty0 (4):\penalty0 911--917, 1995.
\newblock \doi{10.1109/72.392253}.

\bibitem[Child et~al.(2019)Child, Gray, Radford, and Sutskever]{child2019generating}
Rewon Child, Scott Gray, Alec Radford, and Ilya Sutskever.
\newblock Generating long sequences with sparse transformers, 2019.

\bibitem[Choromanski et~al.(2022)Choromanski, Likhosherstov, Dohan, Song, Gane, Sarlos, Hawkins, Davis, Mohiuddin, Kaiser, Belanger, Colwell, and Weller]{choromanski2022rethinking}
Krzysztof Choromanski, Valerii Likhosherstov, David Dohan, Xingyou Song, Andreea Gane, Tamas Sarlos, Peter Hawkins, Jared Davis, Afroz Mohiuddin, Lukasz Kaiser, David Belanger, Lucy Colwell, and Adrian Weller.
\newblock Rethinking attention with performers, 2022.

\bibitem[Chu and Thuerey(2017)]{Chu2017tempoGAN}
Mengyu Chu and Nils Thuerey.
\newblock Data-driven synthesis of smoke flows with {CNN}-based feature descriptors.
\newblock \emph{{ACM} Transactions on Graphics}, 36\penalty0 (4):\penalty0 1--14, jul 2017.
\newblock \doi{10.1145/3072959.3073643}.
\newblock URL \url{https://doi.org/10.1145%2F3072959.3073643}.

\bibitem[Devlin et~al.(2018)Devlin, Chang, Lee, and Toutanova]{devlin2018bert}
Jacob Devlin, Ming-Wei Chang, Kenton Lee, and Kristina Toutanova.
\newblock Bert: Pre-training of deep bidirectional transformers for language understanding.
\newblock \emph{arXiv preprint arXiv:1810.04805}, 2018.

\bibitem[Dong et~al.(2021)Dong, Cordonnier, and Loukas]{dong2021attention}
Yihe Dong, Jean-Baptiste Cordonnier, and Andreas Loukas.
\newblock Attention is not all you need: Pure attention loses rank doubly exponentially with depth, 2021.

\bibitem[Dosovitskiy et~al.(2020)Dosovitskiy, Beyer, Kolesnikov, Weissenborn, Zhai, Unterthiner, Dehghani, Minderer, Heigold, Gelly, et~al.]{dosovitskiy2020image}
Alexey Dosovitskiy, Lucas Beyer, Alexander Kolesnikov, Dirk Weissenborn, Xiaohua Zhai, Thomas Unterthiner, Mostafa Dehghani, Matthias Minderer, Georg Heigold, Sylvain Gelly, et~al.
\newblock An image is worth 16x16 words: Transformers for image recognition at scale.
\newblock \emph{arXiv preprint arXiv:2010.11929}, 2020.

\bibitem[Dresdner et~al.(2022)Dresdner, Kochkov, Norgaard, Zepeda-Núñez, Smith, Brenner, and Hoyer]{dresdner2022correctspectral}
Gideon Dresdner, Dmitrii Kochkov, Peter Norgaard, Leonardo Zepeda-Núñez, Jamie~A. Smith, Michael~P. Brenner, and Stephan Hoyer.
\newblock Learning to correct spectral methods for simulating turbulent flows, 2022.

\bibitem[Fonseca et~al.(2023)Fonseca, Zappala, Caro, and van Dijk]{fonseca2023continuous}
Antonio H de~O Fonseca, Emanuele Zappala, Josue~Ortega Caro, and David van Dijk.
\newblock Continuous spatiotemporal transformers.
\newblock \emph{arXiv preprint arXiv:2301.13338}, 2023.

\bibitem[Fukami et~al.(2019)Fukami, Fukagata, and Taira]{fukami2019superres}
Kai Fukami, Koji Fukagata, and Kunihiko Taira.
\newblock Super-resolution reconstruction of turbulent flows with machine learning.
\newblock \emph{Journal of Fluid Mechanics}, 870:\penalty0 106–120, 2019.
\newblock \doi{10.1017/jfm.2019.238}.

\bibitem[Gao et~al.(2022)Gao, Shi, Wang, Zhu, Wang, Li, and Yeung]{gao2022earthformer}
Zhihan Gao, Xingjian Shi, Hao Wang, Yi~Zhu, Yuyang~Bernie Wang, Mu~Li, and Dit-Yan Yeung.
\newblock Earthformer: Exploring space-time transformers for earth system forecasting.
\newblock \emph{Advances in Neural Information Processing Systems}, 35:\penalty0 25390--25403, 2022.

\bibitem[Geneva and Zabaras(2022)]{geneva2022transformers}
Nicholas Geneva and Nicholas Zabaras.
\newblock Transformers for modeling physical systems.
\newblock \emph{Neural Networks}, 146:\penalty0 272--289, 2022.

\bibitem[Graves et~al.(2014)Graves, Wayne, and Danihelka]{neural-turing-machine-2014}
Alex Graves, Greg Wayne, and Ivo Danihelka.
\newblock Neural turing machines, 2014.
\newblock URL \url{https://arxiv.org/abs/1410.5401}.

\bibitem[Guibas et~al.(2021)Guibas, Mardani, Li, Tao, Anandkumar, and Catanzaro]{guibas2021adaptive}
John Guibas, Morteza Mardani, Zongyi Li, Andrew Tao, Anima Anandkumar, and Bryan Catanzaro.
\newblock Adaptive fourier neural operators: Efficient token mixers for transformers.
\newblock \emph{arXiv preprint arXiv:2111.13587}, 2021.

\bibitem[Guo et~al.(2022)Guo, Cao, and Chen]{guo2022transformer}
Ruchi Guo, Shuhao Cao, and Long Chen.
\newblock Transformer meets boundary value inverse problems.
\newblock \emph{arXiv preprint arXiv:2209.14977}, 2022.

\bibitem[Gupta et~al.(2021)Gupta, Xiao, and Bogdan]{gupta2021multiwaveletbased}
Gaurav Gupta, Xiongye Xiao, and Paul Bogdan.
\newblock Multiwavelet-based operator learning for differential equations, 2021.

\bibitem[Gupta and Brandstetter(2022{\natexlab{a}})]{gupta2022multispatiotemporalscale}
Jayesh~K. Gupta and Johannes Brandstetter.
\newblock Towards multi-spatiotemporal-scale generalized pde modeling, 2022{\natexlab{a}}.

\bibitem[Gupta and Brandstetter(2022{\natexlab{b}})]{gupta2022towards}
Jayesh~K Gupta and Johannes Brandstetter.
\newblock Towards multi-spatiotemporal-scale generalized pde modeling.
\newblock \emph{arXiv preprint arXiv:2209.15616}, 2022{\natexlab{b}}.

\bibitem[Halko et~al.(2010)Halko, Martinsson, and Tropp]{halko2010finding}
Nathan Halko, Per-Gunnar Martinsson, and Joel~A. Tropp.
\newblock Finding structure with randomness: Probabilistic algorithms for constructing approximate matrix decompositions, 2010.

\bibitem[Han et~al.(2018)Han, Jentzen, and E]{han2018solving}
Jiequn Han, Arnulf Jentzen, and Weinan E.
\newblock Solving high-dimensional partial differential equations using deep learning.
\newblock \emph{Proceedings of the National Academy of Sciences}, 115\penalty0 (34):\penalty0 8505--8510, 2018.

\bibitem[Han et~al.(2022)Han, Gao, Pfaff, Wang, and Liu]{han2022predicting}
Xu~Han, Han Gao, Tobias Pfaff, Jian-Xun Wang, and Li-Ping Liu.
\newblock Predicting physics in mesh-reduced space with temporal attention.
\newblock \emph{arXiv preprint arXiv:2201.09113}, 2022.

\bibitem[Hao et~al.(2023{\natexlab{a}})Hao, Liu, Zhang, Ying, Feng, Su, and Zhu]{hao2023physicsinformed}
Zhongkai Hao, Songming Liu, Yichi Zhang, Chengyang Ying, Yao Feng, Hang Su, and Jun Zhu.
\newblock Physics-informed machine learning: A survey on problems, methods and applications, 2023{\natexlab{a}}.

\bibitem[Hao et~al.(2023{\natexlab{b}})Hao, Ying, Wang, Su, Dong, Liu, Cheng, Zhu, and Song]{hao2023gnot}
Zhongkai Hao, Chengyang Ying, Zhengyi Wang, Hang Su, Yinpeng Dong, Songming Liu, Ze~Cheng, Jun Zhu, and Jian Song.
\newblock Gnot: A general neural operator transformer for operator learning.
\newblock \emph{arXiv preprint arXiv:2302.14376}, 2023{\natexlab{b}}.

\bibitem[He et~al.(2016)He, Zhang, Ren, and Sun]{He2016Residual}
Kaiming He, Xiangyu Zhang, Shaoqing Ren, and Jian Sun.
\newblock Deep residual learning for image recognition.
\newblock In \emph{Proceedings of the IEEE Conference on Computer Vision and Pattern Recognition (CVPR)}, June 2016.

\bibitem[Ho et~al.(2020)Ho, Kalchbrenner, Weissenborn, and Salimans]{ho2020axial}
Jonathan Ho, Nal Kalchbrenner, Dirk Weissenborn, and Tim Salimans.
\newblock Axial attention in multidimensional transformers, 2020.
\newblock URL \url{https://openreview.net/forum?id=H1e5GJBtDr}.

\bibitem[Holl et~al.()Holl, Koltun, and Um]{holl2020phiflow}
Philipp Holl, Vladlen Koltun, and Kiwon Um.
\newblock phiflow: A differentiable pde solving framework for deep learning via physical simulations.

\bibitem[Jagtap and Karniadakis(2020)]{jagtap2020extended}
Ameya~D Jagtap and George~Em Karniadakis.
\newblock Extended physics-informed neural networks (xpinns): A generalized space-time domain decomposition based deep learning framework for nonlinear partial differential equations.
\newblock \emph{Communications in Computational Physics}, 28\penalty0 (5):\penalty0 2002--2041, 2020.

\bibitem[JANNY et~al.(2023)JANNY, B{\'e}n{\'e}teau, Nadri, Digne, THOME, and Wolf]{janny2023eagle}
Steeven JANNY, Aur{\'e}lien B{\'e}n{\'e}teau, Madiha Nadri, Julie Digne, Nicolas THOME, and Christian Wolf.
\newblock {EAGLE}: Large-scale learning of turbulent fluid dynamics with mesh transformers.
\newblock In \emph{The Eleventh International Conference on Learning Representations}, 2023.
\newblock URL \url{https://openreview.net/forum?id=mfIX4QpsARJ}.

\bibitem[Jiang et~al.(2020)Jiang, Esmaeilzadeh, Azizzadenesheli, Kashinath, Mustafa, Tchelepi, Marcus, Prabhat, and Anandkumar]{meshfreeflownet}
Chiyu~“Max” Jiang, Soheil Esmaeilzadeh, Kamyar Azizzadenesheli, Karthik Kashinath, Mustafa Mustafa, Hamdi~A. Tchelepi, Philip Marcus, Mr~Prabhat, and Anima Anandkumar.
\newblock Meshfreeflownet: A physics-constrained deep continuous space-time super-resolution framework.
\newblock In \emph{SC20: International Conference for High Performance Computing, Networking, Storage and Analysis}, pages 1--15, 2020.
\newblock \doi{10.1109/SC41405.2020.00013}.

\bibitem[Jin et~al.(2022)Jin, Meng, and Lu]{jin2022mionet}
Pengzhan Jin, Shuai Meng, and Lu~Lu.
\newblock Mionet: Learning multiple-input operators via tensor product.
\newblock \emph{SIAM Journal on Scientific Computing}, 44\penalty0 (6):\penalty0 A3490--A3514, 2022.

\bibitem[Jumper et~al.(2021)Jumper, Evans, Pritzel, Green, Figurnov, Ronneberger, Tunyasuvunakool, Bates, {\v{Z}}{\'\i}dek, Potapenko, et~al.]{jumper2021highly}
John Jumper, Richard Evans, Alexander Pritzel, Tim Green, Michael Figurnov, Olaf Ronneberger, Kathryn Tunyasuvunakool, Russ Bates, Augustin {\v{Z}}{\'\i}dek, Anna Potapenko, et~al.
\newblock Highly accurate protein structure prediction with alphafold.
\newblock \emph{Nature}, 596\penalty0 (7873):\penalty0 583--589, 2021.

\bibitem[Karniadakis et~al.(2021)Karniadakis, Kevrekidis, Lu, Perdikaris, Wang, and Yang]{karniadakis2021physics}
George~Em Karniadakis, Ioannis~G Kevrekidis, Lu~Lu, Paris Perdikaris, Sifan Wang, and Liu Yang.
\newblock Physics-informed machine learning.
\newblock \emph{Nature Reviews Physics}, 3\penalty0 (6):\penalty0 422--440, 2021.

\bibitem[Katharopoulos et~al.(2020)Katharopoulos, Vyas, Pappas, and Fleuret]{transformer-rnn}
Angelos Katharopoulos, Apoorv Vyas, Nikolaos Pappas, and Fran{\c{c}}ois Fleuret.
\newblock Transformers are {RNN}s: Fast autoregressive transformers with linear attention.
\newblock In Hal~Daumé III and Aarti Singh, editors, \emph{Proceedings of the 37th International Conference on Machine Learning}, volume 119 of \emph{Proceedings of Machine Learning Research}, pages 5156--5165. PMLR, 13--18 Jul 2020.
\newblock URL \url{https://proceedings.mlr.press/v119/katharopoulos20a.html}.

\bibitem[Kissas et~al.(2022)Kissas, Seidman, Guilhoto, Preciado, Pappas, and Perdikaris]{kissas2022learning}
Georgios Kissas, Jacob Seidman, Leonardo~Ferreira Guilhoto, Victor~M. Preciado, George~J. Pappas, and Paris Perdikaris.
\newblock Learning operators with coupled attention, 2022.

\bibitem[Kitaev et~al.(2020)Kitaev, Kaiser, and Levskaya]{Kitaev2020Reformer}
Nikita Kitaev, Lukasz Kaiser, and Anselm Levskaya.
\newblock Reformer: The efficient transformer.
\newblock In \emph{International Conference on Learning Representations}, 2020.
\newblock URL \url{https://openreview.net/forum?id=rkgNKkHtvB}.

\bibitem[Kochkov et~al.(2021)Kochkov, Smith, Alieva, Wang, Brenner, and Hoyer]{Kochkov2021mlcfd}
Dmitrii Kochkov, Jamie~A. Smith, Ayya Alieva, Qing Wang, Michael~P. Brenner, and Stephan Hoyer.
\newblock Machine learning{\textendash}accelerated computational fluid dynamics.
\newblock \emph{Proceedings of the National Academy of Sciences}, 118\penalty0 (21), may 2021.
\newblock \doi{10.1073/pnas.2101784118}.
\newblock URL \url{https://doi.org/10.1073%2Fpnas.2101784118}.

\bibitem[Kolda and Bader(2009)]{tensor-decomp-review}
Tamara~G. Kolda and Brett~W. Bader.
\newblock Tensor decompositions and applications.
\newblock \emph{SIAM Rev.}, 51\penalty0 (3):\penalty0 455–500, aug 2009.
\newblock ISSN 0036-1445.
\newblock \doi{10.1137/07070111X}.
\newblock URL \url{https://doi.org/10.1137/07070111X}.

\bibitem[Kossaifi et~al.(2023)Kossaifi, Kovachki, Azizzadenesheli, and Anandkumar]{kossaifi2023multigrid}
Jean Kossaifi, Nikola~Borislavov Kovachki, Kamyar Azizzadenesheli, and Anima Anandkumar.
\newblock Multi-grid tensorized fourier neural operator for high resolution {PDE}s, 2023.
\newblock URL \url{https://openreview.net/forum?id=po-oqRst4Xm}.

\bibitem[Kovachki et~al.(2021{\natexlab{a}})Kovachki, Lanthaler, and Mishra]{kovachki2021universal}
Nikola Kovachki, Samuel Lanthaler, and Siddhartha Mishra.
\newblock On universal approximation and error bounds for fourier neural operators, 2021{\natexlab{a}}.

\bibitem[Kovachki et~al.(2021{\natexlab{b}})Kovachki, Li, Liu, Azizzadenesheli, Bhattacharya, Stuart, and Anandkumar]{kovachki2021neural}
Nikola Kovachki, Zongyi Li, Burigede Liu, Kamyar Azizzadenesheli, Kaushik Bhattacharya, Andrew Stuart, and Anima Anandkumar.
\newblock Neural operator: Learning maps between function spaces.
\newblock \emph{arXiv preprint arXiv:2108.08481}, 2021{\natexlab{b}}.

\bibitem[Lam et~al.(2022)Lam, Sanchez-Gonzalez, Willson, Wirnsberger, Fortunato, Pritzel, Ravuri, Ewalds, Alet, Eaton-Rosen, Hu, Merose, Hoyer, Holland, Stott, Vinyals, Mohamed, and Battaglia]{lam2022graphcast}
Remi Lam, Alvaro Sanchez-Gonzalez, Matthew Willson, Peter Wirnsberger, Meire Fortunato, Alexander Pritzel, Suman Ravuri, Timo Ewalds, Ferran Alet, Zach Eaton-Rosen, Weihua Hu, Alexander Merose, Stephan Hoyer, George Holland, Jacklynn Stott, Oriol Vinyals, Shakir Mohamed, and Peter Battaglia.
\newblock Graphcast: Learning skillful medium-range global weather forecasting, 2022.

\bibitem[Lamorgese et~al.(2004)Lamorgese, Caughey, and Pope]{3dturb-dns}
A.~G. Lamorgese, D.~A. Caughey, and S.~B. Pope.
\newblock {Direct numerical simulation of homogeneous turbulence with hyperviscosity}.
\newblock \emph{Physics of Fluids}, 17\penalty0 (1), 12 2004.
\newblock ISSN 1070-6631.
\newblock \doi{10.1063/1.1833415}.
\newblock URL \url{https://doi.org/10.1063/1.1833415}.
\newblock 015106.

\bibitem[Lebedev et~al.(2015)Lebedev, Ganin, Rakhuba, Oseledets, and Lempitsky]{lebedev2015speedingup}
Vadim Lebedev, Yaroslav Ganin, Maksim Rakhuba, Ivan Oseledets, and Victor Lempitsky.
\newblock Speeding-up convolutional neural networks using fine-tuned cp-decomposition, 2015.

\bibitem[Li et~al.(2019)Li, Wu, Tedrake, Tenenbaum, and Torralba]{li2019dpi}
Yunzhu Li, Jiajun Wu, Russ Tedrake, Joshua~B. Tenenbaum, and Antonio Torralba.
\newblock Learning particle dynamics for manipulating rigid bodies, deformable objects, and fluids, 2019.

\bibitem[Li and Farimani(2022)]{li2022fgn}
Zijie Li and Amir~Barati Farimani.
\newblock Graph neural network-accelerated lagrangian fluid simulation.
\newblock \emph{Computers \& Graphics}, 103:\penalty0 201--211, 2022.
\newblock ISSN 0097-8493.
\newblock \doi{https://doi.org/10.1016/j.cag.2022.02.004}.
\newblock URL \url{https://www.sciencedirect.com/science/article/pii/S0097849322000206}.

\bibitem[Li et~al.(2022)Li, Li, and Farimani]{li2022tpugan}
Zijie Li, Tianqin Li, and Amir~Barati Farimani.
\newblock {TPU}-{GAN}: Learning temporal coherence from dynamic point cloud sequences.
\newblock In \emph{International Conference on Learning Representations}, 2022.
\newblock URL \url{https://openreview.net/forum?id=FEBFJ98FKx}.

\bibitem[Li et~al.(2023{\natexlab{a}})Li, Meidani, and Farimani]{li2023transformer}
Zijie Li, Kazem Meidani, and Amir~Barati Farimani.
\newblock Transformer for partial differential equations{\textquoteright} operator learning.
\newblock \emph{Transactions on Machine Learning Research}, 2023{\natexlab{a}}.
\newblock ISSN 2835-8856.
\newblock URL \url{https://openreview.net/forum?id=EPPqt3uERT}.

\bibitem[Li et~al.(2020{\natexlab{a}})Li, Kovachki, Azizzadenesheli, Liu, Bhattacharya, Stuart, and Anandkumar]{li2020fourier}
Zongyi Li, Nikola Kovachki, Kamyar Azizzadenesheli, Burigede Liu, Kaushik Bhattacharya, Andrew Stuart, and Anima Anandkumar.
\newblock Fourier neural operator for parametric partial differential equations.
\newblock \emph{arXiv preprint arXiv:2010.08895}, 2020{\natexlab{a}}.

\bibitem[Li et~al.(2020{\natexlab{b}})Li, Kovachki, Azizzadenesheli, Liu, Bhattacharya, Stuart, and Anandkumar]{li2020gno}
Zongyi Li, Nikola Kovachki, Kamyar Azizzadenesheli, Burigede Liu, Kaushik Bhattacharya, Andrew Stuart, and Anima Anandkumar.
\newblock Neural operator: Graph kernel network for partial differential equations, 2020{\natexlab{b}}.

\bibitem[Li et~al.(2020{\natexlab{c}})Li, Kovachki, Azizzadenesheli, Liu, Bhattacharya, Stuart, and Anandkumar]{li2020multipole}
Zongyi Li, Nikola Kovachki, Kamyar Azizzadenesheli, Burigede Liu, Kaushik Bhattacharya, Andrew Stuart, and Anima Anandkumar.
\newblock Multipole graph neural operator for parametric partial differential equations, 2020{\natexlab{c}}.

\bibitem[Li et~al.(2020{\natexlab{d}})Li, Kovachki, Azizzadenesheli, Liu, Bhattacharya, Stuart, and Anandkumar]{li2020neural}
Zongyi Li, Nikola Kovachki, Kamyar Azizzadenesheli, Burigede Liu, Kaushik Bhattacharya, Andrew Stuart, and Anima Anandkumar.
\newblock Neural operator: Graph kernel network for partial differential equations.
\newblock \emph{arXiv preprint arXiv:2003.03485}, 2020{\natexlab{d}}.

\bibitem[Li et~al.(2023{\natexlab{b}})Li, Zheng, Kovachki, Jin, Chen, Liu, Azizzadenesheli, and Anandkumar]{li2023physicsinformed}
Zongyi Li, Hongkai Zheng, Nikola Kovachki, David Jin, Haoxuan Chen, Burigede Liu, Kamyar Azizzadenesheli, and Anima Anandkumar.
\newblock Physics-informed neural operator for learning partial differential equations, 2023{\natexlab{b}}.

\bibitem[Liu et~al.(2022)Liu, Xu, and Zhang]{liu2022ht}
Xinliang Liu, Bo~Xu, and Lei Zhang.
\newblock Ht-net: Hierarchical transformer based operator learning model for multiscale pdes.
\newblock \emph{arXiv preprint arXiv:2210.10890}, 2022.

\bibitem[Liu et~al.(2023)Liu, Xu, and Zhang]{liu2023mitigating}
Xinliang Liu, Bo~Xu, and Lei Zhang.
\newblock Mitigating spectral bias for the multiscale operator learning with hierarchical attention, 2023.

\bibitem[L{\"o}tzsch et~al.(2022)L{\"o}tzsch, Ohler, and Otterbach]{lotzsch2022learning}
Winfried L{\"o}tzsch, Simon Ohler, and Johannes Otterbach.
\newblock Learning the solution operator of boundary value problems using graph neural networks.
\newblock In \emph{ICML 2022 2nd AI for Science Workshop}, 2022.
\newblock URL \url{https://openreview.net/forum?id=4vx9FQA7wiC}.

\bibitem[Lu et~al.(2019)Lu, Jin, and Karniadakis]{lu2019deeponet}
Lu~Lu, Pengzhan Jin, and George~Em Karniadakis.
\newblock Deeponet: Learning nonlinear operators for identifying differential equations based on the universal approximation theorem of operators.
\newblock \emph{arXiv preprint arXiv:1910.03193}, 2019.

\bibitem[Lu et~al.(2021)Lu, Meng, Mao, and Karniadakis]{lu2021deepxde}
Lu~Lu, Xuhui Meng, Zhiping Mao, and George~Em Karniadakis.
\newblock Deepxde: A deep learning library for solving differential equations.
\newblock \emph{SIAM review}, 63\penalty0 (1):\penalty0 208--228, 2021.

\bibitem[Lu et~al.(2022)Lu, Meng, Cai, Mao, Goswami, Zhang, and Karniadakis]{Lu2022fair}
Lu~Lu, Xuhui Meng, Shengze Cai, Zhiping Mao, Somdatta Goswami, Zhongqiang Zhang, and George~Em Karniadakis.
\newblock A comprehensive and fair comparison of two neural operators (with practical extensions) based on {FAIR} data.
\newblock \emph{Computer Methods in Applied Mechanics and Engineering}, 393:\penalty0 114778, apr 2022.
\newblock \doi{10.1016/j.cma.2022.114778}.
\newblock URL \url{https://doi.org/10.1016%2Fj.cma.2022.114778}.

\bibitem[Luong et~al.(2015)Luong, Pham, and Manning]{attention-based-nmt-2015}
Minh-Thang Luong, Hieu Pham, and Christopher~D. Manning.
\newblock Effective approaches to attention-based neural machine translation, 2015.
\newblock URL \url{https://arxiv.org/abs/1508.04025}.

\bibitem[Ma et~al.(2019)Ma, Zhang, Zhang, Duan, Hou, Song, and Zhou]{ma2019tensorized}
Xindian Ma, Peng Zhang, Shuai Zhang, Nan Duan, Yuexian Hou, Dawei Song, and Ming Zhou.
\newblock A tensorized transformer for language modeling, 2019.

\bibitem[Mortensen and Langtangen(2016)]{Mortensen2016spectralDNS}
Mikael Mortensen and Hans~Petter Langtangen.
\newblock High performance python for direct numerical simulations of turbulent flows.
\newblock \emph{Computer Physics Communications}, 203:\penalty0 53--65, jun 2016.
\newblock \doi{10.1016/j.cpc.2016.02.005}.
\newblock URL \url{https://doi.org/10.1016%2Fj.cpc.2016.02.005}.

\bibitem[Nguyen et~al.(2022)Nguyen, Pham, Nguyen, Nguyen, Osher, and Ho]{nguyen2022fourierformer}
Tan Nguyen, Minh Pham, Tam Nguyen, Khai Nguyen, Stanley Osher, and Nhat Ho.
\newblock Fourierformer: Transformer meets generalized fourier integral theorem.
\newblock \emph{Advances in Neural Information Processing Systems}, 35:\penalty0 29319--29335, 2022.

\bibitem[Nguyen et~al.(2023)Nguyen, Brandstetter, Kapoor, Gupta, and Grover]{nguyen2023climax}
Tung Nguyen, Johannes Brandstetter, Ashish Kapoor, Jayesh~K. Gupta, and Aditya Grover.
\newblock Climax: A foundation model for weather and climate, 2023.

\bibitem[Novikov et~al.(2015)Novikov, Podoprikhin, Osokin, and Vetrov]{novikov2015tensorizing}
Alexander Novikov, Dmitry Podoprikhin, Anton Osokin, and Dmitry Vetrov.
\newblock Tensorizing neural networks, 2015.

\bibitem[Oseledets(2011)]{tensor-train-decomposition}
I.~V. Oseledets.
\newblock Tensor-train decomposition.
\newblock \emph{SIAM Journal on Scientific Computing}, 33\penalty0 (5):\penalty0 2295--2317, 2011.
\newblock \doi{10.1137/090752286}.
\newblock URL \url{https://doi.org/10.1137/090752286}.

\bibitem[Ovadia et~al.(2023)Ovadia, Kahana, Stinis, Turkel, and Karniadakis]{ovadia2023vito}
Oded Ovadia, Adar Kahana, Panos Stinis, Eli Turkel, and George~Em Karniadakis.
\newblock Vito: Vision transformer-operator.
\newblock \emph{arXiv preprint arXiv:2303.08891}, 2023.

\bibitem[Pang et~al.(2019)Pang, Lu, and Karniadakis]{pang2019fpinns}
Guofei Pang, Lu~Lu, and George~Em Karniadakis.
\newblock fpinns: Fractional physics-informed neural networks.
\newblock \emph{SIAM Journal on Scientific Computing}, 41\penalty0 (4):\penalty0 A2603--A2626, 2019.

\bibitem[Pant et~al.(2021)Pant, Doshi, Bahl, and Farimani]{Pant2021dlrom}
Pranshu Pant, Ruchit Doshi, Pranav Bahl, and Amir~Barati Farimani.
\newblock Deep learning for reduced order modelling and efficient temporal evolution of fluid simulations.
\newblock \emph{Physics of Fluids}, 33\penalty0 (10):\penalty0 107101, oct 2021.
\newblock \doi{10.1063/5.0062546}.
\newblock URL \url{https://doi.org/10.1063%2F5.0062546}.

\bibitem[Pathak et~al.(2020)Pathak, Mustafa, Kashinath, Motheau, Kurth, and Day]{pathak2020using}
Jaideep Pathak, Mustafa Mustafa, Karthik Kashinath, Emmanuel Motheau, Thorsten Kurth, and Marcus Day.
\newblock Using machine learning to augment coarse-grid computational fluid dynamics simulations.
\newblock \emph{arXiv preprint arXiv:2010.00072}, 2020.

\bibitem[Pathak et~al.(2022)Pathak, Subramanian, Harrington, Raja, Chattopadhyay, Mardani, Kurth, Hall, Li, Azizzadenesheli, Hassanzadeh, Kashinath, and Anandkumar]{pathak2022fourcastnet}
Jaideep Pathak, Shashank Subramanian, Peter Harrington, Sanjeev Raja, Ashesh Chattopadhyay, Morteza Mardani, Thorsten Kurth, David Hall, Zongyi Li, Kamyar Azizzadenesheli, Pedram Hassanzadeh, Karthik Kashinath, and Animashree Anandkumar.
\newblock Fourcastnet: A global data-driven high-resolution weather model using adaptive fourier neural operators, 2022.

\bibitem[Pfaff et~al.(2021)Pfaff, Fortunato, Sanchez-Gonzalez, and Battaglia]{pfaff2021learning}
Tobias Pfaff, Meire Fortunato, Alvaro Sanchez-Gonzalez, and Peter~W. Battaglia.
\newblock Learning mesh-based simulation with graph networks, 2021.

\bibitem[Prantl et~al.(2022)Prantl, Ummenhofer, Koltun, and Thuerey]{prantl2022guaranteed}
Lukas Prantl, Benjamin Ummenhofer, Vladlen Koltun, and Nils Thuerey.
\newblock Guaranteed conservation of momentum for learning particle-based fluid dynamics, 2022.

\bibitem[Quarteroni and Valli(1999)]{quarteroni1999domain}
Alfio Quarteroni and Alberto Valli.
\newblock \emph{Domain decomposition methods for partial differential equations}.
\newblock Number BOOK. Oxford University Press, 1999.

\bibitem[Rahimi and Recht(2007)]{rff2007}
Ali Rahimi and Benjamin Recht.
\newblock Random features for large-scale kernel machines.
\newblock In J.~Platt, D.~Koller, Y.~Singer, and S.~Roweis, editors, \emph{Advances in Neural Information Processing Systems}, volume~20. Curran Associates, Inc., 2007.
\newblock URL \url{https://proceedings.neurips.cc/paper_files/paper/2007/file/013a006f03dbc5392effeb8f18fda755-Paper.pdf}.

\bibitem[Rahman et~al.(2023)Rahman, Ross, and Azizzadenesheli]{rahman2023uno}
Md~Ashiqur Rahman, Zachary~E. Ross, and Kamyar Azizzadenesheli.
\newblock U-no: U-shaped neural operators, 2023.

\bibitem[Raissi et~al.(2019)Raissi, Perdikaris, and Karniadakis]{raissi2019physics}
Maziar Raissi, Paris Perdikaris, and George~E Karniadakis.
\newblock Physics-informed neural networks: A deep learning framework for solving forward and inverse problems involving nonlinear partial differential equations.
\newblock \emph{Journal of Computational physics}, 378:\penalty0 686--707, 2019.

\bibitem[Rasp et~al.(2020)Rasp, Dueben, Scher, Weyn, Mouatadid, and Thuerey]{Rasp2020weatherbench}
Stephan Rasp, Peter~D. Dueben, Sebastian Scher, Jonathan~A. Weyn, Soukayna Mouatadid, and Nils Thuerey.
\newblock {WeatherBench}: A benchmark data set for data-driven weather forecasting.
\newblock \emph{Journal of Advances in Modeling Earth Systems}, 12\penalty0 (11), nov 2020.
\newblock \doi{10.1029/2020ms002203}.
\newblock URL \url{https://doi.org/10.1029%2F2020ms002203}.

\bibitem[Rogallo(1981)]{Rogallo1981NumericalEI}
Robert~S. Rogallo.
\newblock Numerical experiments in homogeneous turbulence.
\newblock 1981.

\bibitem[Rombach et~al.(2022)Rombach, Blattmann, Lorenz, Esser, and Ommer]{rombach2022highresolution}
Robin Rombach, Andreas Blattmann, Dominik Lorenz, Patrick Esser, and Björn Ommer.
\newblock High-resolution image synthesis with latent diffusion models, 2022.

\bibitem[Sanchez-Gonzalez et~al.(2020)Sanchez-Gonzalez, Godwin, Pfaff, Ying, Leskovec, and Battaglia]{sanchezgonzalez2020learning}
Alvaro Sanchez-Gonzalez, Jonathan Godwin, Tobias Pfaff, Rex Ying, Jure Leskovec, and Peter~W. Battaglia.
\newblock Learning to simulate complex physics with graph networks, 2020.

\bibitem[Shen et~al.(2020)Shen, Zhang, Zhao, Yi, and Li]{shen2020efficient}
Zhuoran Shen, Mingyuan Zhang, Haiyu Zhao, Shuai Yi, and Hongsheng Li.
\newblock Efficient attention: Attention with linear complexities, 2020.

\bibitem[Shu et~al.(2023)Shu, Li, and Farimani]{shu2023physics}
Dule Shu, Zijie Li, and Amir~Barati Farimani.
\newblock A physics-informed diffusion model for high-fidelity flow field reconstruction.
\newblock \emph{Journal of Computational Physics}, 478:\penalty0 111972, 2023.

\bibitem[Silver et~al.(2016)Silver, Huang, Maddison, Guez, Sifre, Van Den~Driessche, Schrittwieser, Antonoglou, Panneershelvam, Lanctot, et~al.]{silver2016mastering}
David Silver, Aja Huang, Chris~J Maddison, Arthur Guez, Laurent Sifre, George Van Den~Driessche, Julian Schrittwieser, Ioannis Antonoglou, Veda Panneershelvam, Marc Lanctot, et~al.
\newblock Mastering the game of go with deep neural networks and tree search.
\newblock \emph{nature}, 529\penalty0 (7587):\penalty0 484--489, 2016.

\bibitem[Stachenfeld et~al.(2022)Stachenfeld, Fielding, Kochkov, Cranmer, Pfaff, Godwin, Cui, Ho, Battaglia, and Sanchez-Gonzalez]{stachenfeld2022learned}
Kimberly Stachenfeld, Drummond~B. Fielding, Dmitrii Kochkov, Miles Cranmer, Tobias Pfaff, Jonathan Godwin, Can Cui, Shirley Ho, Peter Battaglia, and Alvaro Sanchez-Gonzalez.
\newblock Learned coarse models for efficient turbulence simulation, 2022.

\bibitem[Su et~al.(2022)Su, Lu, Pan, Murtadha, Wen, and Liu]{su2022roformer}
Jianlin Su, Yu~Lu, Shengfeng Pan, Ahmed Murtadha, Bo~Wen, and Yunfeng Liu.
\newblock Roformer: Enhanced transformer with rotary position embedding, 2022.

\bibitem[Sun et~al.(2020)Sun, Gao, Pan, and Wang]{sun2020surrogate}
Luning Sun, Han Gao, Shaowu Pan, and Jian-Xun Wang.
\newblock Surrogate modeling for fluid flows based on physics-constrained deep learning without simulation data.
\newblock \emph{Computer Methods in Applied Mechanics and Engineering}, 361:\penalty0 112732, 2020.

\bibitem[Tancik et~al.(2020)Tancik, Srinivasan, Mildenhall, Fridovich-Keil, Raghavan, Singhal, Ramamoorthi, Barron, and Ng]{tancik2020fourier}
Matthew Tancik, Pratul~P. Srinivasan, Ben Mildenhall, Sara Fridovich-Keil, Nithin Raghavan, Utkarsh Singhal, Ravi Ramamoorthi, Jonathan~T. Barron, and Ren Ng.
\newblock Fourier features let networks learn high frequency functions in low dimensional domains, 2020.

\bibitem[Tang et~al.(2022)Tang, Azevedo, Cordonnier, and Solenthaler]{tang2022neural}
Jingwei Tang, Vinicius~C Azevedo, Guillaume Cordonnier, and Barbara Solenthaler.
\newblock Neural green’s function for laplacian systems.
\newblock \emph{Computers \& Graphics}, 107:\penalty0 186--196, 2022.

\bibitem[Thuerey et~al.(2020)Thuerey, Wei{\ss}enow, Prantl, and Hu]{thuerey2020deep}
Nils Thuerey, Konstantin Wei{\ss}enow, Lukas Prantl, and Xiangyu Hu.
\newblock Deep learning methods for reynolds-averaged navier--stokes simulations of airfoil flows.
\newblock \emph{AIAA Journal}, 58\penalty0 (1):\penalty0 25--36, 2020.

\bibitem[Tran et~al.(2023)Tran, Mathews, Xie, and Ong]{tran2023factfno}
Alasdair Tran, Alexander Mathews, Lexing Xie, and Cheng~Soon Ong.
\newblock Factorized fourier neural operators, 2023.

\bibitem[Tsai et~al.(2019)Tsai, Bai, Yamada, Morency, and Salakhutdinov]{tsai2019transformer}
Yao-Hung~Hubert Tsai, Shaojie Bai, Makoto Yamada, Louis-Philippe Morency, and Ruslan Salakhutdinov.
\newblock Transformer dissection: A unified understanding of transformer's attention via the lens of kernel, 2019.

\bibitem[Ulyanov et~al.(2017)Ulyanov, Vedaldi, and Lempitsky]{ulyanov2017instance}
Dmitry Ulyanov, Andrea Vedaldi, and Victor Lempitsky.
\newblock Instance normalization: The missing ingredient for fast stylization, 2017.

\bibitem[Um et~al.(2020)Um, Brand, Fei, Holl, and Thuerey]{um2020solver}
Kiwon Um, Robert Brand, Yun~Raymond Fei, Philipp Holl, and Nils Thuerey.
\newblock Solver-in-the-loop: Learning from differentiable physics to interact with iterative pde-solvers.
\newblock \emph{Advances in Neural Information Processing Systems}, 33:\penalty0 6111--6122, 2020.

\bibitem[Ummenhofer et~al.(2020)Ummenhofer, Prantl, Thuerey, and Koltun]{Ummenhofer2020Lagrangian}
Benjamin Ummenhofer, Lukas Prantl, Nils Thuerey, and Vladlen Koltun.
\newblock Lagrangian fluid simulation with continuous convolutions.
\newblock In \emph{International Conference on Learning Representations}, 2020.
\newblock URL \url{https://openreview.net/forum?id=B1lDoJSYDH}.

\bibitem[Vaswani et~al.(2017)Vaswani, Shazeer, Parmar, Uszkoreit, Jones, Gomez, Kaiser, and Polosukhin]{Attention-NIPS-2017}
Ashish Vaswani, Noam Shazeer, Niki Parmar, Jakob Uszkoreit, Llion Jones, Aidan~N Gomez, \L~ukasz Kaiser, and Illia Polosukhin.
\newblock Attention is all you need.
\newblock In I.~Guyon, U.~Von Luxburg, S.~Bengio, H.~Wallach, R.~Fergus, S.~Vishwanathan, and R.~Garnett, editors, \emph{Advances in Neural Information Processing Systems}, volume~30. Curran Associates, Inc., 2017.
\newblock URL \url{https://proceedings.neurips.cc/paper/2017/file/3f5ee243547dee91fbd053c1c4a845aa-Paper.pdf}.

\bibitem[Wang et~al.(2020{\natexlab{a}})Wang, Kashinath, Mustafa, Albert, and Yu]{fluidunet-kdd2020}
Rui Wang, Karthik Kashinath, Mustafa Mustafa, Adrian Albert, and Rose Yu.
\newblock Towards physics-informed deep learning for turbulent flow prediction.
\newblock In \emph{Proceedings of the 26th ACM SIGKDD International Conference on Knowledge Discovery \& Data Mining}, KDD '20, page 1457–1466, New York, NY, USA, 2020{\natexlab{a}}. Association for Computing Machinery.
\newblock ISBN 9781450379984.
\newblock \doi{10.1145/3394486.3403198}.
\newblock URL \url{https://doi.org/10.1145/3394486.3403198}.

\bibitem[Wang et~al.(2021{\natexlab{a}})Wang, Wang, and Perdikaris]{pi-deeponet}
Sifan Wang, Hanwen Wang, and Paris Perdikaris.
\newblock Learning the solution operator of parametric partial differential equations with physics-informed deeponets.
\newblock \emph{Science Advances}, 7\penalty0 (40):\penalty0 eabi8605, 2021{\natexlab{a}}.
\newblock \doi{10.1126/sciadv.abi8605}.
\newblock URL \url{https://www.science.org/doi/abs/10.1126/sciadv.abi8605}.

\bibitem[Wang et~al.(2021{\natexlab{b}})Wang, Wang, and Perdikaris]{wang2021learning}
Sifan Wang, Hanwen Wang, and Paris Perdikaris.
\newblock Learning the solution operator of parametric partial differential equations with physics-informed deeponets, 2021{\natexlab{b}}.

\bibitem[Wang et~al.(2022)Wang, Sankaran, and Perdikaris]{wang2022respecting}
Sifan Wang, Shyam Sankaran, and Paris Perdikaris.
\newblock Respecting causality is all you need for training physics-informed neural networks, 2022.

\bibitem[Wang et~al.(2020{\natexlab{b}})Wang, Li, Khabsa, Fang, and Ma]{wang2020linformer}
Sinong Wang, Belinda~Z Li, Madian Khabsa, Han Fang, and Hao Ma.
\newblock Linformer: Self-attention with linear complexity.
\newblock \emph{arXiv preprint arXiv:2006.04768}, 2020{\natexlab{b}}.

\bibitem[Wen et~al.(2022)Wen, Li, Azizzadenesheli, Anandkumar, and Benson]{wen2022ufno}
Gege Wen, Zongyi Li, Kamyar Azizzadenesheli, Anima Anandkumar, and Sally~M Benson.
\newblock U-fno—an enhanced fourier neural operator-based deep-learning model for multiphase flow.
\newblock \emph{Advances in Water Resources}, 163:\penalty0 104180, 2022.

\bibitem[Wright and Gonzalez(2021)]{wright2021transformers}
Matthew~A Wright and Joseph~E Gonzalez.
\newblock Transformers are deep infinite-dimensional non-mercer binary kernel machines.
\newblock \emph{arXiv preprint arXiv:2106.01506}, 2021.

\bibitem[Xiong et~al.(2021)Xiong, Zeng, Chakraborty, Tan, Fung, Li, and Singh]{xiong2021nystromformer}
Yunyang Xiong, Zhanpeng Zeng, Rudrasis Chakraborty, Mingxing Tan, Glenn Fung, Yin Li, and Vikas Singh.
\newblock Nystr\"omformer: A nystr\"om-based algorithm for approximating self-attention, 2021.

\bibitem[Yang et~al.(2017)Yang, Krompass, and Tresp]{yang2017tensorrnn}
Yinchong Yang, Denis Krompass, and Volker Tresp.
\newblock Tensor-train recurrent neural networks for video classification, 2017.

\bibitem[Yu and Koltun(2016)]{yu2016multiscale}
Fisher Yu and Vladlen Koltun.
\newblock Multi-scale context aggregation by dilated convolutions, 2016.

\bibitem[Zaheer et~al.(2021)Zaheer, Guruganesh, Dubey, Ainslie, Alberti, Ontanon, Pham, Ravula, Wang, Yang, and Ahmed]{zaheer2021bigbird}
Manzil Zaheer, Guru Guruganesh, Avinava Dubey, Joshua Ainslie, Chris Alberti, Santiago Ontanon, Philip Pham, Anirudh Ravula, Qifan Wang, Li~Yang, and Amr Ahmed.
\newblock Big bird: Transformers for longer sequences, 2021.

\bibitem[Zhu et~al.(2023)Zhu, Zhang, Jiao, Karniadakis, and Lu]{zhu2023reliable}
Min Zhu, Handi Zhang, Anran Jiao, George~Em Karniadakis, and Lu~Lu.
\newblock Reliable extrapolation of deep neural operators informed by physics or sparse observations.
\newblock \emph{Computer Methods in Applied Mechanics and Engineering}, 412:\penalty0 116064, 2023.

\end{thebibliography}

\newpage
\appendix
\section{Model implementation details}
\label{sec:implementation details}

The major hyperparameters are listed in Table \ref{tab: hyperparameter}.

\begin{table}[h]
\centering
\begin{tabular}{ccccc} 
\toprule
Hyperparameter & 2D Kolmogorov & 3D turbulence & 3D smoke & 2D Darcy flow \\ 
\cmidrule(lr){1-1}
\cmidrule(lr){2-5}
Hidden dimension & 128 & 128 & 128 & 128  \\
Depth & 4 & 4 & 3 & 3 \\
Heads & 8 & 6 & 6  & 12 \\
Kernel dimension & 128 & 192 & 192 & 128 \\
Input encoder & 2D Conv & 2D Conv & 2D Conv & MLP \\
Output decoder & MLP & MLP & MLP & MLP \\
\bottomrule
\end{tabular}
\vspace{+2mm}
\caption{Major hyperparameters for FactFormer}
\label{tab: hyperparameter}
\end{table}

\textit{Hidden dimension} indicates the number of channels in the latent space. \textit{Depth} denotes the number of attention layers. Before entering each attention layer, we modulate the latent encoding with positional encoding: $z_i \leftarrow z_i + \psi(x_i)$, where $x_i$ is the Cartesian coordinate of latent encoding $z_i$, $\psi: \mathbb{R}^{n}\mapsto\mathbb{R}^d$ is a random Fourier feature mapping \citep{rff2007, tancik2020fourier} with a learnable linear transformation. \textit{Kernel dimension} is the dimension of each head, which is equivalent to $d_{k}$, the number of function bases used to compute the kernel: $\kappa(x,\xi)=\sum_{l}^{d_k}q_l(x)k_l(\xi)$.
We train the model with AdamW optimizer and cyclic learning rate scheduler with a maximum learning rate $3e-4$, similar to prior Transformer-for-PDE works \citep{li2023transformer, hao2023gnot, cao2021galerkin}.

For 3D smoke buoyancy and 2D Darcy flow problem, we append a CNN block after attention layers to better account for the boundary values. The CNN block has a U-shape arrangement with 4 CNN layers, with all layers using a kernel size of 3 and padding size of 1 (pad with zeros). The first CNN layer has a stride of 2, while other layers have a stride of 1. The stride-2 convolution will downsample the data by half, so nearest upsampling is applied between the second and third CNN layers to recover the spatial resolution. 

On top of every model (including baselines we will discuss in the next section), we use a 2D convolutional layer to compress the temporal dimension if it is a time-dependent problem. Concretely, we first reshape the input into $(N, T_{\text{in}})$ where $N$ is the number of spatial grid points and $T_{\text{in}}$ is the number of input frames. Then we apply 2D convolution filters of size $(1, T_{\text{in}})$ to compress the temporal dimension to $1$. At the bottom of every model, we use a three-layer MLP to project the latent encoding back to variables of interest such as pressure and velocities. In addition, we adopt a curriculum training strategy for all latent marching models, where we only march for 1 step at the beginning of the training and don't do any pushforward. Then we gradually increase the latent marching steps throughout the training and apply pushforward when the model has been trained for around $6\%$ of the total epochs.

\section{Baseline implementation details}
\label{sec:baseline details}
In this section, we provide the full details of baseline models, namely FNO, F-FNO, Dil-ResNet, and linear attention Transformer.

For Fourier Neural Operator, the implementation is taken from \citet{li2023physicsinformed}'s official implementation: \url{https://github.com/neuraloperator/physics_informed}. And for Factorized Fourier Neural Operator the implementation is taken from \url{https://github.com/alasdairtran/fourierflow}. We add group normalization before the final fully-connected layer. We use a hidden size of 96, a mode number of 12 for 3D problems, 24 for 2D turbulence, 20 for Darcy flow, and a layer number of 4.

For Dil-ResNet, we adopt the implementation from \citet{gupta2022multispatiotemporalscale}:\url{https://github.com/microsoft/pdearena}. Compared to \citet{gupta2022multispatiotemporalscale} and \citet{stachenfeld2022learned}, we simplify the setting by truncating the number of layers inside each block, where we use dilation layers $[1, 3, 8, 3, 1]$ ($[1, 2, 4, 2, 1]$ is chosen for the 3D problem as it has slightly better performance) inside each block instead of $[1, 2, 4, 8, 4, 2, 1]$. The primary reason is that without truncation, the training on 3D problems will take over a week (and cannot fit into a single A6000 GPU for 2-step rollout training), which is significantly slower than other models. In summary, we use 3 residual blocks (each with 5 CNN layers of width 128), an MLP-based or 2D convolution-based (for convolution in temporal domain) encoder/decoder, with group normalization inserted between every residual block. The implementation of the linear attention Transformer follows OFormer's \citep{li2023transformer} attention implementation \url{https://github.com/BaratiLab/OFormer}, with a Galerkin style  normalization scheme \citep{cao2021galerkin}. Other hyperparameters are kept the same as the hyperparameters listed in Table \ref{tab: hyperparameter} - 2D Kolmogorov. For non-Transformer models, they are trained with Adam optimizer and decay learning rate from $5e-4$ to $5e-6$ throughout the training.

All the experiments are carried out using PyTorch 1.8 except for FNO/F-FNO experiment, which uses PyTorch 1.13 for optimizing complex-valued parameters. We train all LM models for 100k iterations and AR models for 64k iterations of gradient updates.

\section{Visualization of error trend}

This section includes the average frame-wise error trend for the time-dependent systems we have investigated.

\vspace{-3mm}
\begin{figure}[h]
    \centering
    \includegraphics[width=0.60\linewidth]{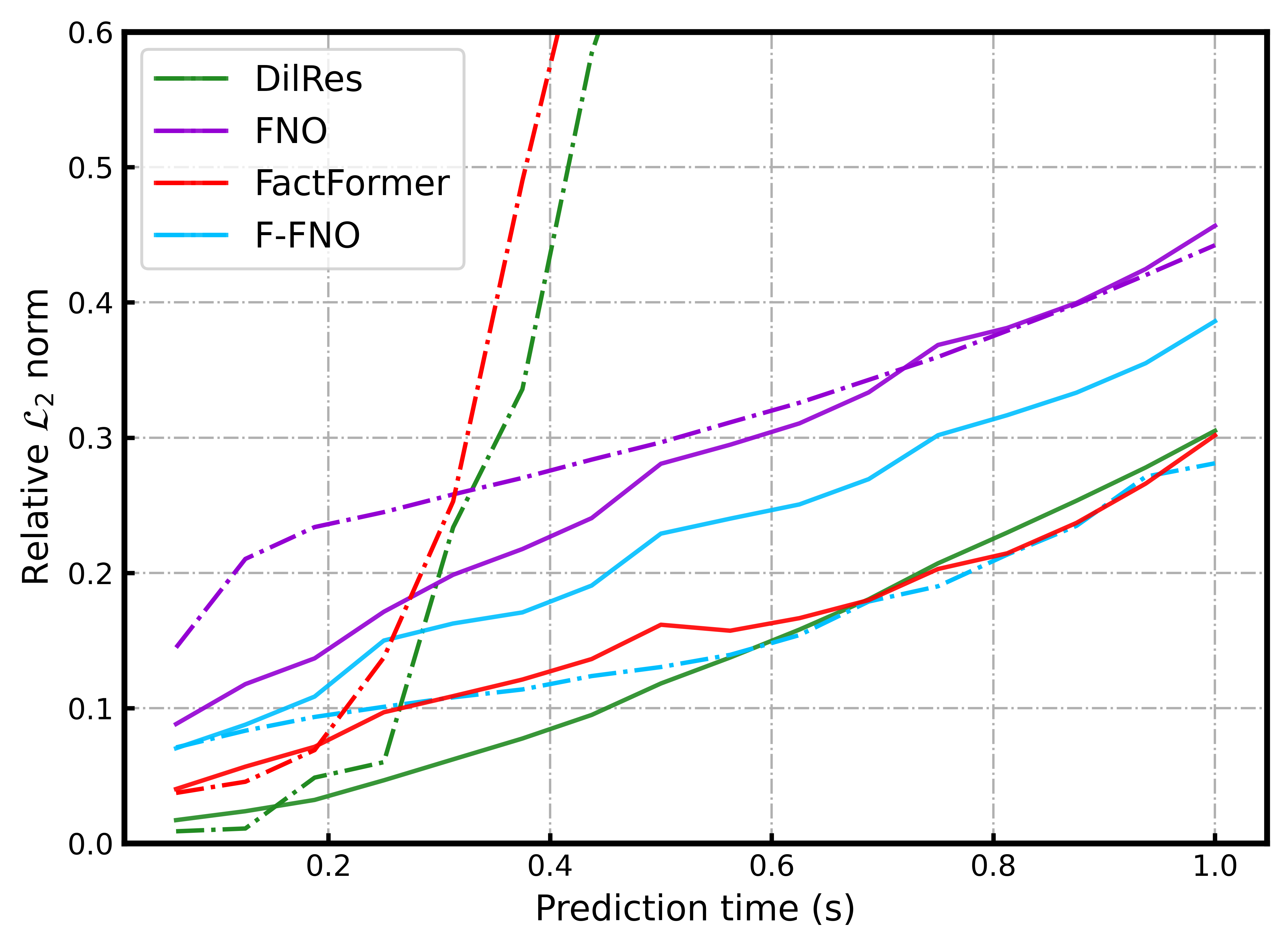}
    \caption{Error trend of vorticity $\omega$ on 2D Kolmogorov flow. \\\textbf{Dashed line}: AR; \textbf{Solid line}: LM}
    \label{fig:err 2d kmflow}
\end{figure}
\begin{minipage}{0.49\textwidth}
    \vspace{-2mm}
     \includegraphics[width=1.0\textwidth]{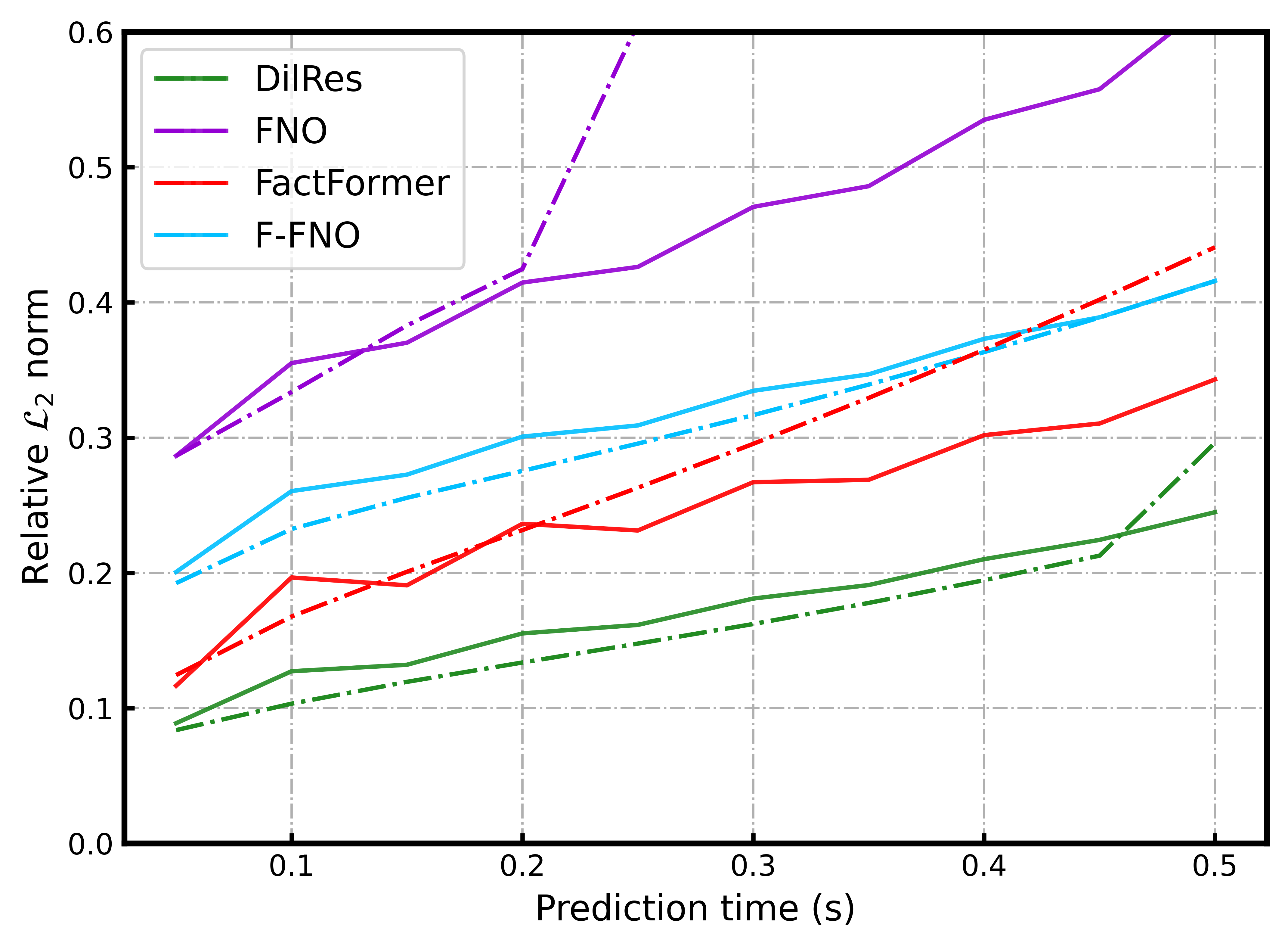}
        \captionsetup{justification=centering}
        \captionof{figure}{Error trend of pressure $p$ on 3D isotropic turbulence.}
        \label{fig:prs err 3d iso}
\end{minipage}
\begin{minipage}{0.49\textwidth}
    \vspace{-2mm}
         \includegraphics[width=1.0\textwidth]{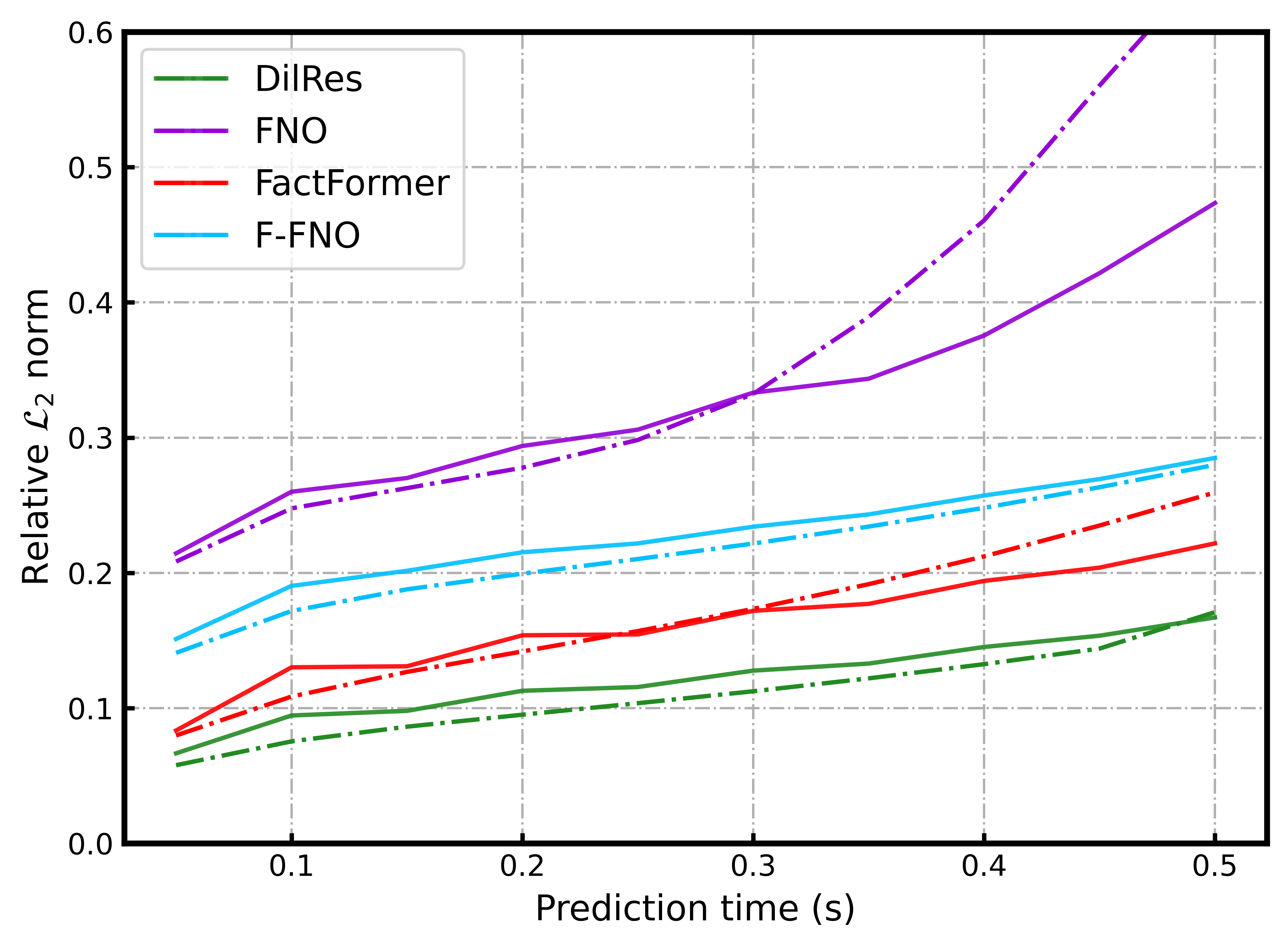}
        \captionsetup{justification=centering}
        \captionof{figure}{Error trend of velocity $\mathbf{u}$ on 3D isotropic turbulence.}
        \label{fig:vel err 3d iso}
\end{minipage}

\begin{minipage}{0.5\textwidth}
    \centering
    \includegraphics[width=1.0\textwidth]{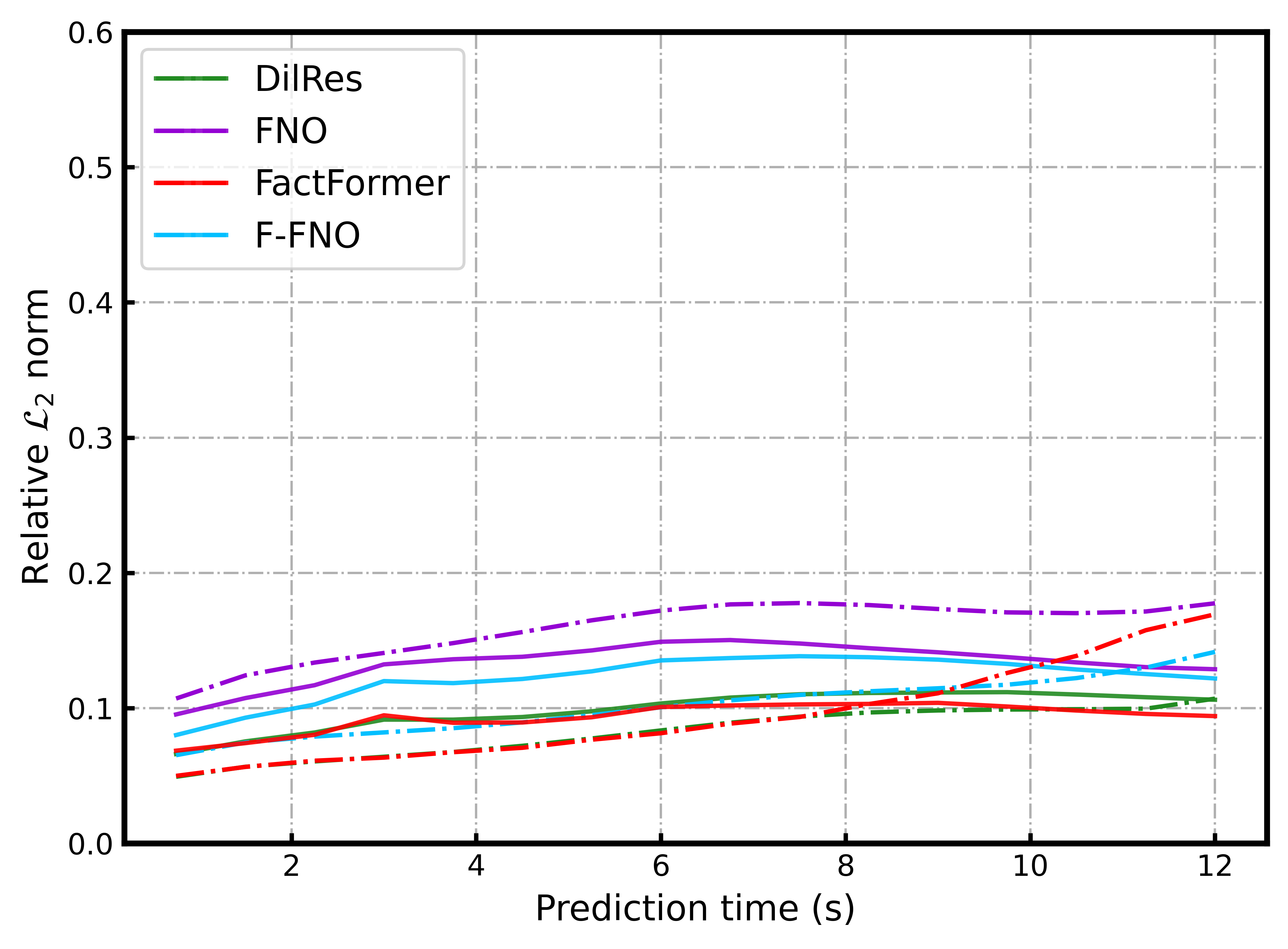}
    \vspace{-2mm}
    \captionsetup{width=0.95\linewidth, justification=centering}
    \captionof{figure}{Error trend of marker field $d$ \\ on 3D smoke buoyancy.}
    \label{fig:dns err 3d smoke} 
\end{minipage}
\begin{minipage}{0.49\textwidth}
\vspace{-1mm}
    \centering
    \includegraphics[width=1.0\textwidth]{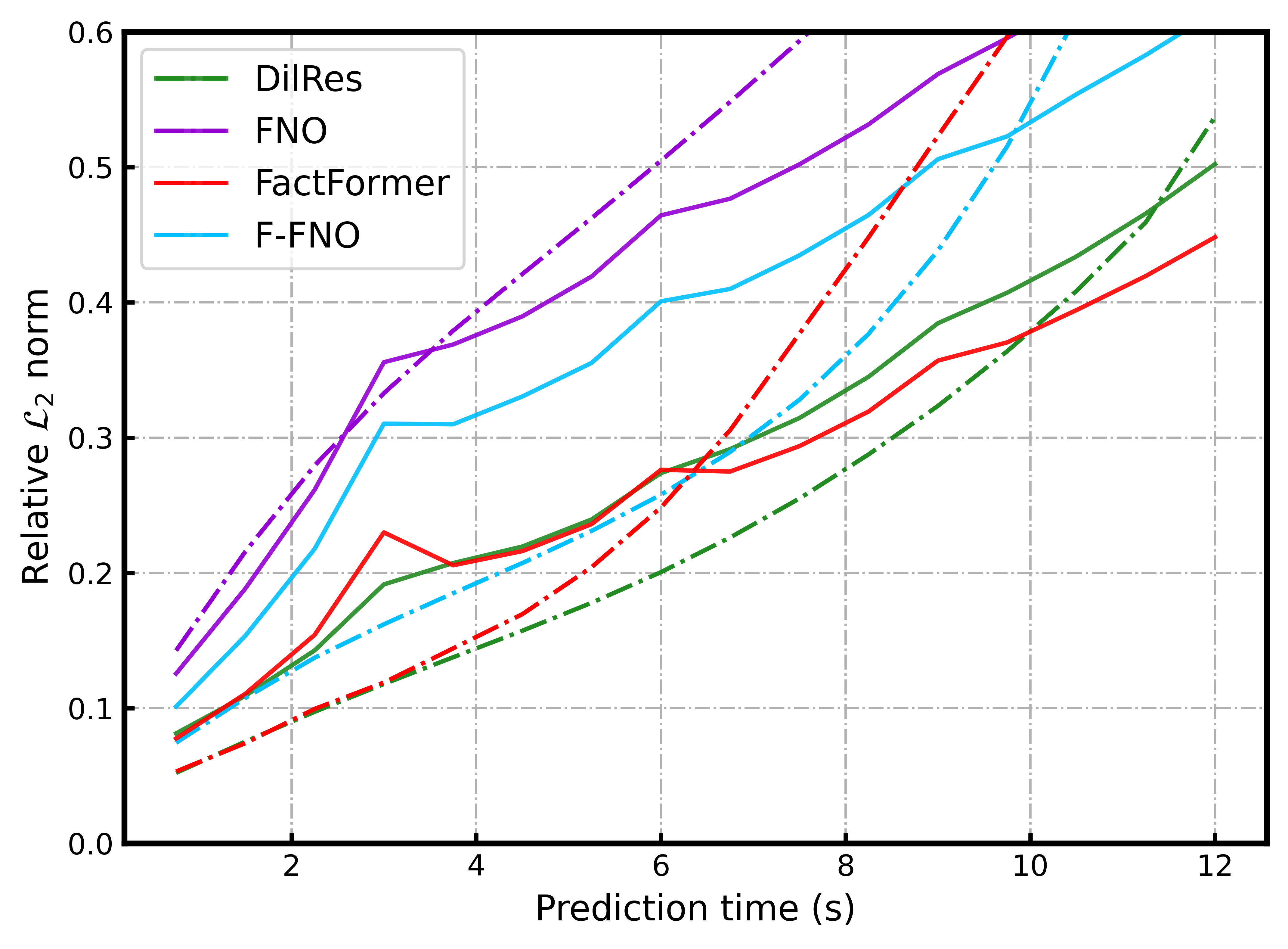}
    \vspace{-2mm}
    \captionsetup{width=0.95\linewidth, justification=centering}
    \captionof{figure}{Error trend of velocity field $\mathbf{u}$ on 3D smoke buoyancy.}
    \label{fig:vel err 3d smoke}
       
\end{minipage}

\section{Further ablation study}
\label{sec:further ablation study}
This section includes further study and numerical experiments on the proposed model.

\paragraph{Note on backpropagation}
Consider a single (softmax-free) attention block that consists of a attention layer and a two layer feedforward network (for simplicity we consider single-headed case):
\begin{equation}
    U_1 = \text{Att}(U_0), \quad
    U_2 = \sigma (U_1 W_1)W_2,
\end{equation}
where $U_0 \in \mathbb{R}^{N \times d}$ is the input, $W_1, W_2 \in \mathbb{R}^{d\times d}$ are learnable weights for the feedforward network. Given a loss function $l(\cdot):\mathbb{R}^{N \times d}\mapsto \mathbb{R}$ (for example, mean squared error), define $\Tilde{U_1}:= U_1 W_1$, $\frac{\partial l}{\partial U_2}:=h$, $\frac{\partial \sigma(\Tilde{U_1})}{\partial \Tilde{U_1}}:=g$, the gradient for the weights and input in the feedforward network are ($\odot$ denotes element-wise multiplication):
\begin{align}
    \frac{\partial l}{\partial W_2} &= \sigma(\Tilde{U_1})^Th \\
    \frac{\partial l}{\partial W_1} &= U_1^T (h W_2^T \odot g) \\
    \frac{\partial l}{\partial U_1} &= (h W_2^T \odot g) W_1^T
\end{align}

For linear (softmax-free) dot-product attention: $U_1=\text{Att}(U_0)=QK^TV$, where $Q=U_0W_q, K=U_0W_k, V=U_0W_v$, and thus {\small$U_1=U_0 W_q W_k^T U_0^T U_0 W_v$}, the gradient of weights are:
\begin{align}
\frac{\partial l}{\partial W_q} &= U_0^T \frac{\partial l}{\partial U_1} W_v^T U_0^T U_0 W_k, \\
\frac{\partial l}{\partial W_k} &= U_0^T U_0 W_v (\frac{\partial l}{\partial U_1})^T U_0 W_q, \\
\frac{\partial l}{\partial W_v} &= U_0^T  U_0 W_k W_q^T U_0^T \frac{\partial l}{\partial U_1},
\end{align}
where $U_0 \in \mathbb{R}^{N \times d}$ has appeared three times in each calculation and thus in backpropagation the computational cost of attention layer is generally more expensive than the feedforward layer where it involves more matrices that grow exponentially with respect to the spatial resolution.

Next we provide a comparison between the backpropagation of linear attention and the proposed factorized attention in scalar summation form. For simplicity, we only consider the attention layer.
For linear attention, it can be written as: 
\begin{equation}
    Z_{i, c} = \sum_{m=1}^d Q_{i, m} (\sum_{j=1}^N K_{j, m} V_{j, c}),
\end{equation} 
where $Z_{i, c}$ denotes the $i$-th row and $j$-th column of matrix $Z$ and similar for other matrices, $Q=XW_q, K=XW_k, V=XW_v$ and $X\in \mathbb{R}^{N \times d}$ is input. The gradient of parameters are computed as:

\begin{align}
\frac{\partial Z_{i, c}}{\partial (W_q)_{r, s}} &= X_{i, r}(\sum_{j=1}^N K_{j, s} V_{j, c}), \\
\frac{\partial Z_{i, c}}{\partial (W_k)_{r, s}} &=  Q_{i, s}(\sum_{j=1}^N X_{j, r} V_{j, c}), \\
  \frac{\partial Z_{i, c}}{\partial (W_v)_{r, s}} &=  
     \begin{cases}
       \sum_{m=1}^d Q_{i, m}(\sum_{j=1}^N K_{j, m} X_{i, r})&: \text{if~} s=c \\
       0&: \text{otherwise.} 
     \end{cases}
\end{align}

For the proposed factorized attention, it can written as:
\begin{equation}
    Z_{i, c} = \sum_{j_1=1}^{S_1}\sum_{j_2=1}^{S_2}\hdots \sum_{j_n=1}^{S_n}A_{i_1, j_1}^{(1)}A_{i_2, j_2}^{(2)}\hdots A_{i_n, j_n}^{(n)} V_{j, c}~, \quad j:=(j_1, j_2, \hdots, j_n),~ i:=i_1, i_2, \hdots, i_n
\end{equation}
where $A^{(m)}=Q^{(n)} (K^{(m)})^T $ ($A^{(m)} \in \mathbb{R}^{S_m \times S_m}, N=S_1\times S_2 \times \hdots \times S_n$) is the axial kernel matrix as defined in \eqref{eq:sub-kernel calculation}, $X^{(m)} \in \mathbb{R}^{S_m \times d}$ is the axial projection along $m$-th axis as defined in \eqref{eq:learnable proj}.

For $W_v$, its gradient is computed as:
\begin{equation}
    \frac{\partial Z_{i, c}}{\partial (W_v)_{r, s}} = 
         \begin{cases}
            \sum_{j_1=1}^{S_1}\sum_{j_2=1}^{S_2}\hdots \sum_{j_n=1}^{S_n}A_{i_1, j_1}^{(1)}A_{i_2, j_2}^{(2)}\hdots A_{i_n, j_n}^{(n)} X_{j, r} &: \text{if~} s=c \\
            0 &: \text{otherwise.} 
     \end{cases}
\end{equation}

For $W^{(n)}_q, W^{(n)}_k$, their gradients are computed as:
\begin{align}
\frac{\partial Z_{i, c}}{\partial (W_q^{(n)})_{r, s}} &= X^{(n)}_{i_n, r}\sum_{j_1=1}^{S_1}\sum_{j_2=1}^{S_2}\hdots \sum_{j_n=1}^{S_n}A_{i_1, j_1}^{(1)}A_{i_2, j_2}^{(2)}\hdots A_{i_{n-1}, j_{n-1}}^{(n-1)} V_{j, c} K^{(n)}_{j_n, s} , \\
\frac{\partial Z_{i, c}}{\partial (W^{(n)}_k)_{r, s}} &=  Q^{(n)}_{i_n, s}\sum_{j_1=1}^{S_1}\sum_{j_2=1}^{S_2}\hdots \sum_{j_n=1}^{S_n}A_{i_1, j_1}^{(1)}A_{i_2, j_2}^{(2)}\hdots A_{i_{n-1}, j_{n-1}}^{(n-1)} V_{j, c} X^{(n)}_{j_n, r},
\end{align}

The major difference is that for factorized attention the summation is taken over each axis separately while for linear attention is taken over all $N$ grid points.

\begin{figure}[h]
\vspace{-2mm}
    \begin{subfigure}{0.5\linewidth}
    \centering
    \includegraphics[width=0.9\textwidth]{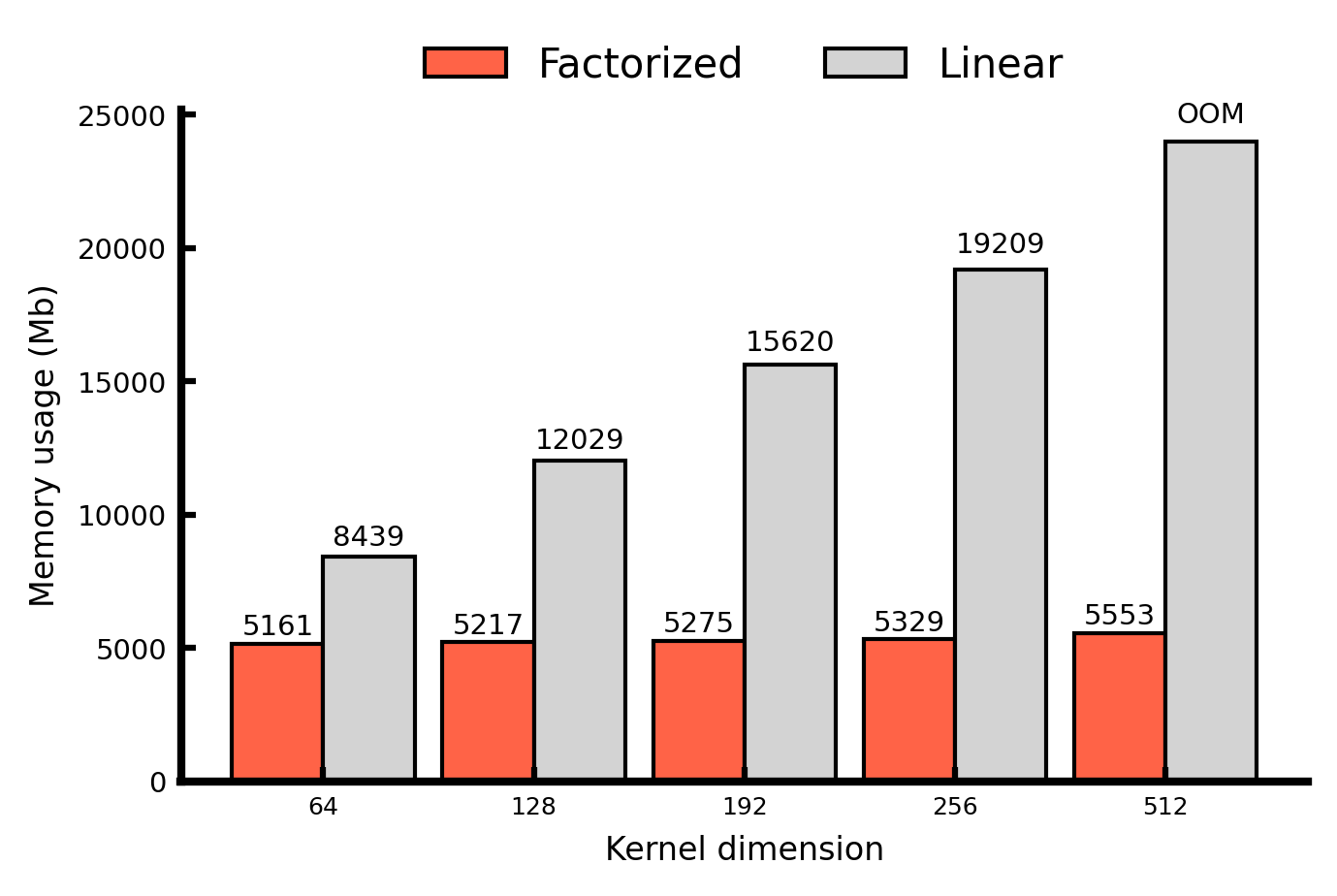}
    \caption{Forward + backward peak memory usage. \label{fig:fwd-bwd memory}
}
\end{subfigure}
\begin{subfigure}{0.49\linewidth}
    \centering
    \includegraphics[width=0.9\textwidth]{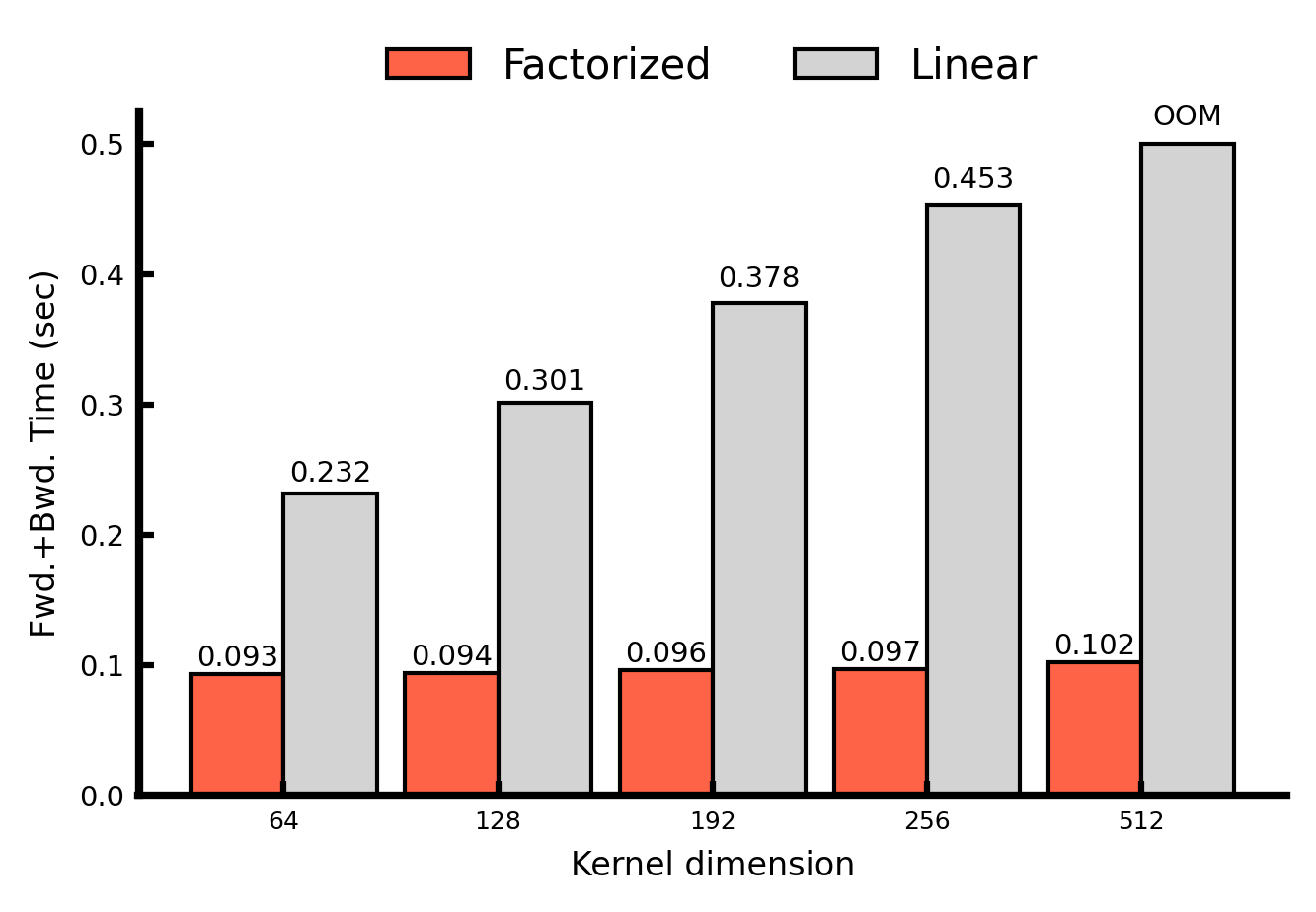}
    \caption{Forward + backward time per iteration.    \label{fig:fwd-bwd time}
}
\end{subfigure}
\vspace{-1mm}
\caption{\small Benchmark of factorized attention and linear attention on 2D domain (with $128\times128$ grid) with varying kernel dimension. Benchmark is done on an RTX 3090 with PyTorch 1.8.2 and a batch size of 4. Hyperparameter setting is the same as in Table \ref{tab: hyperparameter}-2D Kolmogorov flow. "OOM" denotes out of memory.}
\end{figure}

\begin{figure}[h]
\vspace{-2mm}
    \begin{subfigure}{0.5\linewidth}
    \centering
    \includegraphics[width=0.85\textwidth]{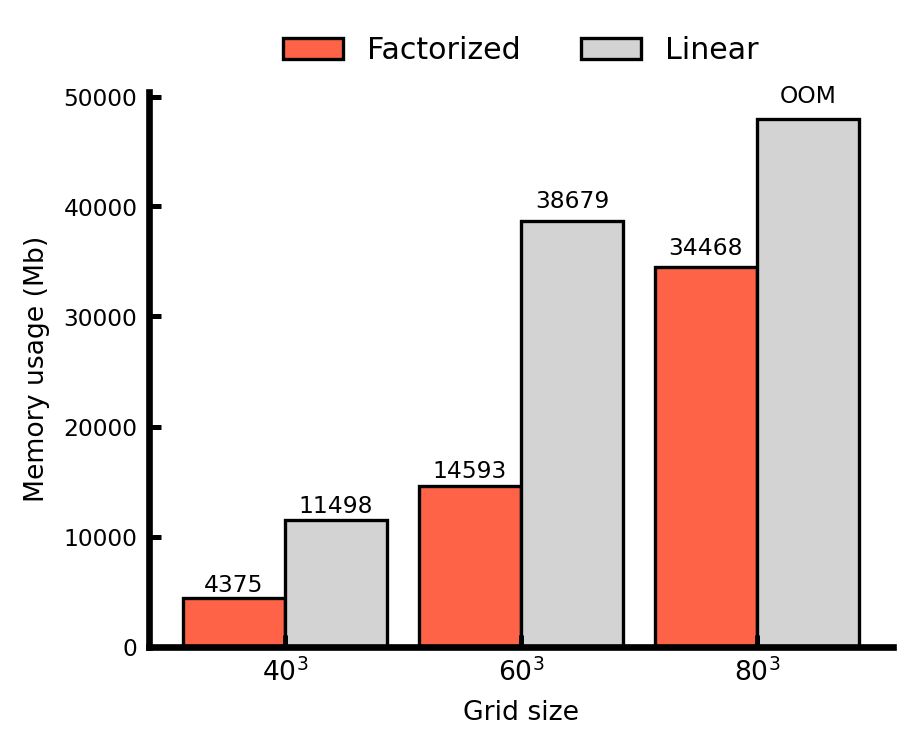}
    \caption{Forward + backward peak memory usage. \label{fig:3d fwd-bwd memory}
}
\end{subfigure}
\begin{subfigure}{0.49\linewidth}
    \centering
    \includegraphics[width=0.85\textwidth]{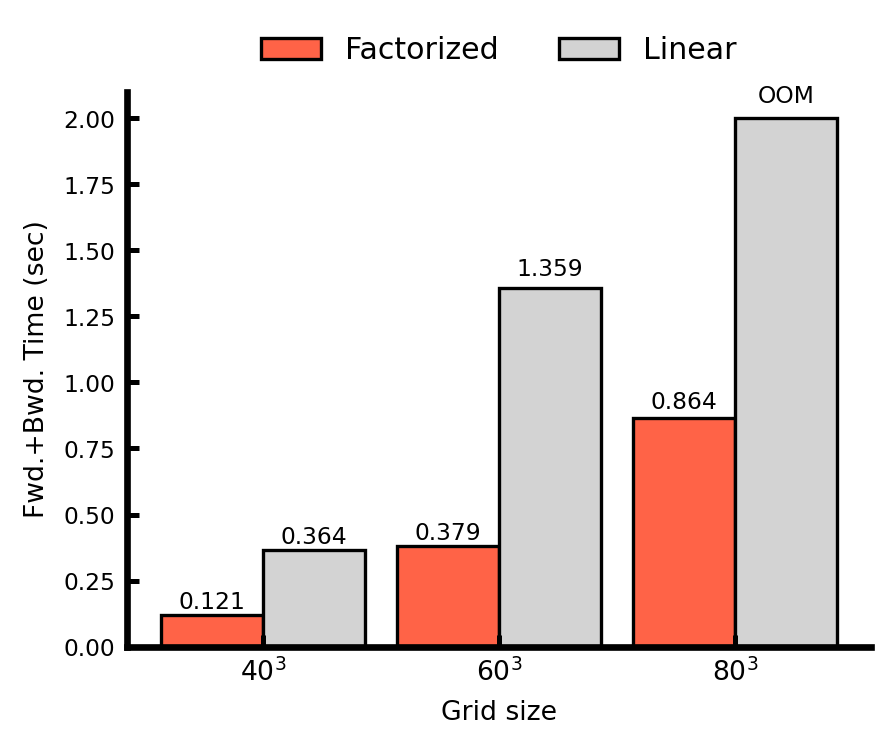}
    \caption{Forward + backward time per iteration.    \label{fig:3d fwd-bwd time}
}
\end{subfigure}
\vspace{-1mm}
\caption{\small Benchmark of factorized attention and linear attention on 3D domain with varying grid size. Benchmark is done on an A6000 with PyTorch 1.8.2 and batch size of 1. Hyperparameter setting is the same as in Table \ref{tab: hyperparameter}-3D isotropic turbulence.}
\end{figure}

\begin{figure}[h]
\vspace{-2mm}
    \centering
    \begin{subfigure}{0.5\linewidth}
    \centering
    \includegraphics[width=0.85\textwidth]{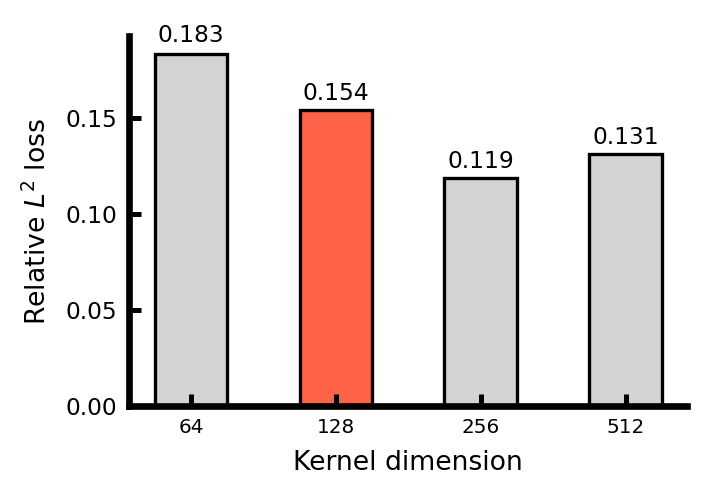}
    \vspace{-2mm}
    \caption{Ablation on the kernel dimension. \label{fig:ablation kernel dim}
}
\end{subfigure}
\begin{subfigure}{0.49\linewidth}
    \centering
    \includegraphics[width=0.85\textwidth]{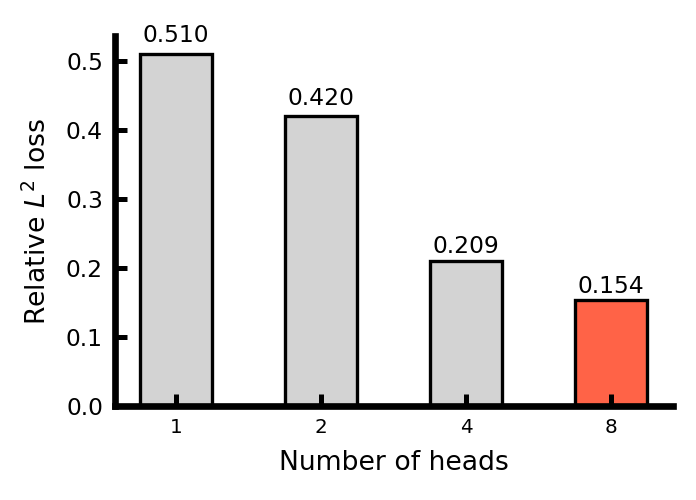}
    \vspace{-2mm}
    \caption{Ablation on the number of heads. \label{fig:ablation num heads}
}
\end{subfigure}
\vspace{-3mm}
\caption{\small Ablation study on the key hyperparameters. Red color denotes the final choice of hyperparameter. Experiments are carried out on the validation fold.}
\end{figure}

\paragraph{Runtime comparison}
As discussed in Section \ref{sec:implementation details}, the kernel dimension indicates how many function bases are used to evaluate the kernel and a larger kernel dimension is beneficial to the learning capacity of the model. As shown in Figure \ref{fig:fwd-bwd memory}, \ref{fig:fwd-bwd time}, linear attention's training cost increases more significantly than the factorized attention as kernel dimension increases, since its complexity is quadratic with respect to kernel dimension. Factorized attention's computational efficiency can be further improved by reducing the spatial resolution, leveraging techniques such as learning the mapping in the latent space (similar to latent diffusion model \citep{rombach2022highresolution}), multi-scale network architecture that resembles multigrid methods \citep{guo2022transformer, liu2023mitigating},  or domain decomposition \citep{quarteroni1999domain, jagtap2020extended}. Furthermore, the training cost of factorized attention is also relatively lower than linear attention on 3D domain as shown in Figure \ref{fig:3d fwd-bwd memory}, \ref{fig:3d fwd-bwd time}.
\vspace{-2mm}
\newpage
\paragraph{Model scaling performance}
We study the impact of the number of heads and size of kernel dimension on prediction loss. For each direction of hyperparameter search, we fix the value of other hyperparameters to that shown in Table \ref{tab: hyperparameter}. The ablation experiments are conducted on 2D Kolmogorov flow (sampled from a $128\times 128$ grid) with a splitting different from the Evaluation Section in the main body of the paper. As shown in Figure \ref{fig:ablation num heads}, the number of attention heads has a crucial impact on the final performance. The model's performance drops significantly when using fewer heads. This highlights the importance of multi-head mechanism in the factorized attention. In addition, we observe that the final accuracy of our model benefits from an increased kernel dimension (as shown in Figure \ref{fig:ablation kernel dim}).

\begin{figure}[h]
\vspace{-2mm}
\captionsetup{justification=centering}
    \begin{subfigure}{0.245\linewidth}
    \centering
    \includegraphics[width=1.03\textwidth]{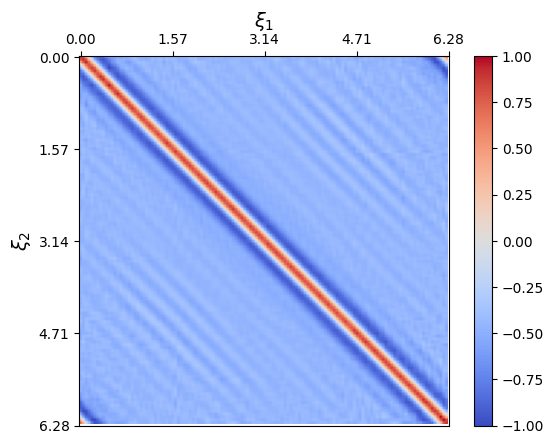}
    \caption{Layer 1, Head 1: \\kernel of $y$ axis \label{fig:kernel l1 k1}
}
\end{subfigure}
\begin{subfigure}{0.245\linewidth}
    \centering
    \includegraphics[width=1.03\textwidth]{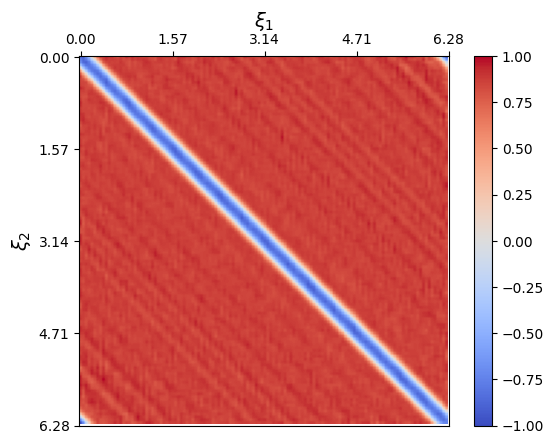}
    \caption{Layer 1, Head 1: \\kernel of $x$ axis    \label{fig:kernel l1 k2}
}
\end{subfigure}
    \begin{subfigure}{0.245\linewidth}
    \centering
    \includegraphics[width=1.03\textwidth]{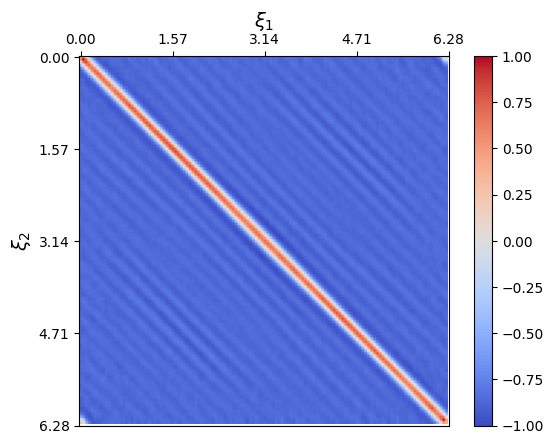}
    \caption{Layer 4, Head 1: \\kernel of $y$ axis \label{fig:kernel l4 k1}
}
\end{subfigure}
\begin{subfigure}{0.245\linewidth}
    \centering
    \includegraphics[width=1.03\textwidth]{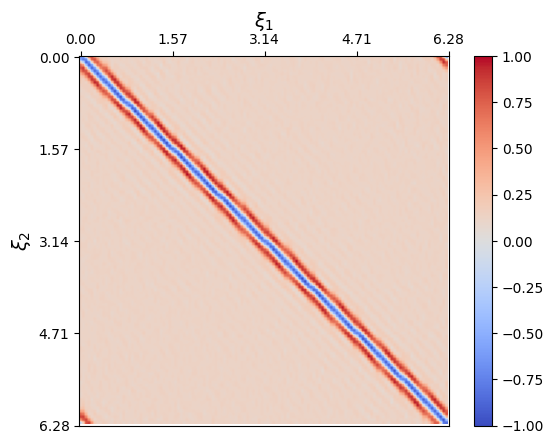}
    \caption{Layer 4, Head 1: \\kernel of $x$ axis    \label{fig:kernel l4 k2}
}
\end{subfigure}
\caption{Visualization of normalized attention kernel \label{fig:kernel vis}.}
\end{figure}

\paragraph{Visualization of learned kernels}
We visualize the learned kernel as shown in Figure \ref{fig:kernel vis}. Due to the presence of Rotary positional encoding \citep{su2022roformer}, all kernels have a stationary pattern (the kernel value $\kappa(\xi_1, \xi_2)$ depends only on the relative distance between two points, e.g. $L^2$ distance: $\left|\left|\xi_1- \xi_2\right|\right|_{2}$). The kernel matrices also exhibits symmetric pattern despite the non-symmetric nature of dot product $QK^T$ and all kernel matrices are diagonal dominated.

\begin{figure}[h]
    \centering
    \includegraphics[width=0.65\linewidth]{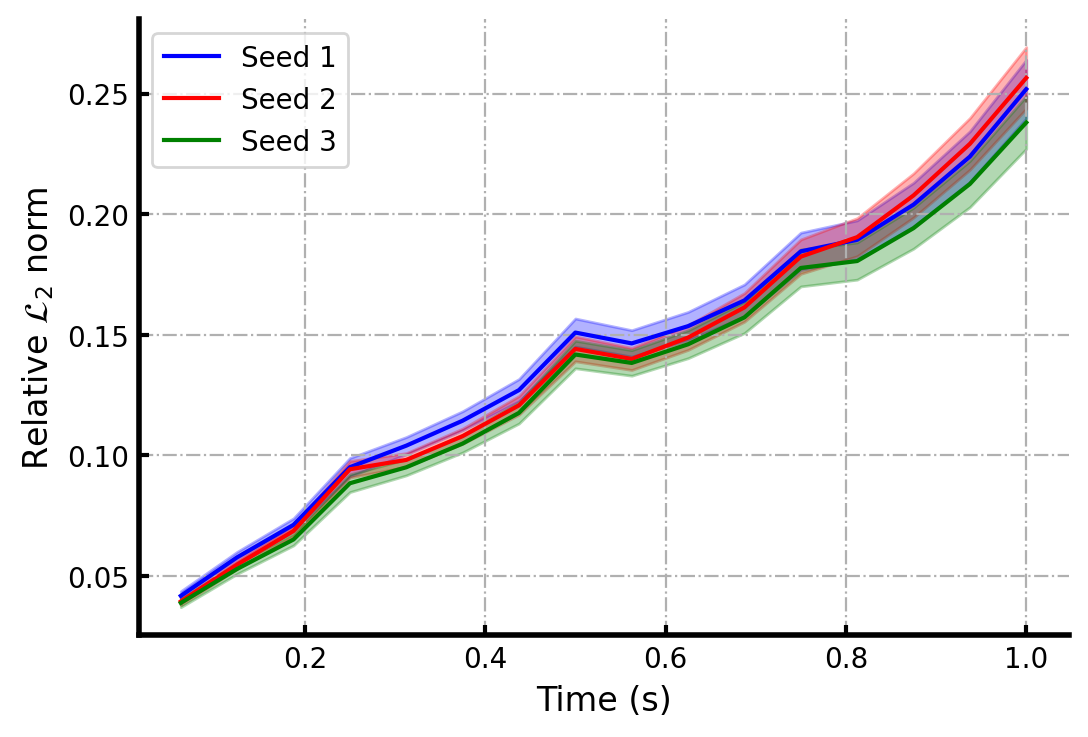}
    \vspace{-2mm}
    \caption{ Averaged frame-wise loss trends. Each loss curve corresponds to a model initialized under a specific seed.}
    \label{fig:seed influence}
\end{figure}

\paragraph{Influence of random seed}
We investigate the influence of random seeds by training the model with three different seeds. As shown in Figure \ref{fig:seed influence}, all models converge to the similar level of loss with marginal difference.

\vspace{-2mm}
\section{Dataset details}
\label{sec:dataset details}
In this section we provide the details for each dataset.
\paragraph{2D Kolmogorov flow} The incompressible Navier-Stokes equation under vorticity form reads as,
\begin{flalign}
 \frac{\partial \omega (\mathbf{x},t)}{\partial t} + \mathbf{u}(\mathbf{x},t) \cdot \nabla \omega (\mathbf{x},t) &= \frac{1}{\textit{Re}} \nabla^2 \omega (\mathbf{x},t) + f(\mathbf{x}) , && \mathbf{x} \in (0,2\pi)^2,  t \in (0,T], \notag\\
\nabla \cdot \mathbf{u}(\mathbf{x},t) &= 0, &&  \mathbf{x} \in (0,2\pi)^2, t \in [0,T], \\
\omega (\mathbf{x},0) &= \omega_0(\mathbf{x}), && \mathbf{x} \in (0,2\pi)^2, \notag
\label{eq:kmflow}   
\end{flalign}
where $\omega$ denotes vorticity, $\mathbf{u}$ denotes velocity, $\textit{Re}$ denotes the Reynolds number, $\mathbf{x}=(x_1, x_2)$ denotes the spatial coordinates and $f(\cdot)$ is the forcing term that is set to $f(\mathbf{x})=-n\cos{(n x_2)}-0.1\omega(\mathbf{x})$. The equation is periodic in all spatial directions. Compared to the cases discussed in \citet{li2023physicsinformed}, we set the forcing factor $n$ to 8 \citep{chandler2013kmflow, large-scale-kmflow} and introduce dragging force term $0.1\omega(\mathbf{x})$ as described in \citet{Kochkov2021mlcfd}. The initial condition $\omega_0$ is sampled from a prescribed Gaussian random field same as \citet{li2023physicsinformed}. The dataset consists of $100$ trajectories for training and $20$ trajectories for testing, with the length of each trajectory being $10$ seconds and $160$ frames.

We modify the pseudo-spectral solver (under Apache License 2.0) from \url{https://github.com/neuraloperator/physics_informed/blob/master/solver/kolmogorov_flow.py} to generate the data. The referenced direct numerical simulation is carried out with a spatial resolution of $2048\times 2048$ and a temporal resolution of $1e-4$.
\paragraph{3D isotropic turbulence} The incompressible Navier-Stokes equation for this problem is given as:
\begin{flalign}
\frac{\partial \mathbf{u} (\mathbf{x},t)}{\partial t} +\mathbf{u}(\mathbf{x},t) \cdot \nabla\mathbf{u}(\mathbf{x},t)  &= \nu \nabla^2 \mathbf{u} (\mathbf{x},t) -\frac{1}{\rho} \nabla p(\mathbf{x},t)+ \mathbf{f}(\mathbf{x}) , &&\mathbf{x} \in (0,2\pi)^3,  t \in (0,T], \notag \\
\nabla \cdot \mathbf{u}(\mathbf{x},t) &= 0, && \mathbf{x} \in (0,2\pi)^3, t \in [0,T], \\
\mathbf{u}(\mathbf{x},0) &= \mathbf{u}_0(\mathbf{x}), && \mathbf{x} \in (0,2\pi)^3 \notag,
\label{eq:3diso}
\end{flalign}
where $\mathbf{u}$ denotes velocity, $p$ denotes the pressure, $\nu$ is the viscosity parameter, $\mathbf{x}=[x_1, x_2, x_3]$ denotes the spatial coordinates and $f(\cdot)$ is the forcing term. The equation is periodic in all three spatial dimensions. The initialization of $\mathbf{u}$ and the forcing settings follow \citet{Rogallo1981NumericalEI} and \citet{3dturb-dns} respectively, with Taylor Reynolds number set to $84$ \citep{3dturb-dns}. The dataset consists of $1000$ trajectories for training and $100$ trajectories for testing, with the length of each trajectory being $1$ second and $20$ frames.

We use the spectral Galerkin solver (under GNU GPL license 3.0) from \url{https://github.com/spectralDNS}. The referenced  simulation is carried out with a spatial resolution of $60\times60\times60$ and a temporal resolution of $0.005s$. 
\paragraph{3D smoke buoyancy}
The governing equations for the 3D smoke buoyancy problem are incompressible Navier-Stokes equation (similar as above) coupled with advection equation:
\begin{equation}
    \begin{aligned}
\frac{\partial \mathbf{u} (\mathbf{x},t)}{\partial t} +\mathbf{u}(\mathbf{x},t) \cdot \nabla\mathbf{u}(\mathbf{x},t)  &= \nu \nabla^2 \mathbf{u} (\mathbf{x},t) -\frac{1}{\rho} \nabla p(\mathbf{x},t)+ \mathbf{f}(\mathbf{x}, t) , &&\mathbf{x} \in (0,L)^3,  t \in (0,T], \notag \\
\frac{\partial d(\mathbf{x}, t)}{\partial t} +\mathbf{u}(\mathbf{x}, t) \cdot \nabla d(\mathbf{x}, t) &= 0, && \mathbf{x} \in (0,L)^3, t \in (0,T], \\
\nabla \cdot \mathbf{u}(\mathbf{x},t) &= 0, && \mathbf{x} \in (0,L)^3, t \in [0,T], \\
\mathbf{u}(\mathbf{x},0) = 0, \quad \mathbf{d}(\mathbf{x},0) &= d_0(\mathbf{x}), && \mathbf{x} \in (0,L)^3 \notag,
\end{aligned}
\label{eq:3dsmoke}
\end{equation}
where $\mathbf{f}(\mathbf{x}, t)=\left[0, 0, \eta d(\mathbf{x},t)\right]$, $\eta$ is the buoyancy factor, the velocity field $\mathbf{u}$ has a Dirichlet boundary condition: $\mathbf{u}(\mathbf{x}, \cdot)=0, \forall \mathbf{x} \in \partial \Omega$, and the scalar density field for smoke has a Neumann boundary condition: $\nabla d(\mathbf{x}, \cdot)=0, \forall \mathbf{x} \in \partial \Omega$. The initial condition $d_0(\mathbf{x})$ is a random field \footnote{Implemented with \textit{phiflow}'s Noise class, see:\url{https://tum-pbs.github.io/PhiFlow/phi/field/}} with scaling of Fourier coefficient set to $15.0$, smoothness factor set to $4.0$. The length of the rectangular domain $L$ is set to $8$. The dataset consists of $2000$ trajectories for training and $200$ trajectories for testing, with the length of each trajectory being $15$ seconds and $20$ frames.

We modify the 2D solver (under MIT license) from \url{https://github.com/microsoft/pdearena} to generate the data. The solver applies an advection-project scheme. The referenced  simulation is carried out with a spatial resolution of $64\times64\times64$ and a temporal resolution of $0.75s$. 

\paragraph{2D Darcy flow} The equation for the 2D Darcy flow is defined as:
\begin{equation}
\begin{aligned}
- \nabla \cdot (a(x) \nabla u(x)) = f(x), \quad & x \in (0,1)^2, \\
u_0(x) = 0, \quad & x \in \partial (0,1)^2,
\end{aligned}
\end{equation}
where $f(x)$ is the forcing function that is set to constant $1$. The coefficient function $a(x)$ is sampled from Gaussian Random Field with zero Neumann boundary condition. The data is generated via second-order finite 
difference solver on a $421 \times 421$ resolution grid. We use the pre-generated dataset from \citet{li2020fourier} (under MIT license). The dataset consists of $1000$ samples for training and $100$ samples for testing.
\newpage
\section{Results visualization}
\label{sec:results vis}
In this section, we provide exemplary visualization of the model's prediction. For 3D problems, the cross-section at the middle of the first axis is shown.

\begin{figure}[H]
    \centering
    \vspace{-2mm}
    \includegraphics[width=0.86\textwidth]{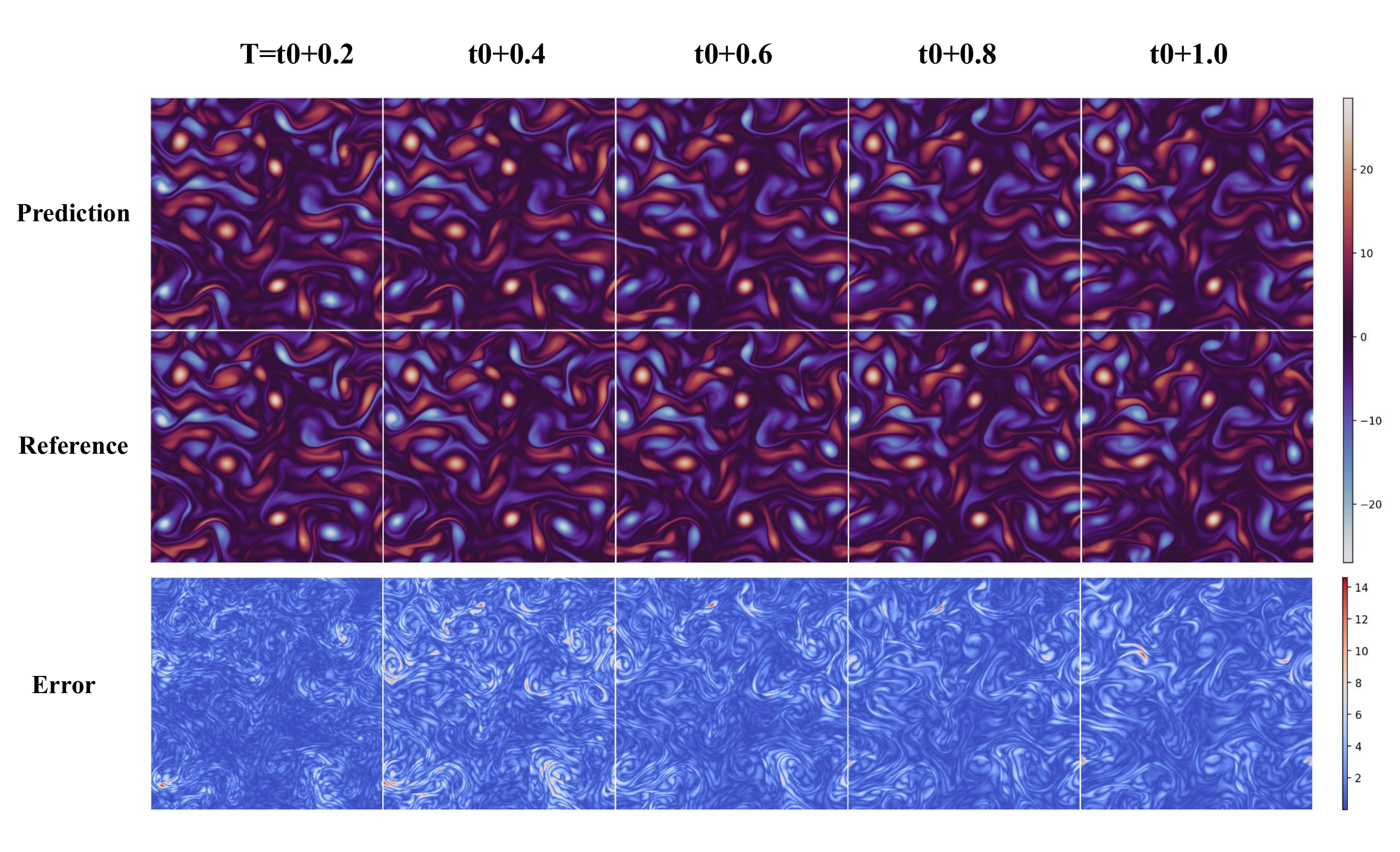}
    \vspace{-2mm}
    \caption{Sample 1 of 2D Kolmogorov flow.}
\end{figure}
\begin{figure}[H]
    \centering
    \vspace{-4mm}
    \includegraphics[width=0.86\textwidth]{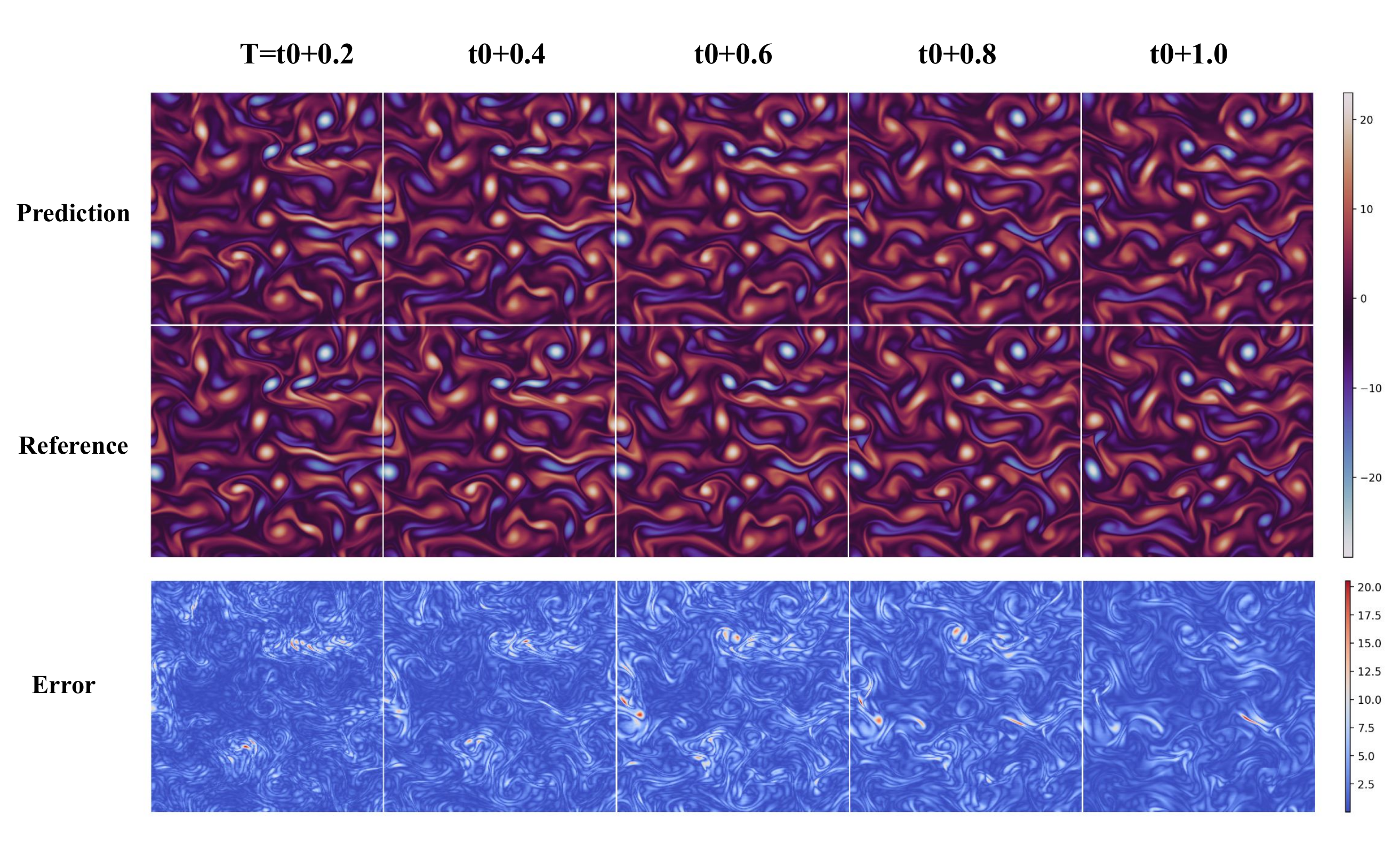}
        \vspace{-4mm}
    \caption{Sample 2 of 2D Kolmogorov flow.}
\end{figure}%
\begin{figure}[H]
    \centering
    \vspace{-4mm}
    \includegraphics[width=0.86\textwidth]{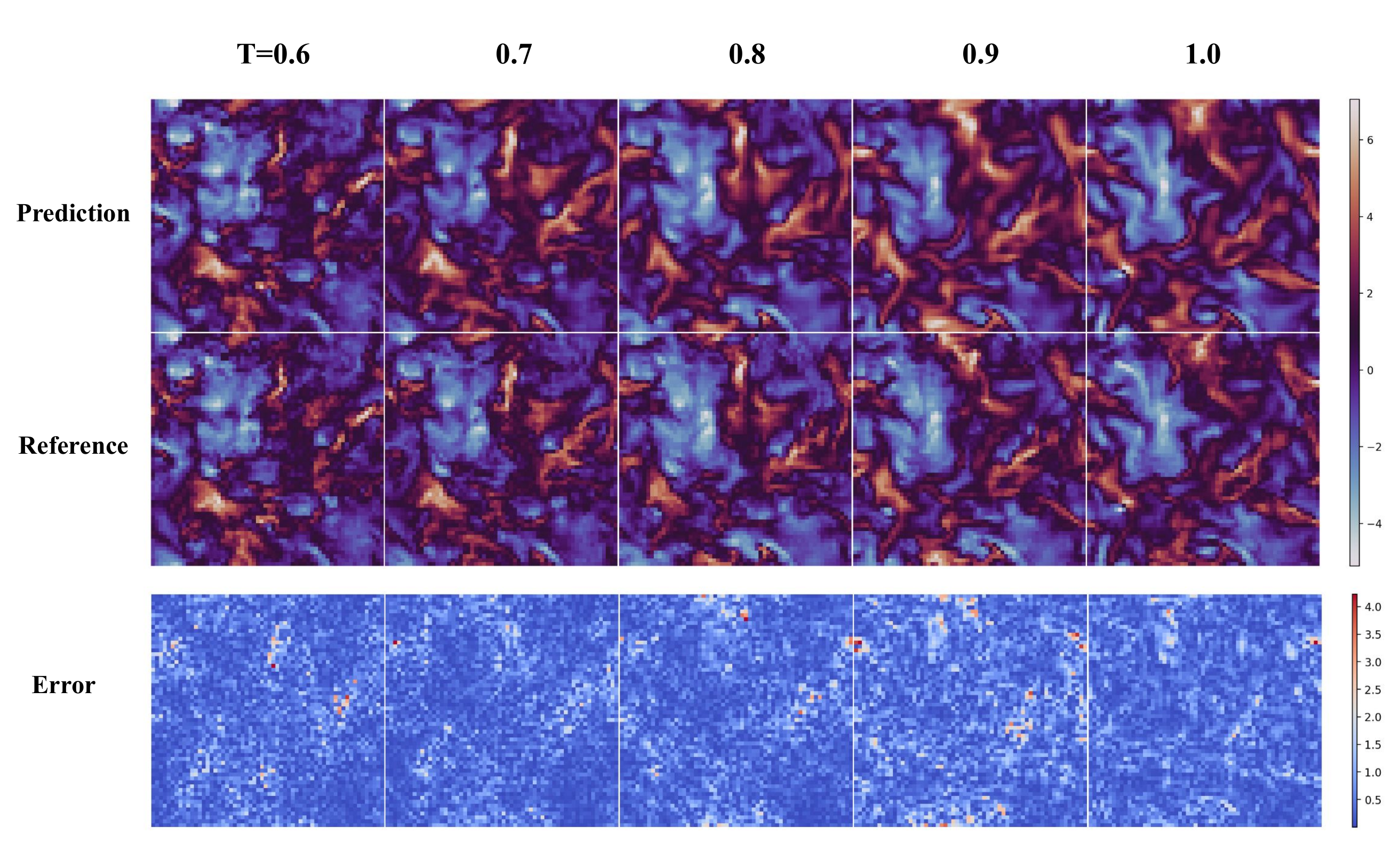}
        \vspace{-3mm}
    \caption{Pressure in 3D isotropic turbulence.}
\end{figure}%
\begin{figure}[H]
    \centering
        \vspace{-4mm}
    \includegraphics[width=0.86\textwidth]{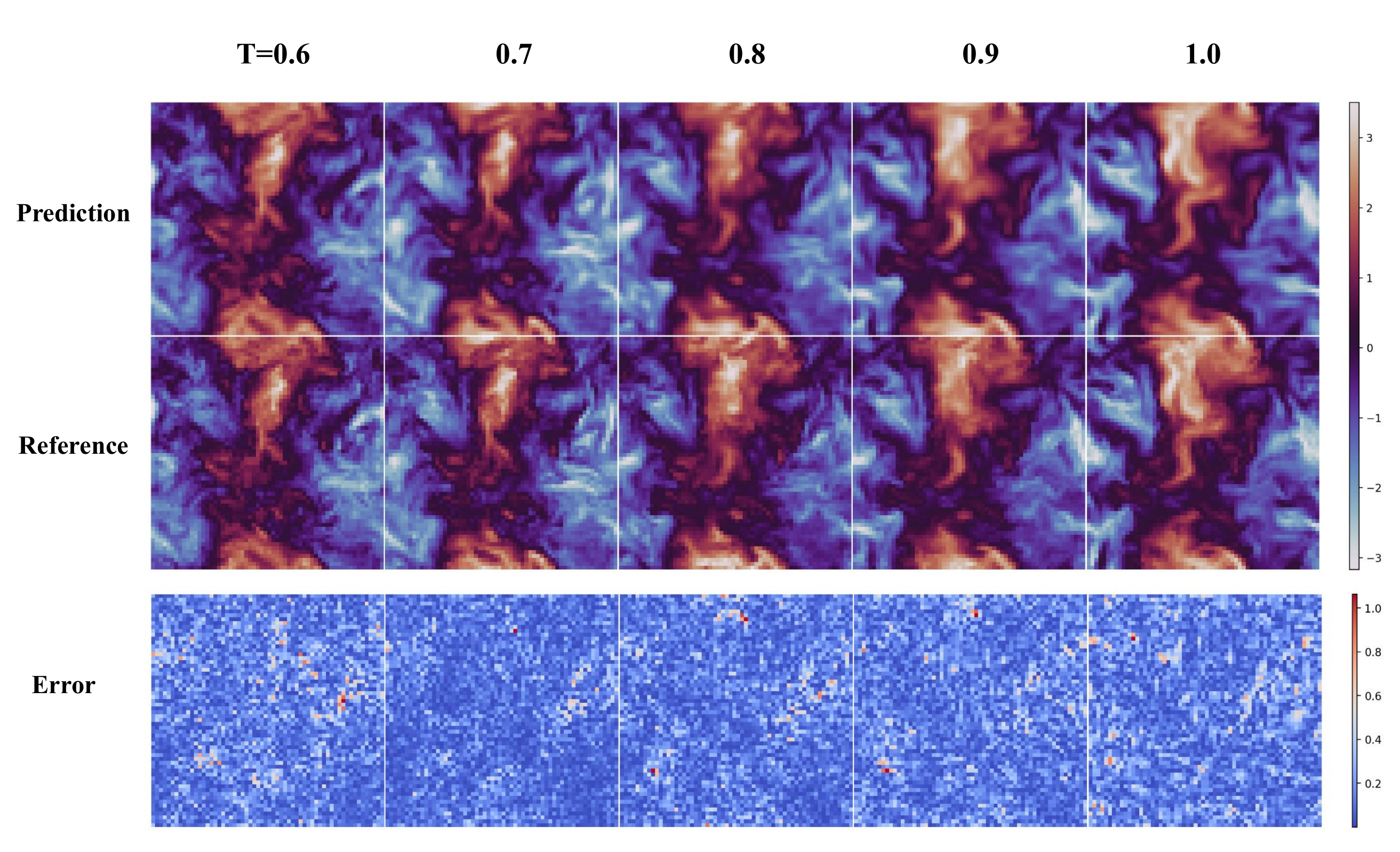}
        \vspace{-3mm}
    \caption{$x$-component of velocity in 3D isotropic turbulence.}
    \vspace{-3mm}
\end{figure}
\begin{figure}[H]
    \centering
    \vspace{-4mm}
    \includegraphics[width=0.86\textwidth]{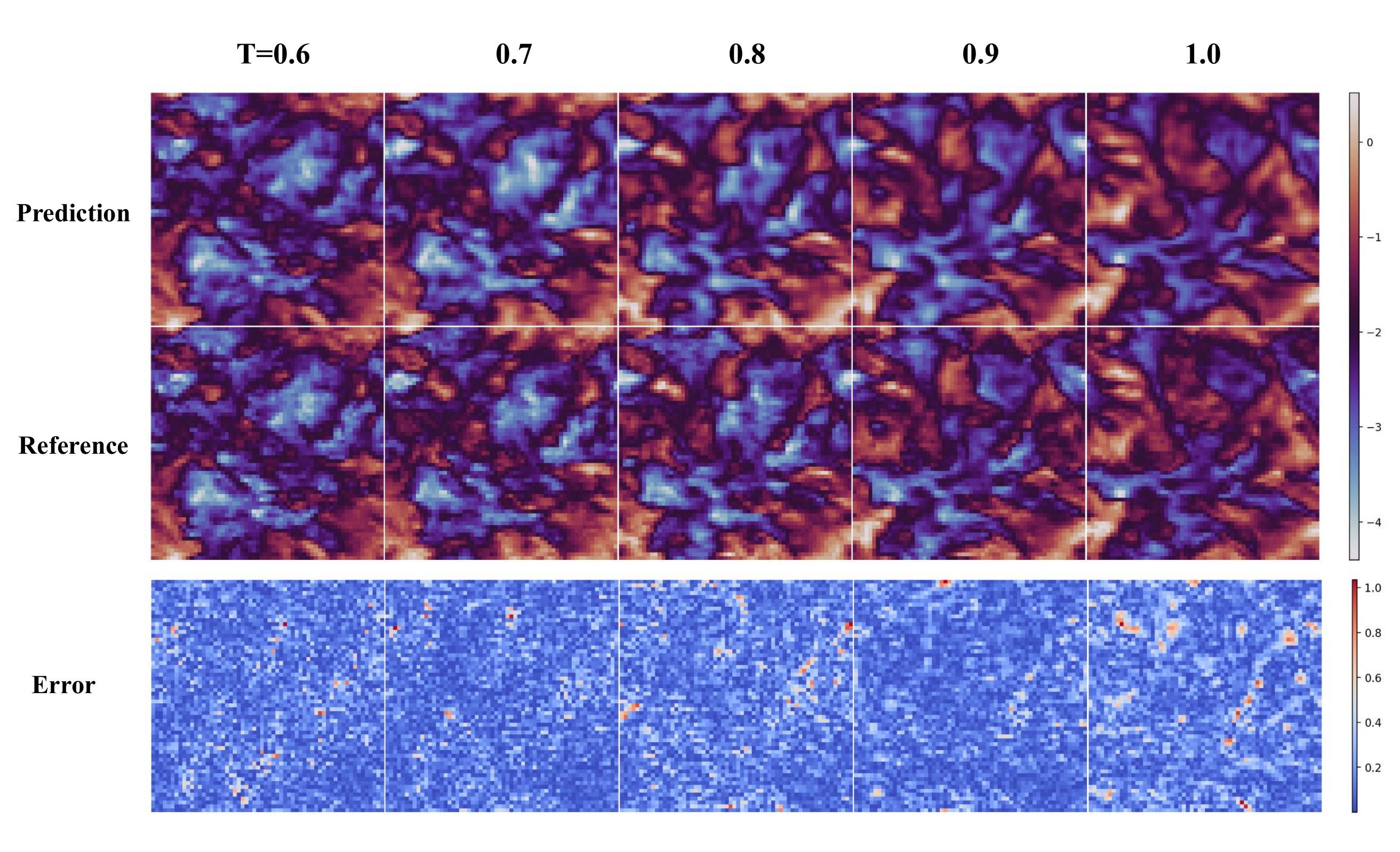}
        \vspace{-3mm}
    \caption{$y$-component of velocity in 3D isotropic turbulence.}
        \vspace{-3mm}
\end{figure}
\begin{figure}[H]
    \centering
    \vspace{-4mm}
    \includegraphics[width=0.86\textwidth]{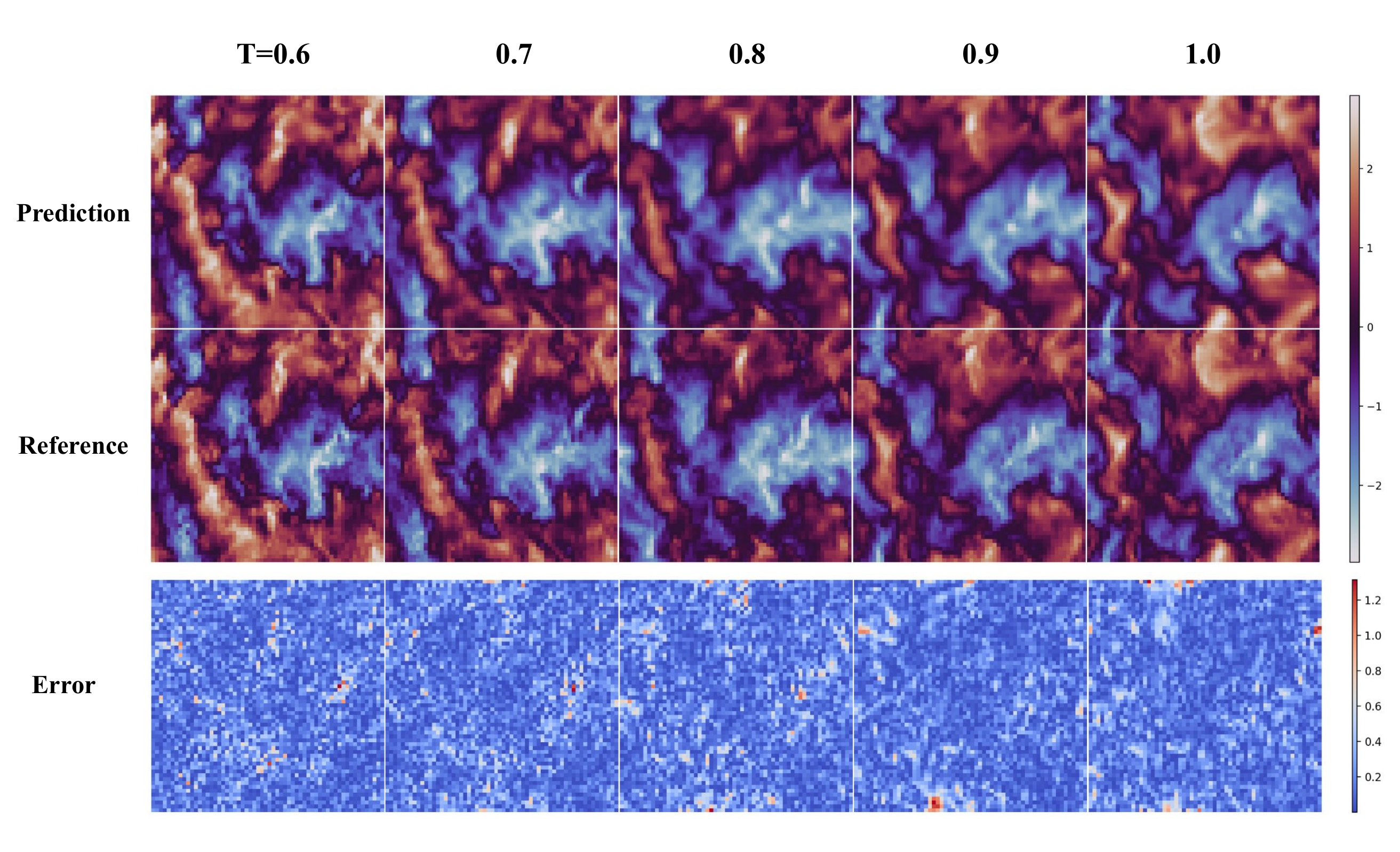}
         \vspace{-3mm}
    \caption{$z$-component of velocity in 3D isotropic turbulence.}
\end{figure}
\begin{figure}[H]
    \centering
    \vspace{-4mm}
    \includegraphics[width=0.86\textwidth]{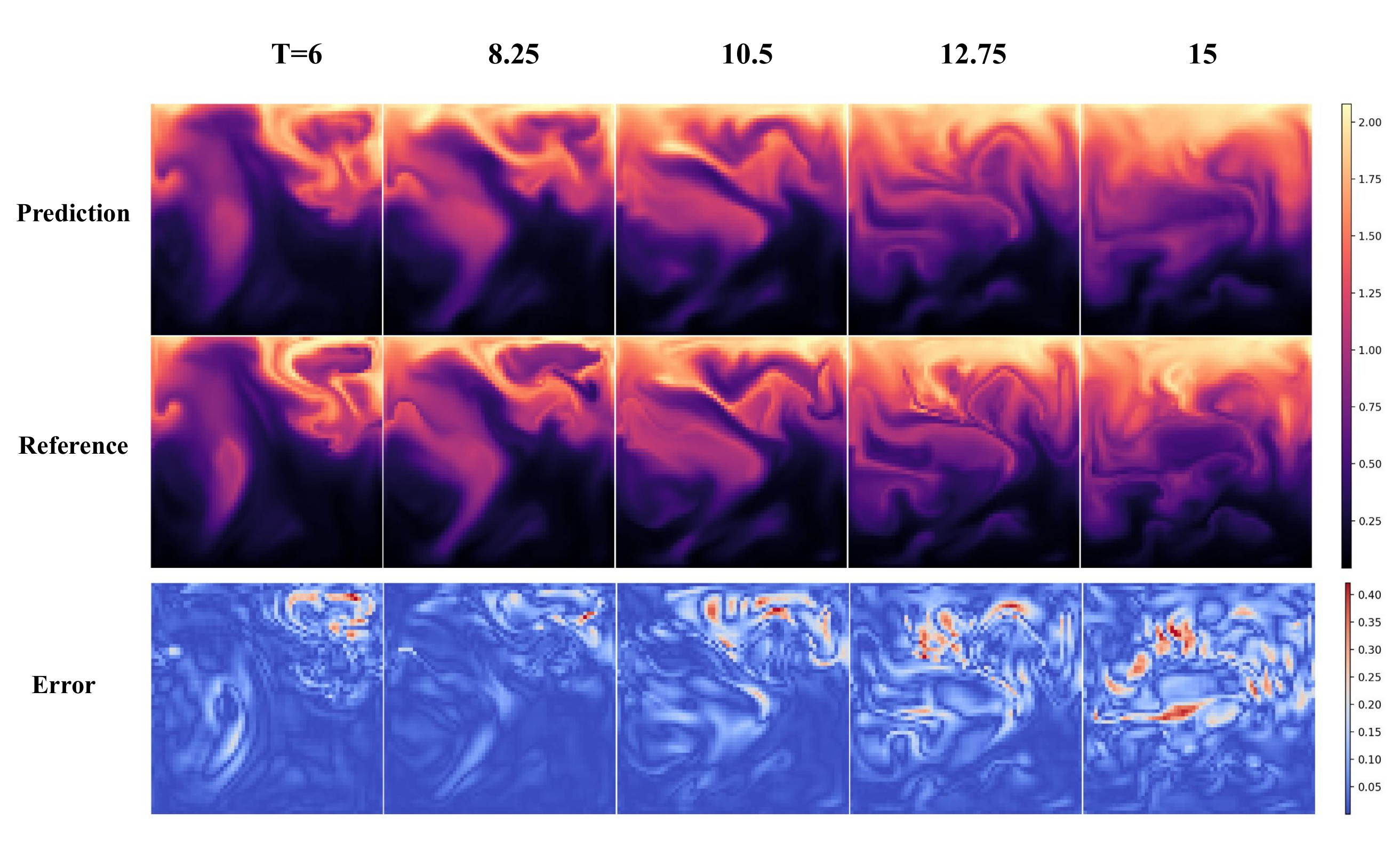}
            \vspace{-3mm}
    \caption{Smoke marker field in 3D smoke buoyancy.}
        \vspace{-4mm}
\end{figure}
\begin{figure}[H]
    \centering
    \vspace{-4mm}
    \includegraphics[width=0.86\textwidth]{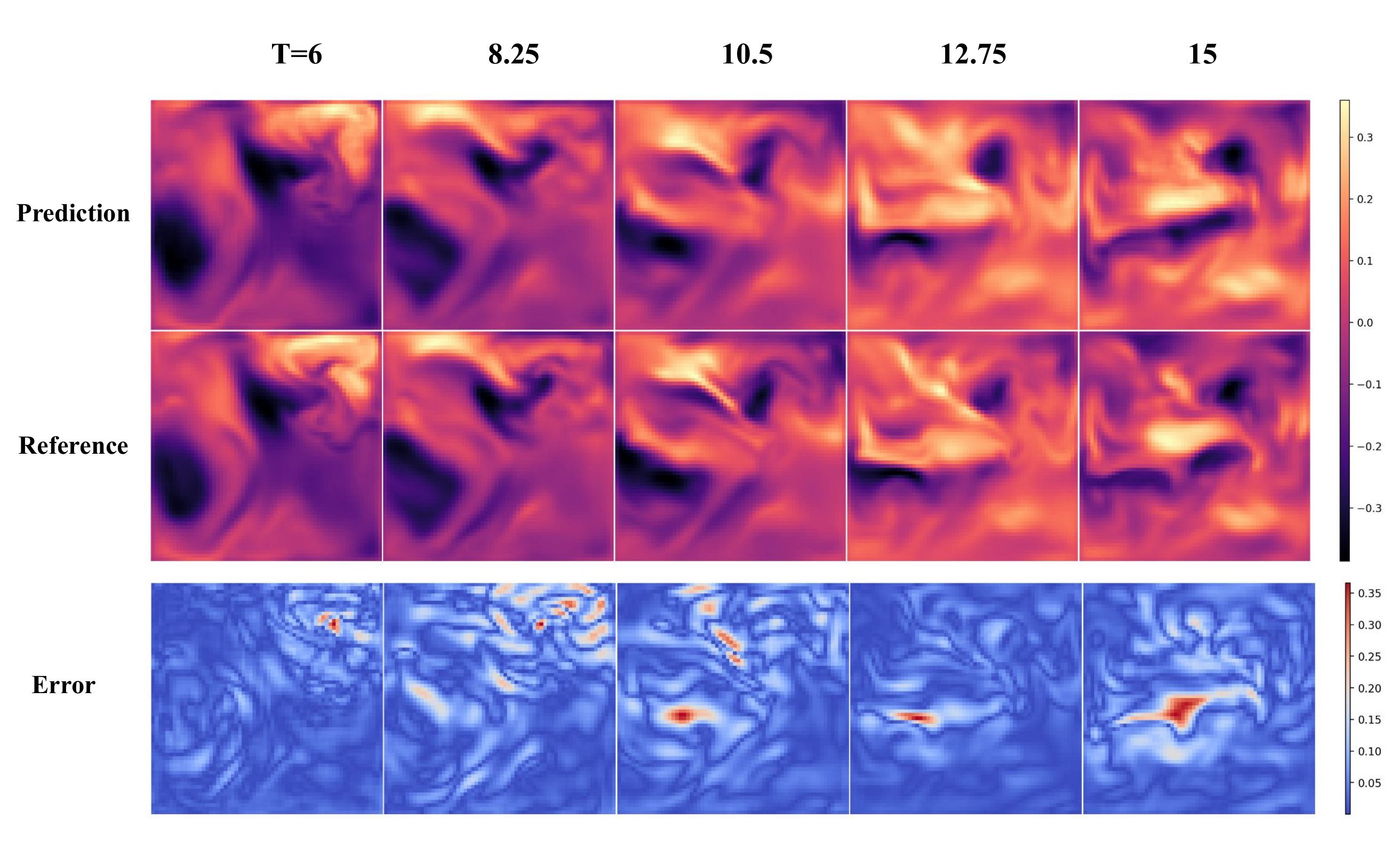}
    \vspace{-3mm}
    \caption{$x$-component of velocity in 3D smoke buoyancy.}
\end{figure}
\begin{figure}[H]
    \centering
    \vspace{-4mm}
    \includegraphics[width=0.86\textwidth]{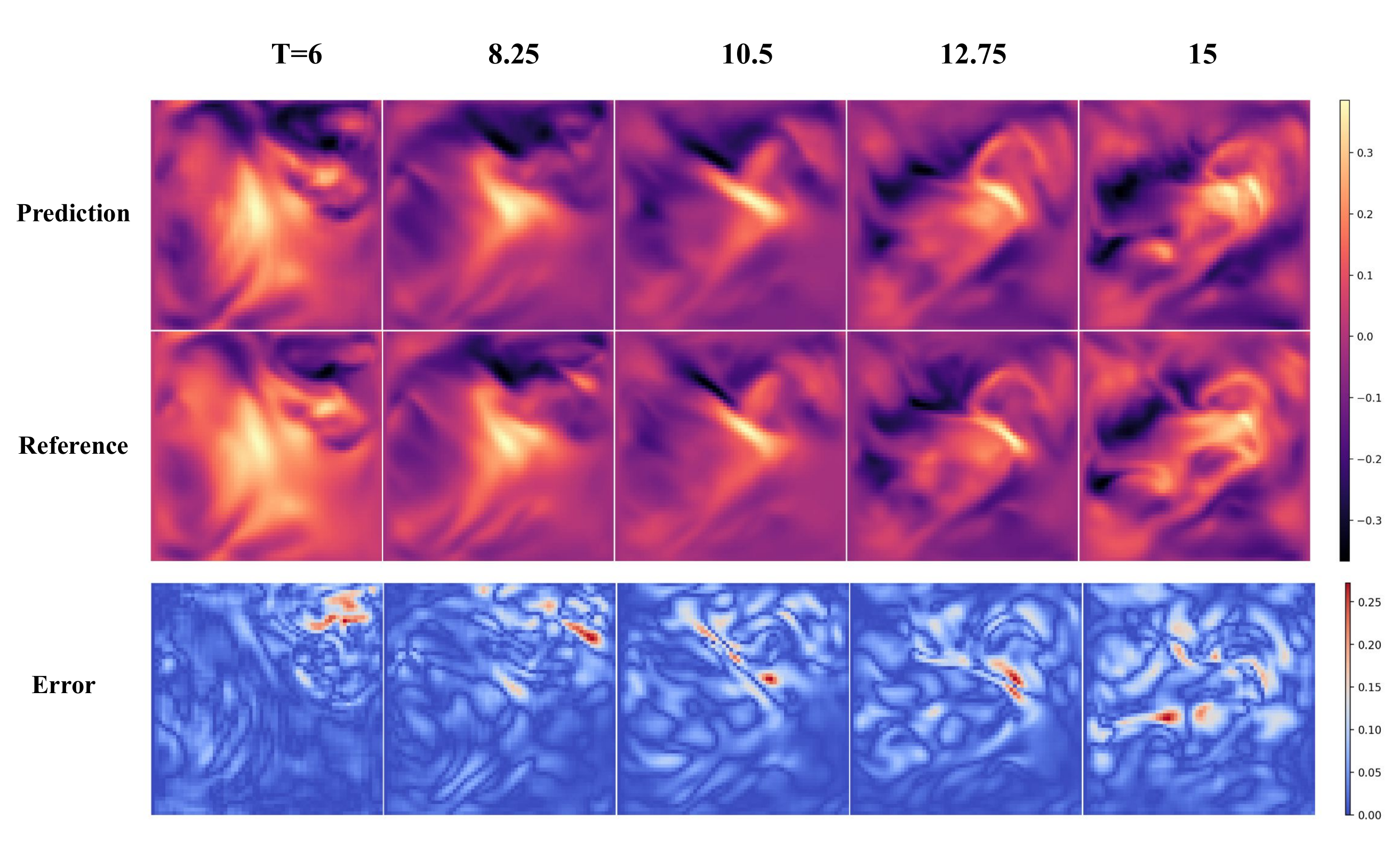}
        \vspace{-3mm}
    \caption{$y$-component of velocity in 3D smoke buoyancy.}
\end{figure}
\begin{figure}[H]
    \centering
    \includegraphics[width=0.86\textwidth]{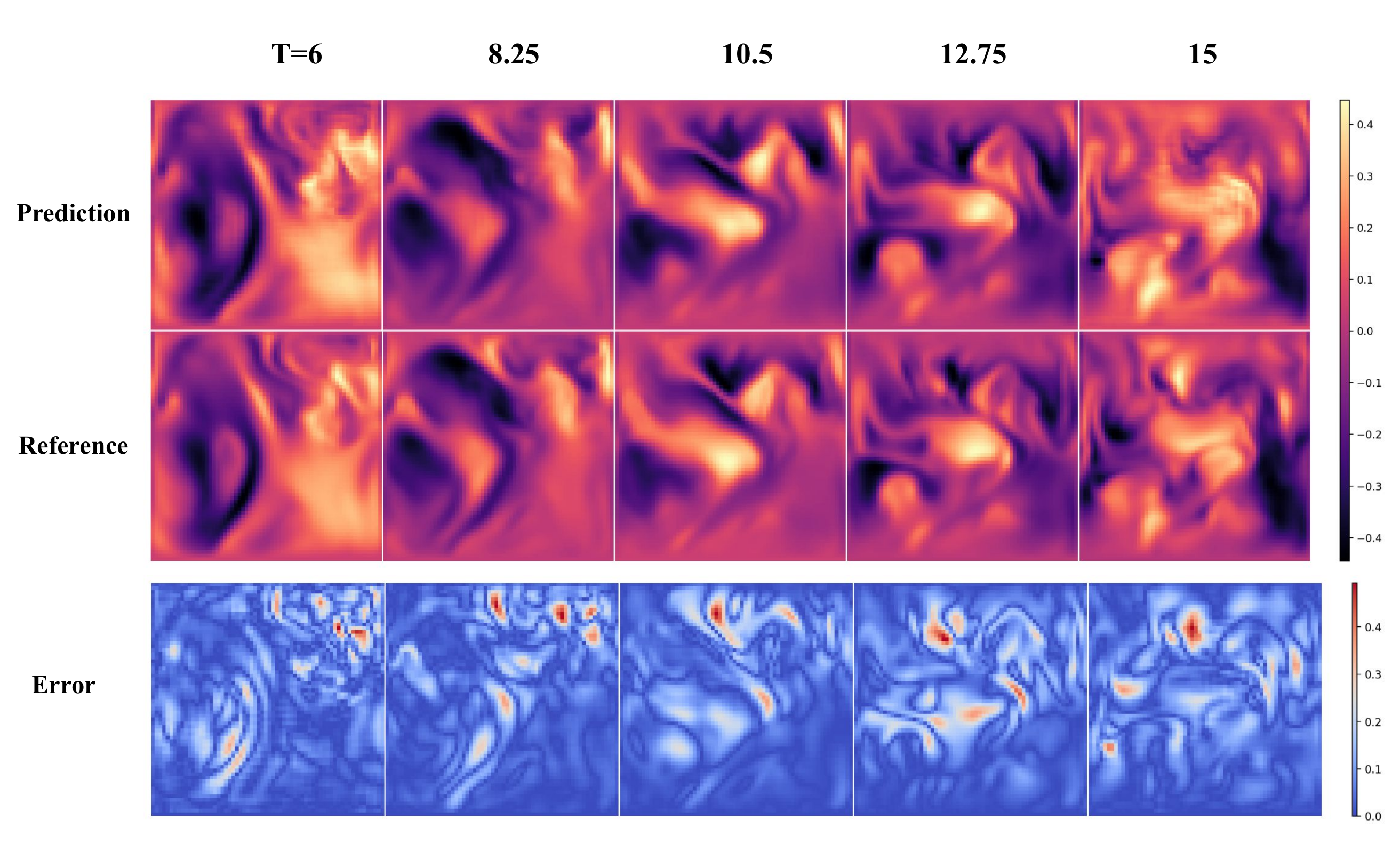}
    \vspace{-3mm}
    \caption{$z$-component of velocity in 3D smoke buoyancy.}
            \vspace{-4mm}
\end{figure}
\begin{figure}[H]
    \centering
    \includegraphics[width=0.95\textwidth]{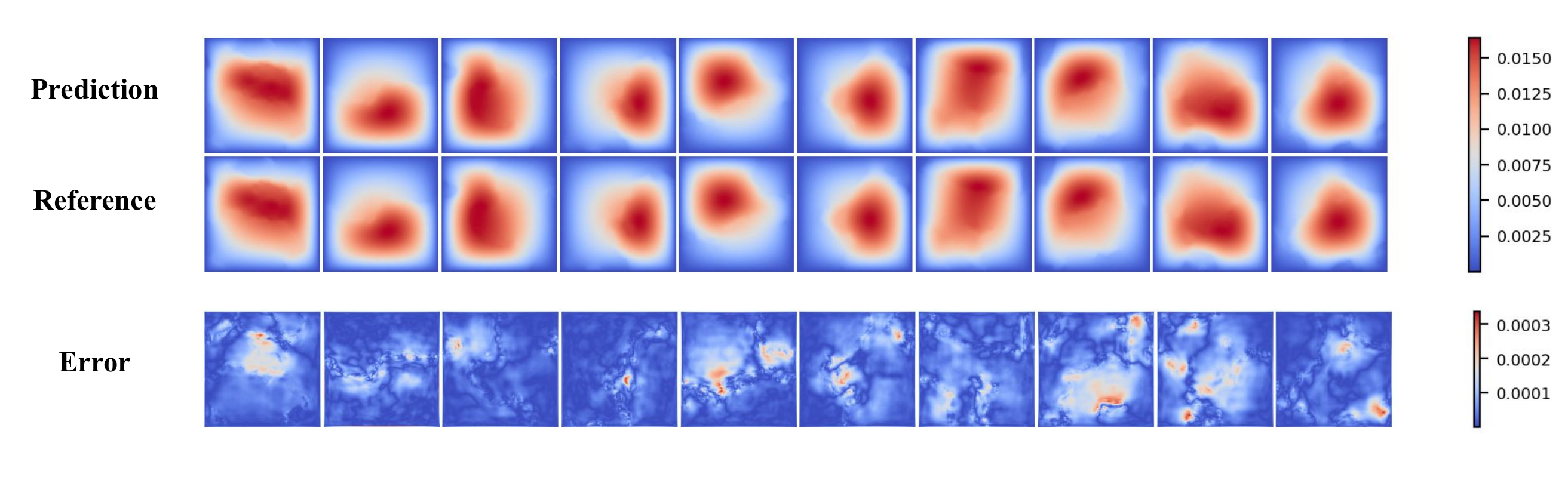}
    \vspace{-3mm}
    \caption{Flow field of 2D Darcy flow.}
\end{figure}
\newpage
\section{Broader impact}
\label{sec:broader impact}
The numerical simulation of PDEs is of extensive application in various fields, such as manufacturing, weather forecasting, and engineering design. Meanwhile, Transformer has shown promising performance on a wide range of data-driven applications including PDE modeling. Our work can help improve the stability and computational efficiency of the attention-based PDE surrogate models.  Our experiments demonstrate that the proposed model serves as an efficient surrogate for numerical solvers, maintaining a balance between accuracy and efficiency, and thus pushing the Pareto front of accuracy-efficiency. However, as there exists a large variety of PDEs and each with very unique properties, there is no guarantee that one type of data-driven model can rule all. Additionally, just like most concurrent works on neural PDE solvers, the long-term stability of the proposed model cannot be guaranteed. Therefore it is important to acknowledge the limitations and potential risks associated with the application of neural PDE solvers.

Apart from enriching the existing architecture design choice of attention-based models, our work also has the potential to be combined with other neural PDE solvers design formulas (e.g., explicitly take into account the relationship between different output variables), or common neural network architectures (e.g., U-Net).

\section{Schematic of Axial Transformer and FactFormer}
\label{sec:axial vs fact}
\begin{figure}[h]
    \centering
    \includegraphics[width=\linewidth]{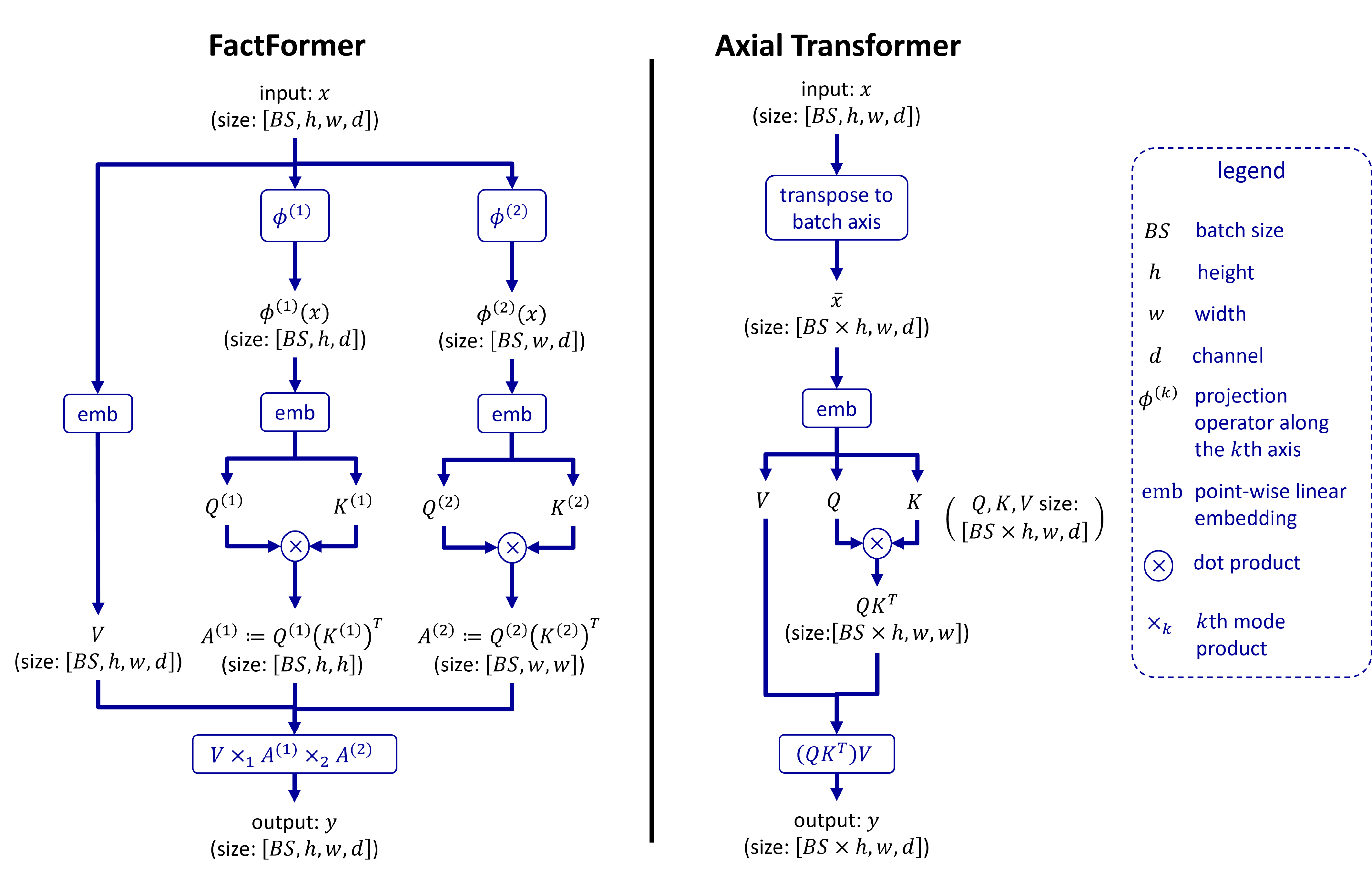}
    \caption{An illustrative example of FactFormer and Axial Transformer applying to 2D input data, with some details such as positional encoding, multi-head mechanism and softmax in Axial Transformer omitted for simplicity. For Axial Transformer, column-wise attention block is shown as an example.}
    \label{fig:axial vs fact}
\end{figure}

\end{document}